\documentclass[letterpaper, 10 pt, conference]{ieeeconf}  % Comment this line out if you need a4paper

\IEEEoverridecommandlockouts                              % This command is only needed if 
\usepackage{times}

\usepackage[table]{xcolor}
\usepackage{colortbl}
\usepackage{graphicx}
\usepackage{array,booktabs}
\usepackage{tabularx}
\graphicspath{{figs/}}
\usepackage{amsmath}
\usepackage{amssymb}
\usepackage{algorithm}
\usepackage{algpseudocode}
\usepackage{array}
\usepackage{makecell}
\usepackage{multirow}
\usepackage[position=bottom]{subfig}
\usepackage{siunitx}
\usepackage{pifont}
\newcommand{\cmark}{\ding{51}}
\newcommand{\xmark}{\ding{55}}
\newcommand{\eqnlab}[1]{\label{eq:#1}}
\newcommand{\eqnref}[1]{Equation \ref{eq:#1}}
\newcommand{\seclab}[1]{\label{section:#1}}
\newcommand{\secref}[1]{Section \ref{section:#1}}
\newcommand{\figlab}[1]{\label{figure:#1}}
\newcommand{\figref}[1]{Fig.\ref{figure:#1}}
\newcommand{\tablab}[1]{\label{table:#1}}
\newcommand{\tabref}[1]{Table \ref{table:#1}}
\newcommand{\argmax}{\operatornamewithlimits{argmax}}

\title{\LARGE \bf
Probabilistic 3D Multilabel Real-time Mapping \\
for Multi-object Manipulation
}

\author{Kentaro Wada and Kei Okada and Masayuki Inaba% <-this % stops a space
  \\
  University of Tokyo, JSK Laboratory \\
  \{wada, okada, inaba\}@jsk.imi.i.u-tokyo.ac.jp%
}

\begin{document}

\maketitle

\begin{abstract}
Probabilistic 3D map has been applied to object segmentation with multiple camera viewpoints,
however, conventional methods lack of real-time efficiency and
functionality of multilabel object mapping.
% Recently, the machine ability of 2D segmentation has been improved
% significantly by works on deep convolutional networks.
% In order to apply these methods to real world tasks with
% robotic manipulation, however, transformation of 2D segmentation to 3D is required,
% because robot needs to understand 3D representation of the world
% to plan collision safe trajectory and effective grasp approaching.
In this paper, we propose a method to generate three-dimensional map
with multilabel occupancy in real-time.
Extending our previous work \cite{wada20163d} in which only target label occupancy is mapped,
we achieve multilabel object segmentation in a single looking around action.
% The mapping method is integrated with 2D object segmentation based on deep learning,
% and we present a 3D segmentation system for multi-objects.
We evaluate our method by testing segmentation accuracy with 39 different objects,
and applying it to a manipulation task of multiple objects in the experiments.
Our mapping-based method outperforms the conventional projection-based method by
40 - 96\% relative (12.6 mean $IU_{3d}$),
and robot successfuly recognizes (86.9\%) and
manipulates multiple objects (60.7\%) in an environment with heavy occlusions.

% Segmenting 3D objects is crutial recognition process for robot
% for environment understanding and planning for manipulation.
% In an environment with occlusions, however,
% conventional segmentation method by registrating 2D segmentation to 3D
% is insufficient, because of lack of point cloud of objects.
% We approach this issue by accumulating segmentation result
% into an occupancy voxel map in each camera viewpoint,
% and generating voxels for each object.
% Our approach consists of two components:
% first is object probability prediction for input RGB image with
% deep convolutional networks,
% and second is generating voxl map with object label occupancy.

\end{abstract}

% 1. Introduction
\section{Introduction}

%%%%%%%%%%%%%%%%%%%%%%%%%%%%%%%%%%%%%%%%%%%%%%%%%%%%%%%%%%%%%%%%%%%%%%%%%%%%%%%%%%%%%%%%%%%%%%%%%%%
% Construction:
% 1. For tasks in real world, 3D object segmentation is important.
% 2. Deep learning works for 2D image segmentation.
% 3. For 3D segmentation, pixel to point registration is used in general,
%    but it does not work with occulusions.
% 4. We propose 3D mapping method for object segmentation
%    to memorize the registration in each view.
%%%%%%%%%%%%%%%%%%%%%%%%%%%%%%%%%%%%%%%%%%%%%%%%%%%%%%%%%%%%%%%%%%%%%%%%%%%%%%%%%%%%%%%%%%%%%%%%%%%

%%%%%%%%%%%%%%%%%%%%%%%%%%%%%%%%%%%%%%%%%%%%%%%%%%%%%%%%%%%%%%%%%%%%%%%%%%%%%%%%%%%%%%%%%%%%%%%%%%%
% ## Identify Issue: 問題提起
Probabilistic three-dimensional map has been applied
to navigation and manipulation in previous works
\cite{ciocarlie2014towards,hornung2013octomap},
however, the generated map has only the collision information,
and is not applicable to multi-object manipulation
because of lacking label information of objects.
%%
% 3D object segmentation, which identifies object class and its location,
% is crucial for robots to understand environment and complete tasks by manipulating objects.
% Recent works on object segmentation with learning-based method
% boosted the ability of 2D object segmentation by machine,
% however, segmenting object three-dimensionally is still a problem because of
% occlusions by the object itself and others as shown in \figref{what_is_this}.
%%
%%%%%%%%%%%%%%%%%%%%%%%%%%%%%%%%%%%%%%%%%%%%%s%%%%%%%%%%%%%%%%%%%%%%%%%%%%%%%%%%%%%%%%%%%%%%%%%%%%%

%%%%%%%%%%%%%%%%%%%%%%%%%%%%%%%%%%%%%%%%%%%%%%%%%%%%%%%%%%%%%%%%%%%%%%%%%%%%%%%%%%%%%%%%%%%%%%%%%%%
% ## Address Issue: 問題展開
Recently, the effectiveness of probabilistic map for object segmentation
is reported \cite{stuckler2012semantic}, which uses the map to improve 2D segmentation result.
However, their research lacks of consideration of real-time efficiency,
which is crucial for 3D segmentation of objects and its manipulation.
On the other hand, we have proposed real-time 3D mapping for a single label object
\cite{wada20163d}.
In order for robot to conduct tasks that demands multi-object segmentation at once;
for example multi-object manipulation (\figref{what_is_this}),
we propose a method to construct three-dimensional map for multi-label objects in real-time.
% another approach is required.
%%
% In general, the pipeline for 3D object segmentation, with large number of object classes,
% consists of 2D segmentation on RGB image and pixel-to-point registration of
% the segmentation result to point cloud \cite{eppner2016lessons}.
% The former component is improved by recent progress in methods
% with deep convolutional networks \cite{long2015fully}\cite{badrinarayanan2015segnet}.
% % however, in most works, the transformation of 2D segmentation to 3D is not tackled
% % more than the registration in a single view.
% As for the transformation of 2D segmentation to 3D,
% previous work tackled this using 3D mapping method in multi-views \cite{wada20163d},
% however, it only generated the map for a single label
% in a situation of single target object picking.
% In order for robot to conduct tasks that demands multi-object segmentation one time;
% for example multi-object manipulation (\figref{what_is_this}),
% another approach is required.
%%
%%%%%%%%%%%%%%%%%%%%%%%%%%%%%%%%%%%%%%%%%%%%%%%%%%%%%%%%%%%%%%%%%%%%%%%%%%%%%%%%%%%%%%%%%%%%%%%%%%%

%%%%%%%%%%%%%%%%%%%%%%%%%%%%%%%%%%%%%%%%%%%%%%%%%%%%%%%%%%%%%%%%%%%%%%%%%%%%%%%%%%%%%%%%%%%%%%%%%%%
% ## Solve Issue: 問題解決
% In this paper,
% we propose a method to construct three-dimensional map for multi-label objects in real-time.
The proposed method extends our previous method for a single label of objects \cite{wada20163d},
and represents object-label and collision with multilabel occupancies in each voxel.
For real-time map generation,
we extend octomap \cite{hornung2013octomap} for multi-label objects,
which is firstly proposed to efficiently map single-label, collision object, occupancy.
By accumulating the 2.5D segmentation results in possible camera viewpoints,
our method segments multiple objects three-dimensionally all at once (\figref{what_is_this}).
We show the efficiency of our method compared to non-mapping-based method
and a multi-object manipulation task in the experiment.
%%%%%%%%%%%%%%%%%%%%%%%%%%%%%%%%%%%%%%%%%%%%%%%%%%%%%%%%%%%%%%%%%%%%%%%%%%%%%%%%%%%%%%%%%%%%%%%%%%%

\begin{figure}[tbhp]
  \centering
  \begin{tabular}{c}
    \subfloat[Manipulating objects with result of 3D segmentation.]{
      \includegraphics[width=0.92\columnwidth]{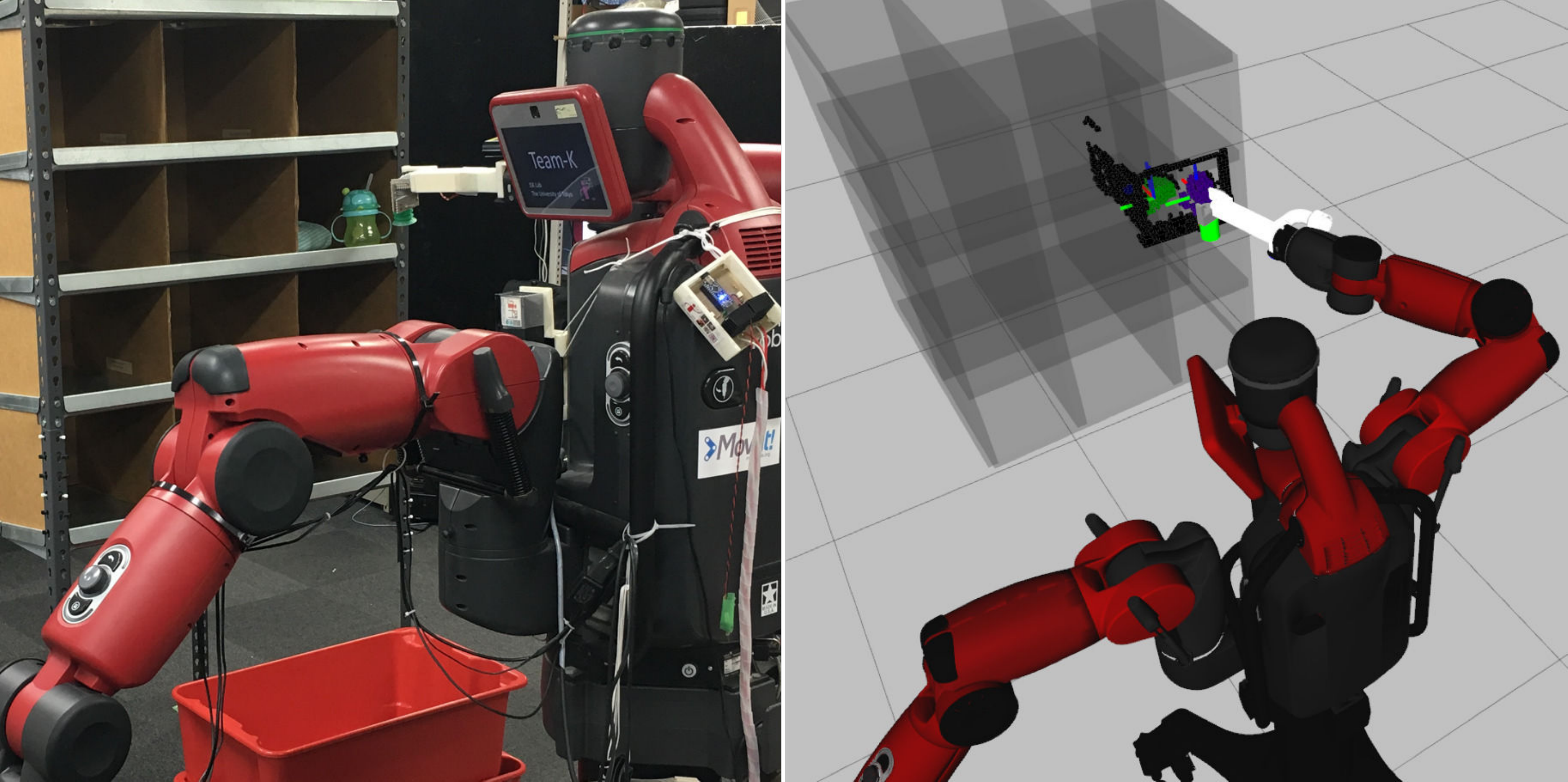}
      \figlab{what_is_this_outlook}
    }
  \end{tabular}
  \begin{tabular}{c}
    \subfloat[View 1]{
      \includegraphics[width=0.29\columnwidth]{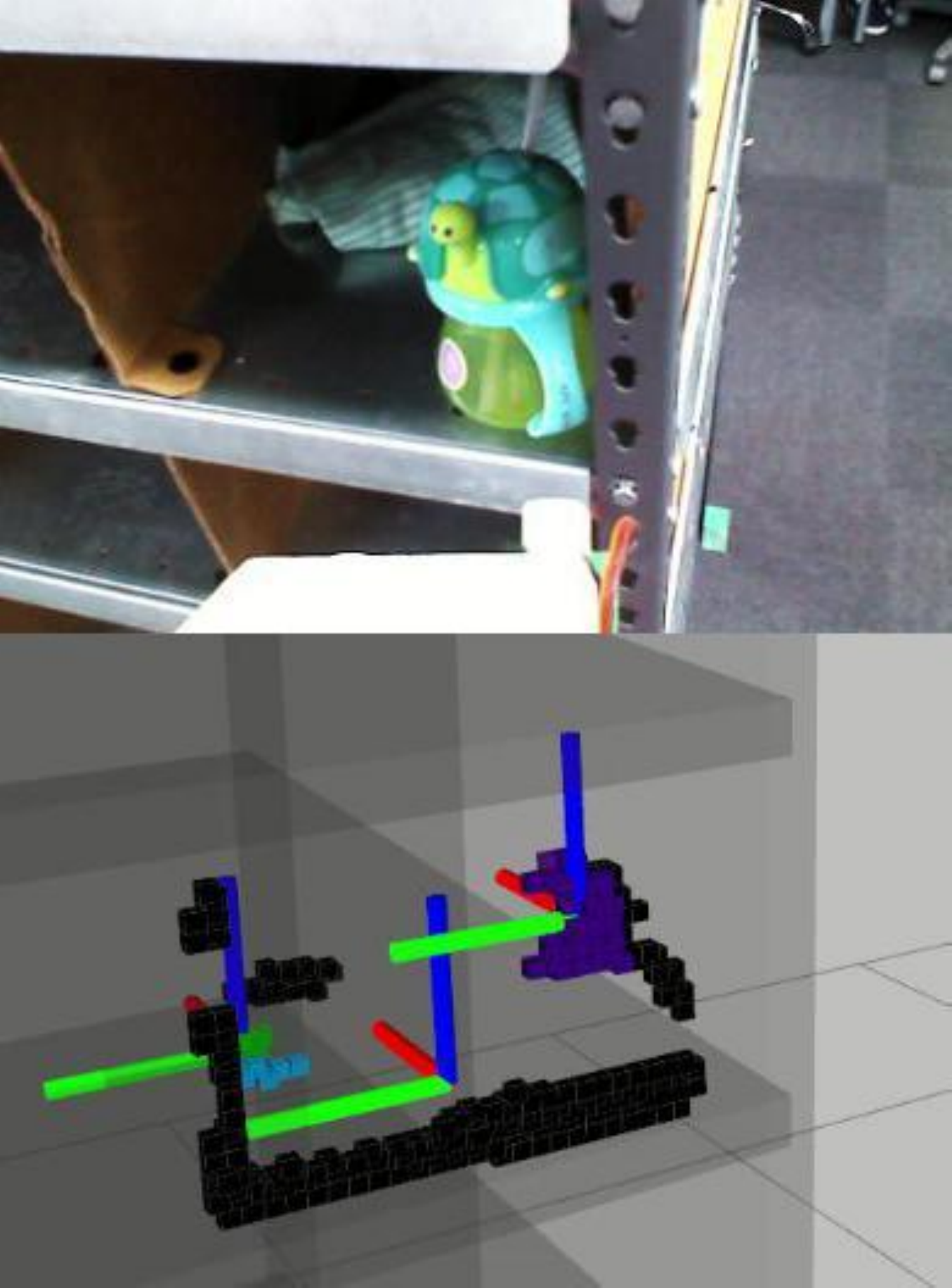}
      \figlab{what_is_this_view1}
    }
    \subfloat[View $\sim$ 3]{
      \includegraphics[width=0.29\columnwidth]{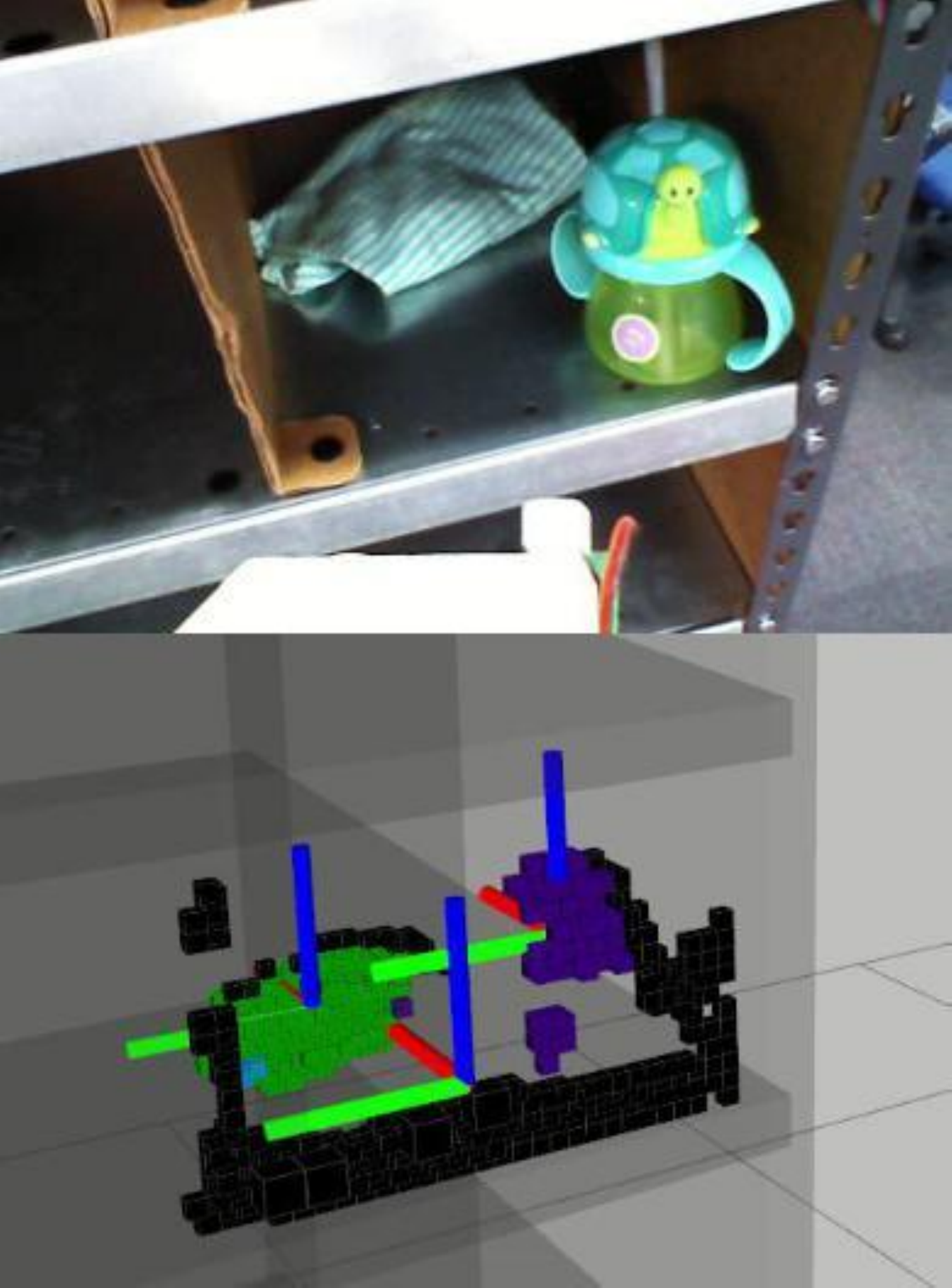}
      \figlab{what_is_this_view2}
    }
    \subfloat[View $\sim$ 5]{
      \includegraphics[width=0.29\columnwidth]{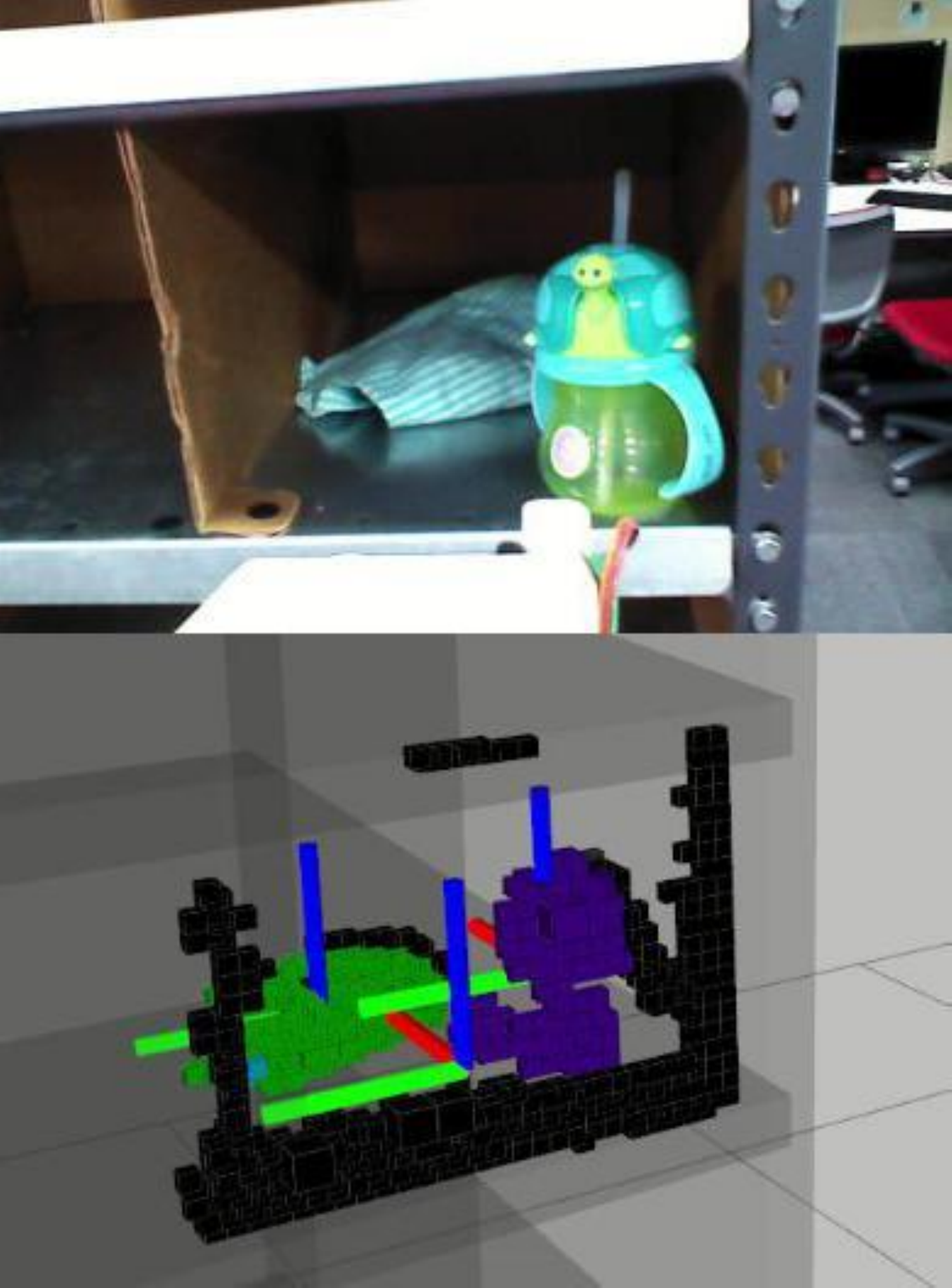}
      \figlab{what_is_this_view3}
    }
  \end{tabular}
  \caption{\textbf{Real-time mapping for multi-object manipulation.}
    \small{
      In this example, target object (t-shirts) is occluded by non-target (green cup),
      and robot needs to remove the obstacle object in order to pick the target
      with recognizing the necessity of doing so.
    }
  }
  \figlab{what_is_this}
\end{figure}
\newpage

% 2. Semantic Object Localization for Humanoids Picking
\section{3D Multilabel Mapping for \\ Object Segmentation and Manipulation}

\subsection{Related Works}

% -------------------------------------------------------------------------------------------------
\subsubsection{Object Segmentation}
% ### 物体セグメンテーションに関する研究: 2D-3D, Mapping-Maplessの対比
In previous works, 2D object segmentation is tackled as
a contour finding problem with a guide of region-of-interest
\cite{rother2004grabcut}\cite{xu2007object},
a superpixel classification
\cite{fulkerson2009class}\cite{gould2008multi}
and pixel-wise classification
\cite{eppner2016lessons}\cite{long2015fully}\cite{badrinarayanan2015segnet}
with learning-based approach with class segmentation dataset.
In addition to these works on 2D segmentation,
three-dimentional segmentation is required for robot to conduct tasks in the real world.
In order to achieve this, previous works propose
projection-based approach projecting segmented pixels to 3D points in a single view (2.5D)
\cite{eppner2016lessons},
mapping-based approach with binary object existence \cite{zeng2016multi} and
probabilistic existence \cite{wada20163d} for a single target object.
And as for fully 3D-based approach, model matching is tackled
\cite{mian2006three}\cite{hinterstoisser2011multimodal}
using various 3D features \cite{aldoma2011cad}\cite{rusu2009fast}.
We use a mapping-based approach with multiple views
extending our previous method \cite{wada20163d} for multi-label objects
to deal with object occlusions and
flexible objects for which static 3D model is less effective.
Our method segments multi-label objects in a single multi-view action,
and effective to acquire the dense 3D information of objects
in an environment with heavy occlusions: self-occlusion and occlusion by others.
% This approach works in a occlusion free environment
% where the 3D information of object is easily sensed with RGB-D sensor.
% and different approach to collect the object 3D information is required.
% We accumulate the 3D segmentation result as a map by sensing with multiple camera angles,
% in order to acquire dense 3D segmentation result based on the generated map.
% -------------------------------------------------------------------------------------------------

% -------------------------------------------------------------------------------------------------
\subsubsection{Probabilistic Grid Map}
% ### 確率的グリッドマップに関する研究: 各グリッドの確率表現の対比
In previous works on probabilistic grid map,
object exploration with updating object existence probability on 2D map
\cite{furukawa2006recursive},
collision object \cite{hornung2013octomap}\cite{elfes1989using}
and object segmentation on 3D map \cite{stuckler2012semantic}\cite{wada20163d}
are tackled.
Our proposed method is most closely related to two prior works
\cite{stuckler2012semantic}\cite{wada20163d},
and the contribution of this paper is the proposed octomap
that has a single octree with multilabel probabilities in the nodes.
Compared to the previous methods to update a single-label occupancy,
our method updates the multilabel occupancies in each voxel,
and segments all objects in a single mapping.

\subsubsection{Multi-object Manipulation}
We define multi-object manipulation as a task to handle multiple objects
with determination of the order to manipulate.
Its typical situation is a picking task with occluded target object,
where robot needs to recognize both the target and occluding objects
in order to remove firstly the occluding and pick the target later.
In previous works, multi-object manipulation is tackled
in simulation \cite{moldovan2012learning},
and in a simple real world environment with single-class objects
\cite{harada2006pushing}\cite{pajarinen2015robotic},
We address manipulation task of multi-class multiple objects
in a clutter environment that contains heavy occlusions.
% -------------------------------------------------------------------------------------------------

% -------------------------------------------------------------------------------------------------
\subsection{Proposed Method and System}

\begin{figure}[htbp]
  \centering
  \includegraphics[width=0.48\textwidth]{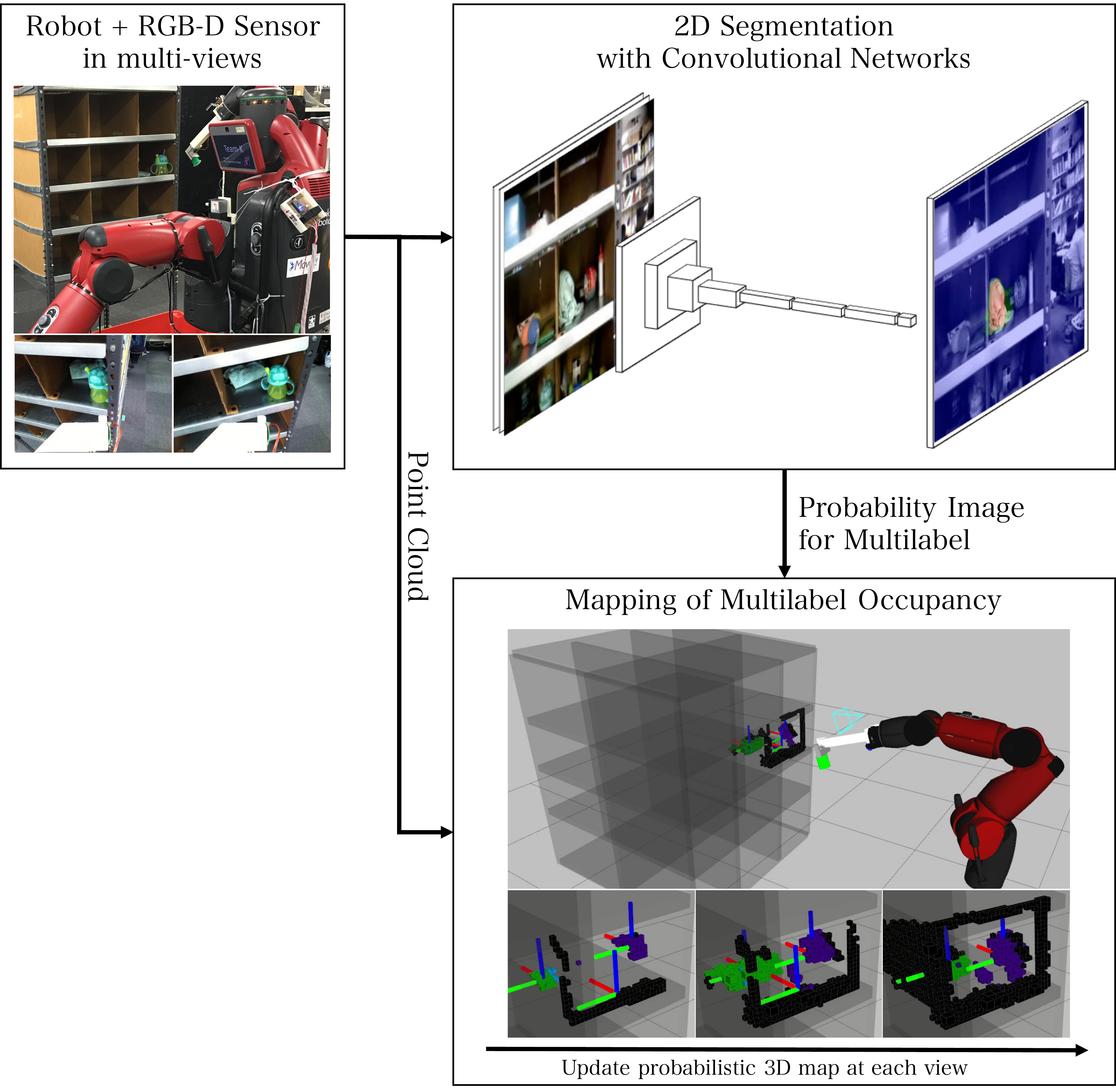}
  \caption{\textbf{Proposed system for 3D object segmentation.}}
  \figlab{proposed_system}
\end{figure}

% ### 提案手法
% LabelOctoMap
In this paper, we propose a probabilistic mapping method for multilabel occupancies,
in which each voxel grid has probabilities for all object labels
($n_{label} \sim 40$ in the experiment).
In a voxel, each label occupancy is updated separately and the maximum occupancy is used
to transform the map to voxels: if the maximum occupancy is over a threshold,
it is occupied by the label and otherwise unoccupied.

We also propose a 3D segmentation system shown in \figref{proposed_system},
which is extended version of system in prior work \cite{wada20163d} for multilabel segmentation.
Our system consists of 2 components: 2D segmentation and 3D mapping.
In the first component,
2D object probability image $I^{prob}_t$ is predicted at time $t$ from input image $I_t$,
and the output probability image $I^{prob}_t$ has same number of channels as
the number of object classes $n_{label}$.
In the second component, our 3D mapping method for multilabel $M_t = g(M_{t-1}, I^{prob}_t, C_t)$
is used to update the grid map $M_{t-1}$ with input point cloud $C_t$,
after projecting the pixel to 3D point with index of height and width axies $u, v$,
\begin{eqnarray}\eqnlab{registration}
  z_{t,uv} = (C_{t,uv}, I^{prob}_{t,uv})
\end{eqnarray}
with correspondence between image and point cloud:
assuming that the point cloud comes from RGB-D camera sensor.
By accumulating the object probability prediction result three-dimensionally as a map,
we improve 3D object segmentation result in a situation with occlusions
and generate dense 3D object voxel map.

In the following sections, we introduce the method to predict object probability
from input image in \secref{fcn}, and the method of 3D mapping in \secref{octomap}.
In \secref{experiment}, we evaluate the efficiency of our proposed method
with quantitative evaluation of the segmentation and
achievement of multi-object manipulation by a robot.
% -------------------------------------------------------------------------------------------------

% 3. 2D Object Segmentation with Convolutional Networks
% ## 畳み込みニューラルネットワークによる物体セグメンテーション
\section{Probabilistic Object Segmentation \\ in a Single View}  \seclab{fcn}

In this section, we introduce the function $(C_{t,uv}, I^{prob}_{t,uv}) = f(I_t)$ which receives
RGB image $I_t$ as the input and outputs multi-class probabilities $I_{t,uv}^{prob}$
and 3D point $C_{t,uv}$ at pixel $(u, v)$ as the output.
We briefly describe the network model for this function
because it is same as that developed in our previous work \cite{wada20163d},
and discuss about the efficiency by using 2D segmentation
as a component in the 3D segmentation system.
% The network model, dataset and training are same as that in our previous work \cite{wada20163d},
% We construct deep neural network for this function,
% and we describe the network architecture and the way of training in the following subsections.
% Most parts of this section are same as the previous our work \cite{wada20163d} to appear.

% ### ネットワーク構造
\subsection{Convolutional Network Model}

We use a previously proposed convolutional network model for 2D segmentation,
Fully Convolutional Networks (FCN) \cite{long2015fully},
applying a small change to convert n-class object score map $I^{score}$,
the output of the network,
to probability image $I^{prob}$ using pixel-wise softmax
with index for height $u \in [0, H]$,
width $v \in [0, W]$ and channel $k \in [0, n_{label}-1]$ for image height $H$ and width $W$:

\begin{eqnarray}
  I^{prob}_{uv} = \sigma (I^{score}_{uv})
    = \frac{exp(I^{score}_{uv})}{\sum^N_{k=0}exp(I^{score}_{uvk})}.
\end{eqnarray}

\subsection{Projection of 2D segmentation as 3D points}

For 3D mapping, the pixel-wise 2D segmentation needs to be projected to 3D points.
By using a calibrated camera, we can project
the pixel point in 2D segmentation to the 3D point:
\begin{eqnarray} \eqnlab{projection1}
  \begin{bmatrix}
    u \\
    v \\
    Z_c
  \end{bmatrix}
  =
  K
  \begin{bmatrix}
    X_c \\
    Y_c \\
    Z_c
  \end{bmatrix},
  K =
  \begin{bmatrix}
    f_x  & 0  & c_x \\
    0 & f_y & c_y \\
    0 & 0 & 1
  \end{bmatrix}.
\end{eqnarray}
$(X_c, Y_c, Z_c)$ are the coordinates of a 3D point in the camera coordinate space,
$(u, v)$ are the coordinates of a projection point in camera pixels,
and $K$ is the matrix of camera intrinsic parameters.

Using RGB-D camera,
whose extrinsic parameters between color and depth sensor coordinate are well calibrated,
the pixelwise depth $d$ in color coordinate is given,
and the 3D point coodinates are given as follows:
\begin{eqnarray}
  X_c = \frac{u - c_x Z_c}{f_x}, Y_c = \frac{v - c_y Z_c}{f_y}, Z_c = d.
\end{eqnarray}
Therefore, 3D point in world coordinate space is given
with a homogeneous transformation matrix $A$ from camera to world coodinate:
\begin{eqnarray} \eqnlab{projection2}
  \begin{bmatrix}
    X \\ Y \\ Z \\ 1
  \end{bmatrix}
  =
  A
  \begin{bmatrix}
    X_c \\ Y_c \\ Z_c \\ 1
  \end{bmatrix},
  A =
  \begin{bmatrix}
    r_{11} & r_{12} & r_{13} & t_{1} \\
    r_{21} & r_{22} & r_{23} & t_{2} \\
    r_{31} & r_{32} & r_{33} & t_{3} \\
    0 & 0 & 0 & 1
  \end{bmatrix}
\end{eqnarray}
In the experiment,
since the RGB-D sensor is mounted on the robotic arm and the task is conducted without navigation,
we use the transformation matrix $A$ computed from joint angles of the arm and forward kinematics.

In the 3D mapping we describe at afterward sections,
the set of multi-class probabilities $I^{prob}_{uv}$, and 3D point $C_{t,uv} = (X, Y, Z)$
is used to accumulate the pixel-wise 2D segmentation into 3D map.

% 4. Map generation for 3D Object Segmentation
% ## Map Generation
\section{3D Mapping of Multilabel Occupancy} \seclab{octomap}

In this section, we introduce the function $M_t = g(M_{t-1}, I_{t,uv}^{prob}, C_{t,uv})$
to generate voxel grid map $M_t$ from 2D probability map $I_t^{prob}$ and input point cloud $C_{t,uv}$.
We describe the method to contruct voxel map for multilabel occupancy,
and to update the map in real-time.

% -------------------------------------------------------------------------------------------------
% ### 占有確率をグリッドによって更新するマップ更新法
\subsection{Generating Occupancy Map for Multilabel}

% % 導入: What is occupancy grid map?
% In this section, we introduce LabelOctoMap
% which is map updating method for object segmentation with multi-class,
% and the variation of Occupancy Grid Map \cite{elfes1989using},
% where the environment is splitted into cells with static grid size
% and the occupancy of each cell is represented probabilistically.
% The map is usually used for path planning for mobile robot navigation \cite{elfes1989using} and
% reaching behavior of robot \cite{murooka2016planning},
% but LabelOctoMap is designed for multi-class object segmentation
% by replacing occupancy probability in conventional occupancy map with object probability
% for each class.

% 確率の値の意味
In conventional occupancy grid map \cite{elfes1989using},
the $l$-th cell $m_s$ holds the probability of occupancy for collision object $p^{collision}(m_s)$.
In our occupancy voxel map, on the other hand,
each grid has the same number of probabilities as the number of labels $n_{label}$.
Each probability has the value between 0 and 1:
\begin{eqnarray}
  0 \le p_k(m_s) \le 1 \quad (k = 0 \cdots n_{label}-1) \nonumber
\end{eqnarray}
and meaning of the value is represented as following:
\begin{eqnarray} \eqnlab{reconstruction}
  \begin{cases}
    m_s: unoccupied      & (\max_{k}(p_k(m_s)) < p_{thresh}) \\
    m_s: occupied~by~l_s & (\max_{k}(p_k(m_s)) >= p_{thresh}) \\
  \end{cases}
\end{eqnarray}
with occupied label $l_s = \argmax_k (p_k(m_s))$.
The occupancy threshold $p_{thresh}$ is set to $0.5$ in our experiments.

% \begin{eqnarray}
%   b_s = \argmax_{k}(p_k(m_s)) \\
%   p_{k,max}(m_s) = \max_{k}(p_k(m_s))
% \end{eqnarray}

% So the LabelOctoMap with $n_{label}=1$ has equivalent function with the conventional occupancy voxel map
% , in which the single label represents collision object.

% LabelOctoMap Specific
We explain the method of updating $p_k(m_s)$ from sensor information $z_t$;
$t$ is the index in series of sensor input,
$z_t$ is the set of sensor information at the time index $t$,
and $z_{1:t}$ means the set of the sensor information from first to the $t$-th time.
Notice that $z_{t,uv}$ (\eqnref{registration}) is
set of the point cloud $C_t$ and 2D probability image $I_t^{prob}$,
and the probability in each pixel $I_{t,uv}^{prob}$
is registered to a 3D point $C_t = (X, Y, Z)$ by
\eqnref{projection1} - \ref{eq:projection2}.

By computing $p_k(m_s | z_{1:t})$ with $p_k(m_s | z_{1:t-1})$ and $z_t$,
we can integrate new sensor information and update the map periodically.
% and we firstly acquire $p_k(m_s | z_t)$ from $z_t$.
The grid map $M_t$ is the set of voxels
with multilabel probabilities accumulated in them $p_k(m_s | z_{1:t})$:
$m_s$ is one of the voxels whose size is $S$: $s = 0 \cdots S$.
Using each 3D point $z_{t,uv} = (C_{uv}, I^{prob}_{uv})$,
occupancies in multiple voxels are updated,
because the sensed 3D point $C_{uv}$ has two parts of information:
the voxel at $C_{uv}$ has collision object (hit),
the voxels on the ray from camera coordinate to $C_{uv}$ has no collision objects (miss).
Using these information, the new probability $p_k(m_s | z_t)$ is acquired as following
with $m_s$ is given by $C_{t, uv}$:
\begin{eqnarray}
  \begin{cases}
    p_k(m_s | z_t) = I_{t, uvk}^{prob} & (m_s: hit)  \\
    p_k(m_s | z_t) = p_{miss}          & (m_s: miss) \\
  \end{cases}
\end{eqnarray}
% The probability acquired from the new sensor information $z_t$ is $p_k(m_s | z_{1:t-1})$,
% and multiple voxels are updated for a 3D point with registered object probability $z_{t,uv}$.
% The equation to update each voxel is mostly same as the one for the conventional occupancy map,
% but in our label occupancy map the probability that is added to the map at each time is acquired from
% the probabilistic 2D segmentation, with the index of height $i$, width $j$ and channel $k$.
% \begin{eqnarray}
%   \begin{cases}
%   p_k(m_s | z_t) = I_{t, uvk}^{prob} & (C_{t, uv} = valid) \\
%   p_k(m_s | z_t) = p_{miss} & (C_{t, uvk} = invalid) \\
%   \end{cases}
% \end{eqnarray}
In the following experiments,
we used $p_{miss} = 0.3$, which is the probability for the voxel where the point cloud is missing.

% 確率更新
% 式
Next we need to compute $p_k(m_s | z_{1:t})$ from $p_k(m_s | z_{1:t-1})$ and $p_k(m_s | z_t)$.
Following formula is derived from Bayes' theorem and independence of the conditional probability:
\begin{eqnarray}
  p_k(m_s | z_{1:t})
  &=& \frac{p_k(z_t | m_s, z_{1:t-1}) \ p_k(m_s | z_{1:t-1})}{p_k(z_t | z_{1:t-1})} \nonumber\\
  &=& \frac{p_k(m_s | z_t) \ p_k(z_t) \ p_k(m_s | z_{1:t-1})}{p_k(m_s) \ p_k(z_t | z_{1:t-1})}.
\end{eqnarray}
The ratio of occupied probability $p_k(m_s | z_{1:t})$
and free probability $p_k(\lnot m_s | z_{1:t})$ becomes simple as following:
\begin{eqnarray}
  \frac{p_k(m_s | z_{1:t})}{p_k(\lnot m_s | z_{1:t})} =
  \frac{p_k(m_s | z_{1:t-1})}{p_k(\lnot m_s | z_{1:t-1})}
  \frac{p_k(m_s | z_t)}{p_k(\lnot m_s | z_t)} \frac{p_k(\lnot m_s)}{p(m_s)}.
\end{eqnarray}
We can deform the formula by introducing log-odds (logit) as following:
\begin{eqnarray}
  {\cal L}(m_s | z_{1:t}) = {\cal L}(m_s | z_{1:t-1}) + {\cal L}(m_s | z_t) - {\cal L}(m_s). \\
  {\rm where} \ \ \ \ {\cal L}(x) = {\rm logit}(p(x)) = \log (p(x)/(1-p(x))) \nonumber
\end{eqnarray}
Presuming that we have no prior knowledge of the environment at the first time ($t = 0$):
${\cal L}(m_s) = 0~(\because p(m_s) = 0.5)$,
therefore ${\cal L}(m_s | z_{1:t})$ is updated as following:
\begin{eqnarray}\eqnlab{update}
  {\cal L}(m_s | z_{1:t}) = {\cal L}(m_s | z_{1:t-1}) + {\cal L}(m_s | z_t).
\end{eqnarray}
We can update the map probabilistically using this formula, and
obtain probability $p_k(m_s | z_{1:t})$ from the log-odds ${\cal L}(m_s | z_{1:t})$ as following:
\begin{eqnarray}
  p_k(m_s | z_{1:t}) = 1 - (1 + \exp( {\cal L}(m_s | z_{1:t}) ))^{-1}.
\end{eqnarray}
% -------------------------------------------------------------------------------------------------

\subsection{Extending OctoMap for Multilabel Occupancy}

For real-time updation of the map,
we extend the OctoMap \cite{hornung2013octomap},
which is an efficient framework to generate occupancy grid map using octree.
In the octomap with multilabel occupancy (LabelOctoMap), which we propose,
each node in octree has multiple probabilities
and the number of nodes is same as that of voxels $S$,
which depends on the resolution of voxels not the number of labels.
So the search of the voxel to be updated from 3D coordinate in LabelOctoMap
is as efficient about speed as the conventional octomap.

Having a single tree is also effective for reconstruction of voxels from the map
compared to the use of multiple octrees for multi-labels.
As shown in \eqnref{reconstruction}, we find label with maximum likelihood for each voxel,
and this operation is required because it is possible that probabilities of multiple labels
can be over threshold $p_{threshold}$ for voxels at the same 3D coordinate.
The comparison of probabilities is time consuming when using multiple octrees,
because each comparison requires the search of the corresponding voxel in a tree
from 3D coordinate of the voxel in other tree.
\section{Experiments}  \seclab{experiment}

In this section, we show the evaluation result of our proposed method
with the segmentation accuracy tests and the robotic manipulation task application.

% -------------------------------------------------------------------------------------------------
\subsection{Evaluate 2D Segmentation}

We describe the evaluation results of the trained model,
in order to show the effectiveness of using 2D segmentation as a component of the system,
though there are previous works which use fully 3D approach
\cite{mian2006three}\cite{hinterstoisser2011multimodal}.
We use the validation dataset ($\sim1900$)
with splitting the whole dataset in 4:1 for training and validation.

\subsubsection{Dataset and training}
% ### 学習方法
% \subsection{Training the Network with Dataset}

We handle 39 objects (\figref{objects}), which were used at Amazon Picking Challenge 2016 (APC2016),
in an environment shown in \figref{what_is_this}.
So the number of object classes is 40: 39 item labels and 1 background label.
The number written in \figref{objects} is each item's label and 0 is the background label.
For dataset, we combined the dataset we previously collected \cite{wada20163d} ($\sim250$)
and that is published by other work \cite{zeng2016multi} ($\sim7300$).
% In spite of the large number of classes,
% we achieve close accuracy to previous work \cite{long2015fully}
% with small number of images as the dataset:
% 218 images in setup with single/multiple objects in a shelf-bin.
% It is conceivable that this is because of
% the limited background of objects, shelf-bin (\figref{what_is_this}),
% and fine-tuning technique \cite{yosinski2014transferable}
% with pretrained network for classification
% of larger number of classes ($\sim 1000$) \cite{simonyan2014very}.
% 153 images in setup with single object and 65 with multiple objects.
% % 40クラスのセグメンテーション問題であり，そこそこの難しさ．
% We handle items shown in \figref{segmentation_problem}, so the
% whole object labels for segmentation is 40 labels: 39 labels for the items and 1 background label.
% Compared to the previous researches \cite{everingham2015pascal} \cite{long2015fully},
% which handle 21 object classes, the number of that we are handling is larger.
% But in our segmentation experiment, the objects are only located inside the shelf, and the location
% where objects is located is limited.
%
\begin{figure}[tbhp]
  \centering
  \includegraphics[width=\columnwidth]{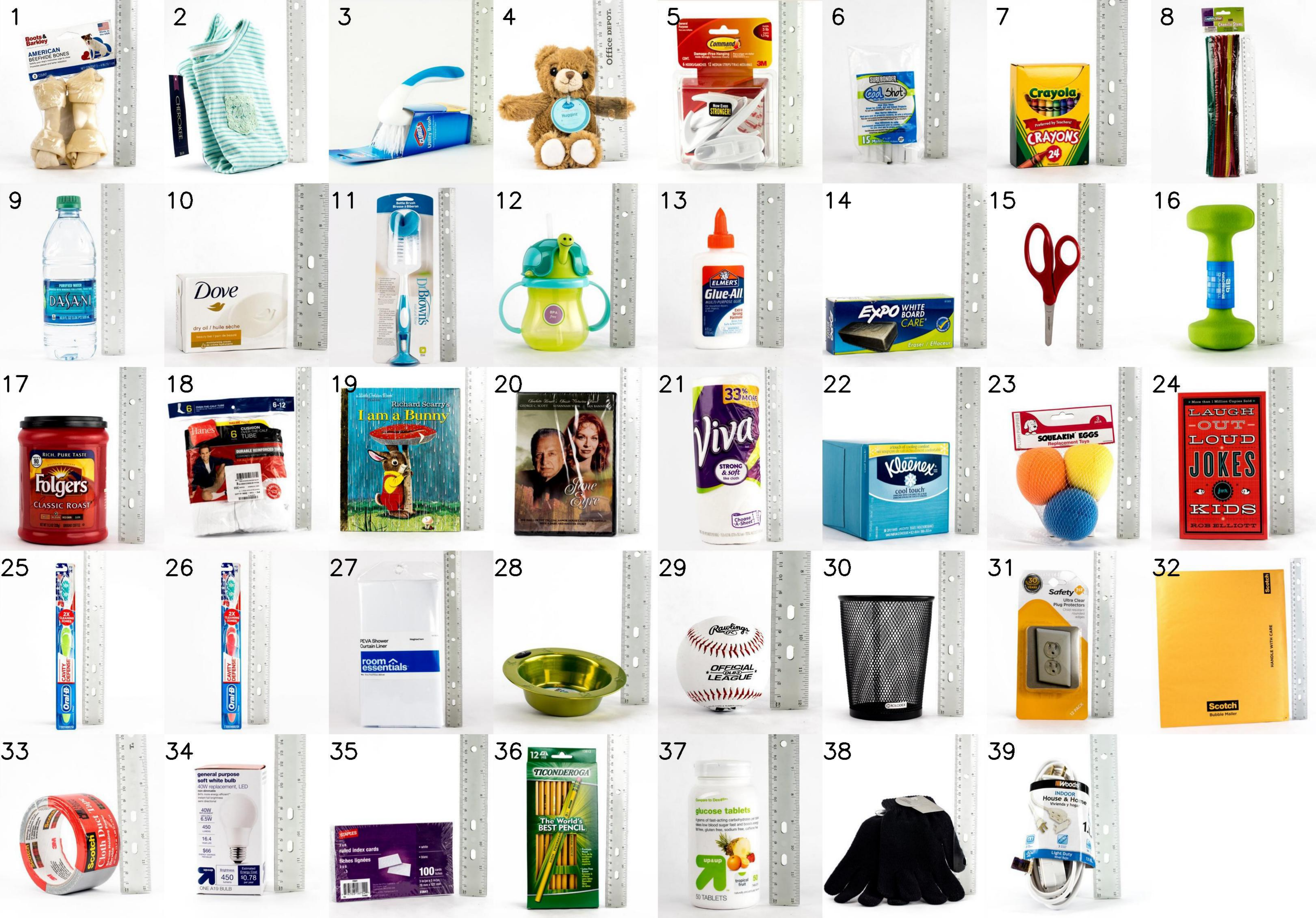}
  \caption{\textbf{39 objects used at APC2016.} \protect\footnotemark}
  \figlab{objects}
\end{figure}
\footnotetext{These item images are originally corrected by Amazon Robotics and modified by the authors.}

\subsubsection{Evaluation Metric}

The metric of quantitative evaluation is same as one used in previous work \cite{long2015fully}:
\begin{itemize}
  \item pixel accuracy: $\sum_i n_{ii} / \sum_i t_i$
  \item mean accuracy: $(1/n_{label}) \sum_i n_{ii} / t_i$
  \item mean intersect-over-union (IU): \\
    $(1/n_{label}) \sum_i n_{ii} / (t_i + \sum_j n_{ji} - n_{ii})$
  \item frequency weighted IU: \\
    $(\sum_k t_k)^{-1} \sum_i t_i n_{ii} / (t_i + \sum_j n_{ji} - n_{ii})$
\end{itemize}
where $n_{ij}$ is the number of pixels of class $i$ predicted to belong to class $j$,
$n_{label}$ the number of classes,
and $t_i = \sum_j n_{ij}$ is the total number of pixels of class $i$.
All metrics have value range $[0, 1]$ so we values multiplied by 100.

\subsubsection{Result}
The quantitative result of the FCN model with different datasets (\tabref{fcn_quantitative_result})
shows that the model trained with our dataset is as good as that with others
with larger values in all metrics.

\setlength{\tabcolsep}{3pt}
\definecolor{Gray}{gray}{0.85}
\begin{table}[htbp]
  \centering
  \caption{\textbf{FCN segmentation results in different datasets.}}
  \tablab{fcn_quantitative_result}
  \begin{tabular}{c|cccc}
    dataset & \shortstack{pixel acc.} & \shortstack{mean acc.} & \shortstack{mean IU} & \shortstack{f.w. IU} \\
    \hline
    \shortstack{VOC2012 \cite{long2015fully}} & 89.1 & 73.3 & 59.4 & 81.4 \\
    \rowcolor{Gray}
    APC2016 & 98.2 & 93.5 & 84.7 & 96.8
  \end{tabular}
\end{table}

We also evaluated the model comparing with other 2D segmentation methods used
by the winner at Amazon Picking Challenge in 2015 \cite{jonschkowski2016probabilistic}.
Since they do not use deep learning,
we can see how effective the segmentation method using deep learning is.
The result is shown in \tabref{seg2d_model_comparison},
and we compared with two of thier proposed methods:
BP (Histogram backprojection) and RF (Random forest),
both with and without class candidates.

\begin{table}[htbp]
  \centering
  \caption{\textbf{Model comparison.}
    \small{We used only the dataset previously collected \cite{wada20163d}
           (train: $\sim200$, validation: $\sim50$).}
  }
  \tablab{seg2d_model_comparison}
  \begin{tabular}{cc|cccc}
    model & w/ candidates & \shortstack{pixel acc.} & \shortstack{mean acc.} & \shortstack{mean IU} & \shortstack{f.w. IU} \\
    \hline
    BP \cite{jonschkowski2016probabilistic} & no & 67.8 & 22.1 & 9.8 & 54.2 \\
    RF \cite{jonschkowski2016probabilistic} & no & 57.7 & 37.4 & 17.4 & 47.6 \\
    \rowcolor{Gray}
    FCN & no & 93.0 & 66.0 & 53.6 & 87.8 \\
    \hline
    BP \cite{jonschkowski2016probabilistic} & yes & 73.8 & 47.4 & 39.2 & 56.6 \\
    RF \cite{jonschkowski2016probabilistic} & yes & 79.1 & 67.0 & 55.4 & 63.8 \\
    \rowcolor{Gray}
    FCN & yes & 94.3 & 74.3 & 67.1 & 89.4
  \end{tabular}
\end{table}

The qualitative result is shown with ground truth in \tabref{fcn_qualitative_result},
and it shows that the model successfully segments flexible objects (object 2,23),
transparent objects (object 9), in environment with occlusions by object each other.
These results represent the effectiveness of 2D segmentation
compared to the 3D model-based approaches,
which are hard to use for flexible and transparent objects
because of missing of 3D model and insensible depth respectively.

\newcommand{\imgwidth}{2.0cm}
\begin{table}[htbp]
  \centering
  \caption{\textbf{Qualitative results of 2D segmentation.}
    \small{
      The image in the final row shows the correspondence between object label in \figref{objects}
      and label color.
    }
  }
  \tablab{fcn_qualitative_result}
  \begin{tabular}{c|ccc}
   & image &  inference &  gt. \\
  \hline
  scene 1 &
  \parbox[c]{\imgwidth}{\includegraphics[width=\imgwidth]{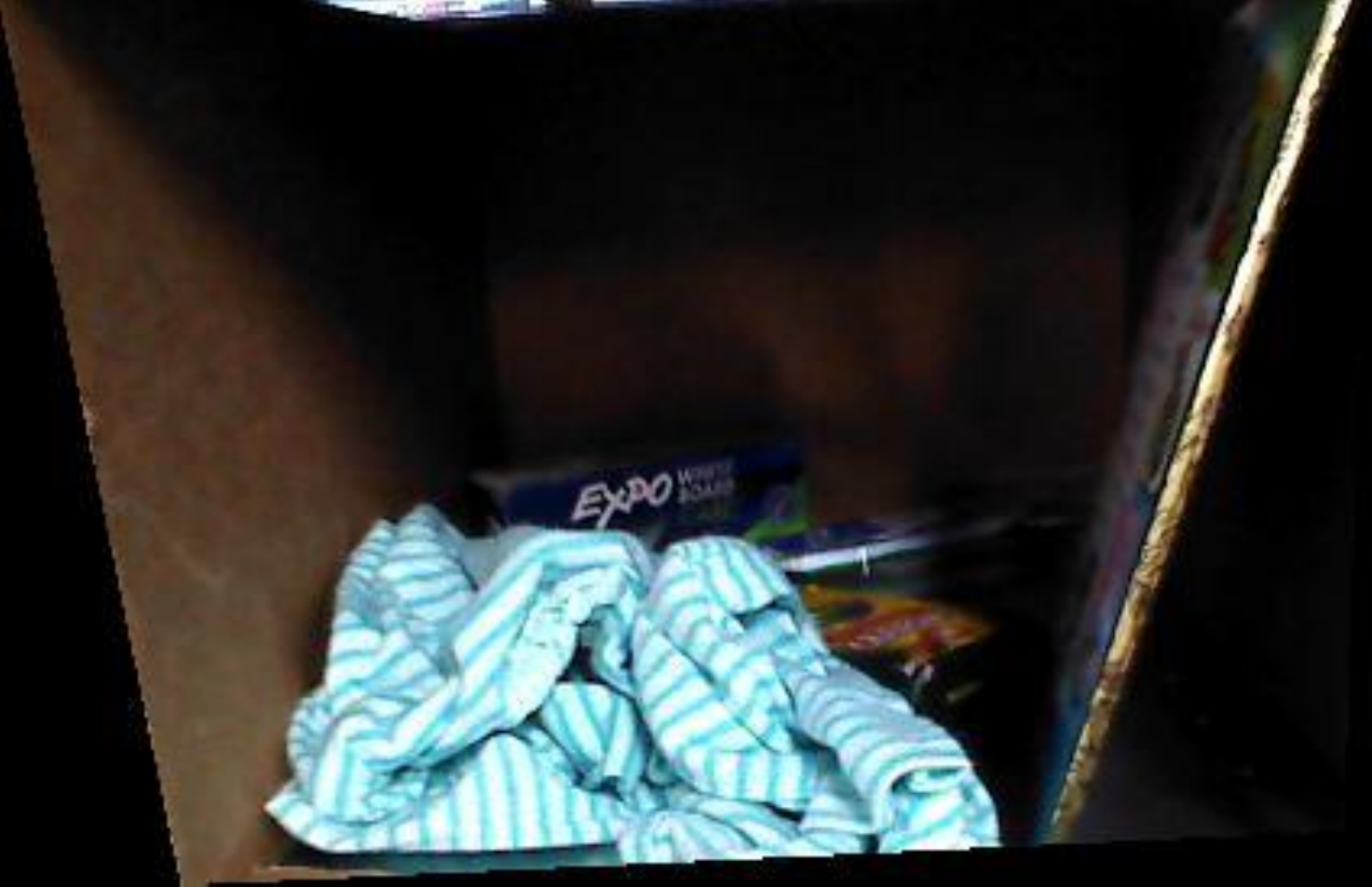}} &
  \parbox[c]{\imgwidth}{\includegraphics[width=\imgwidth]{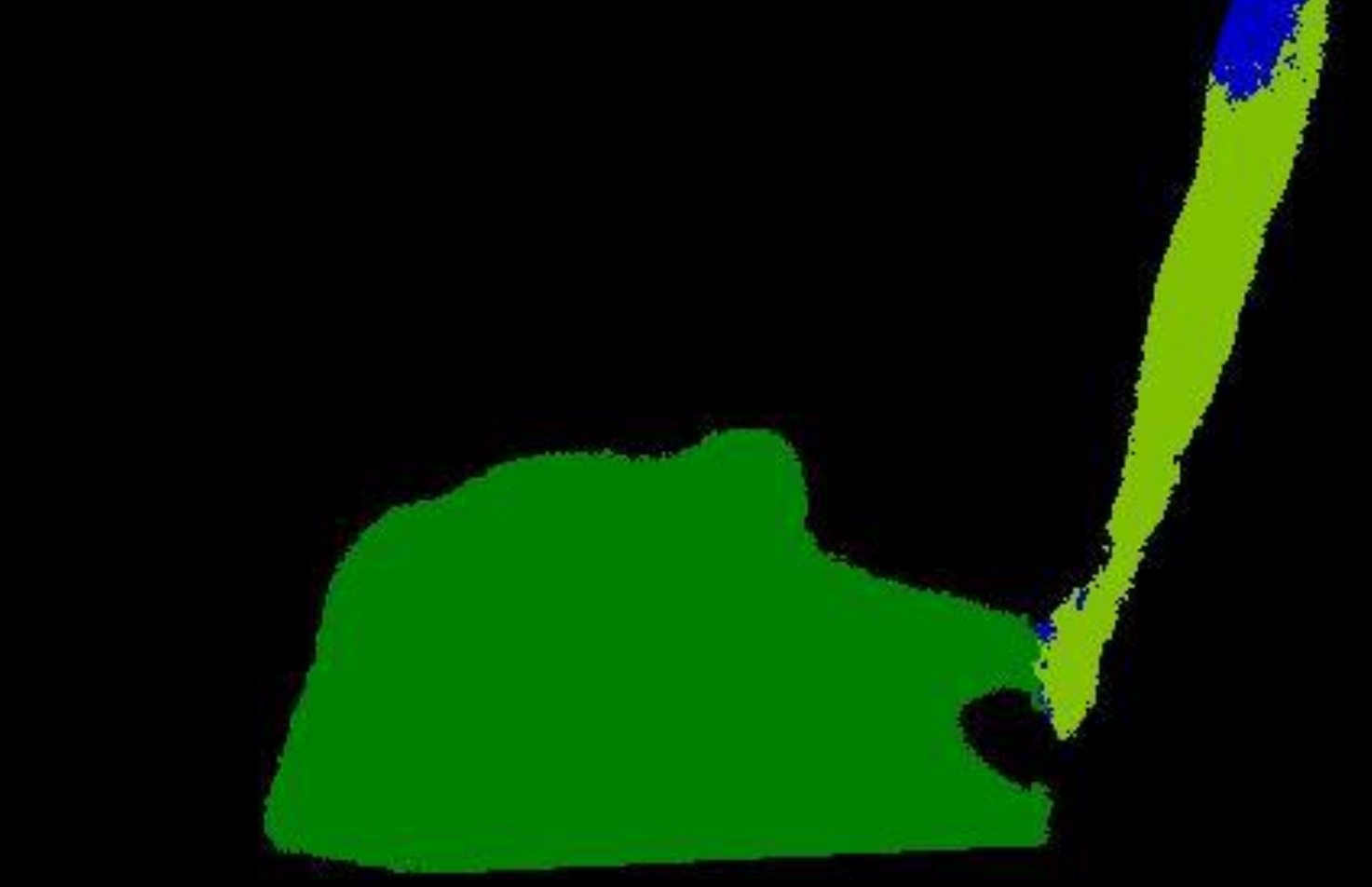}} &
  \parbox[c]{\imgwidth}{\includegraphics[width=\imgwidth]{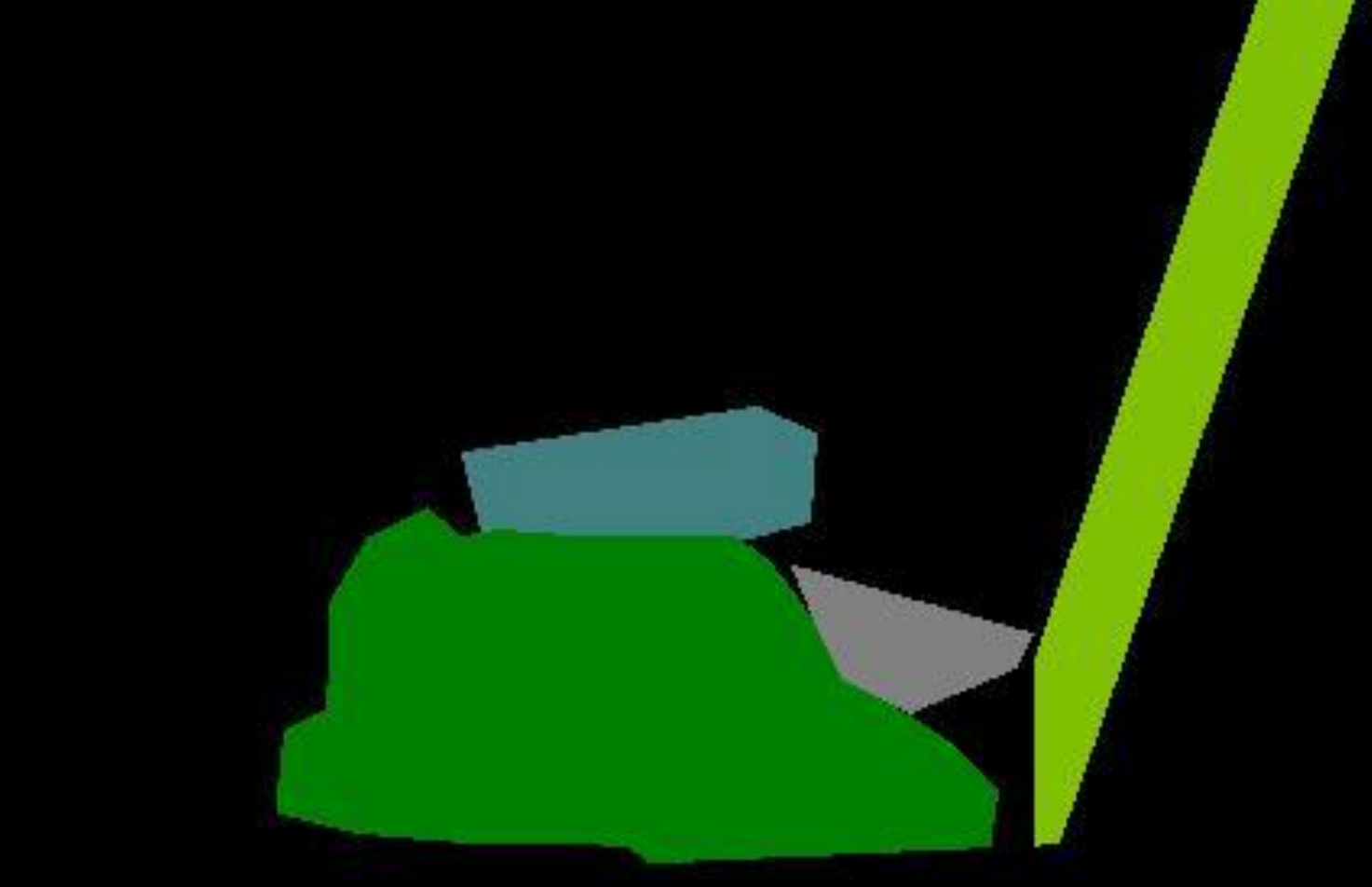}} \\
  scene 2 &
  \parbox[c]{\imgwidth}{\includegraphics[width=\imgwidth]{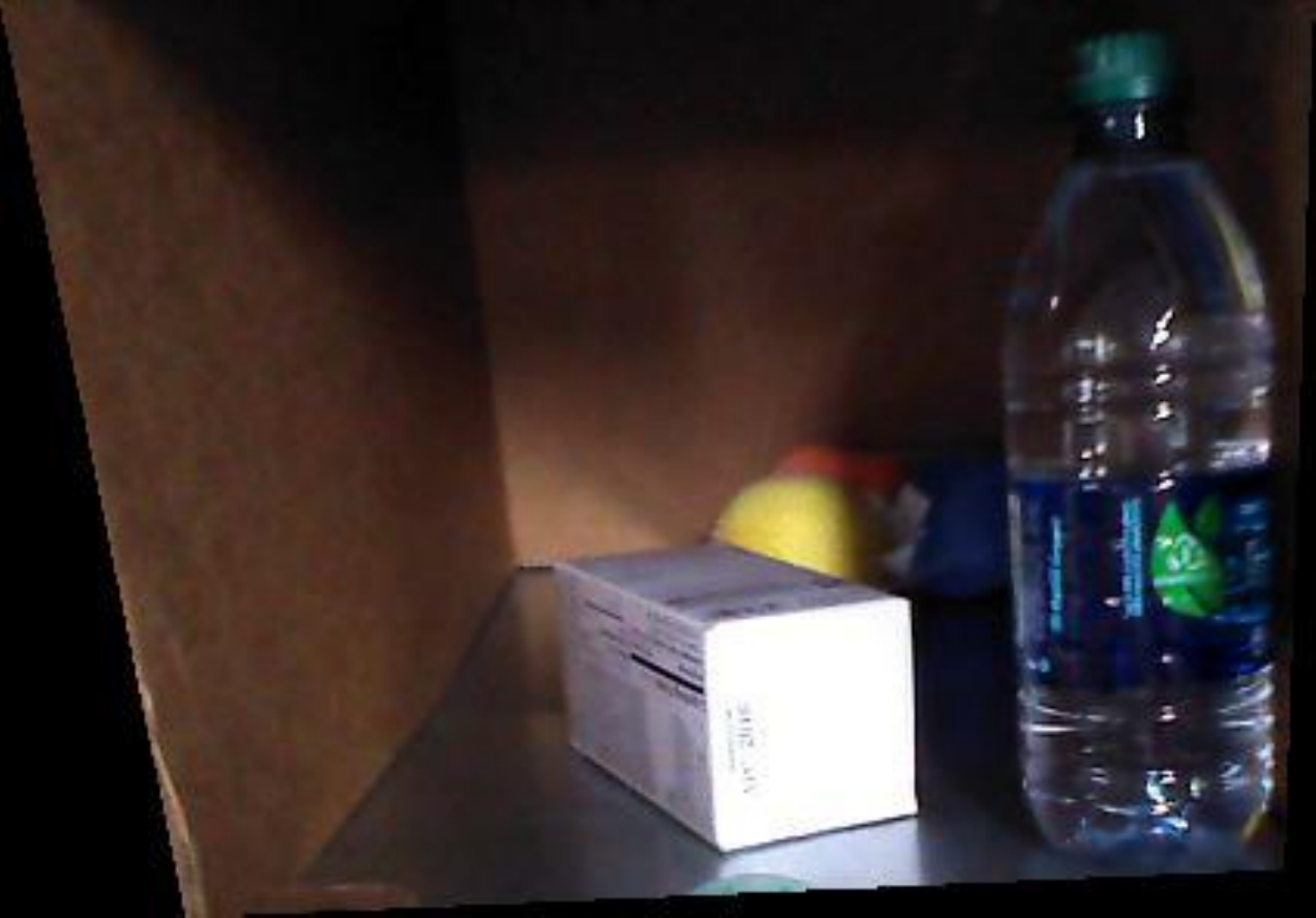}} &
  \parbox[c]{\imgwidth}{\includegraphics[width=\imgwidth]{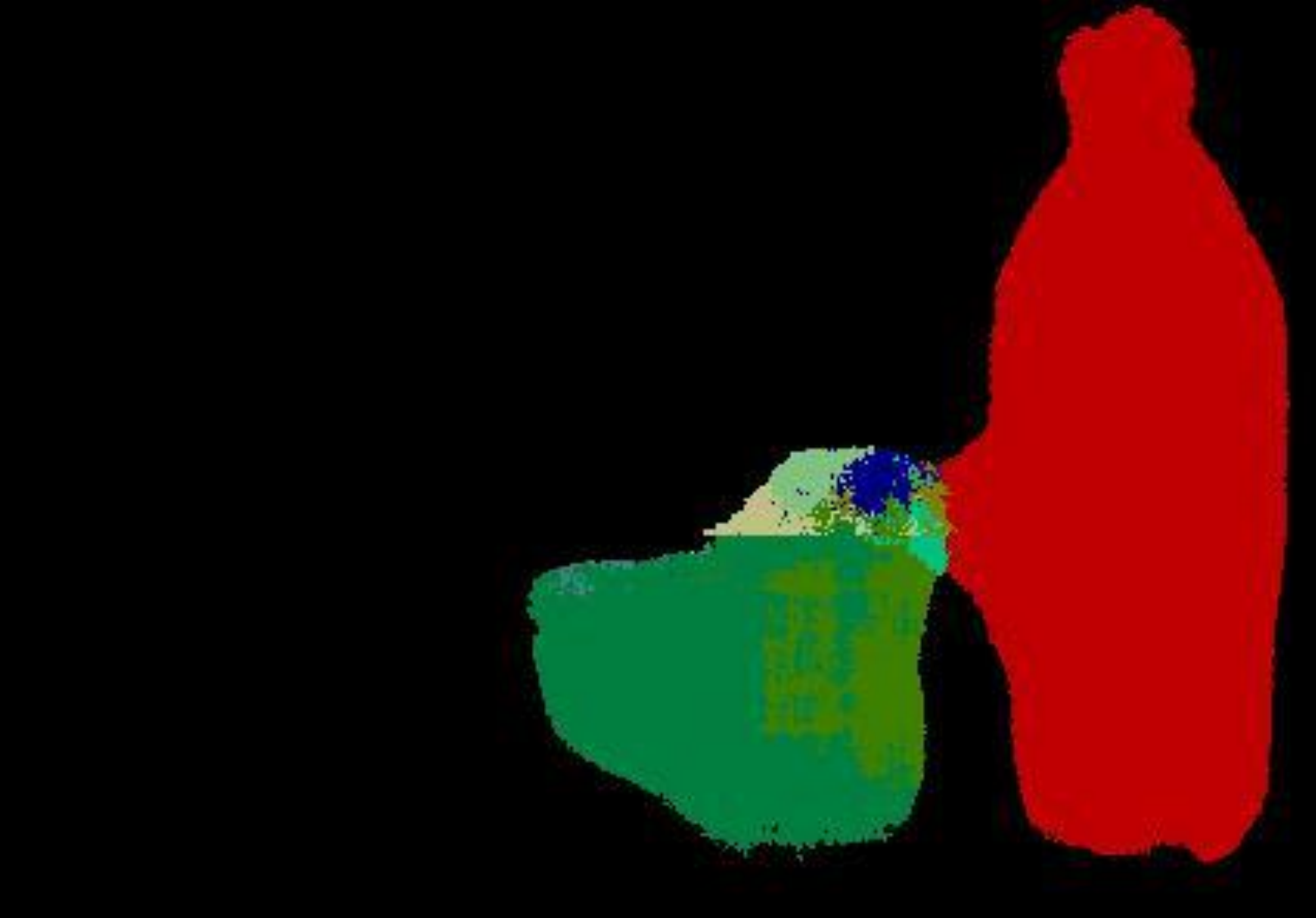}} &
  \parbox[c]{\imgwidth}{\includegraphics[width=\imgwidth]{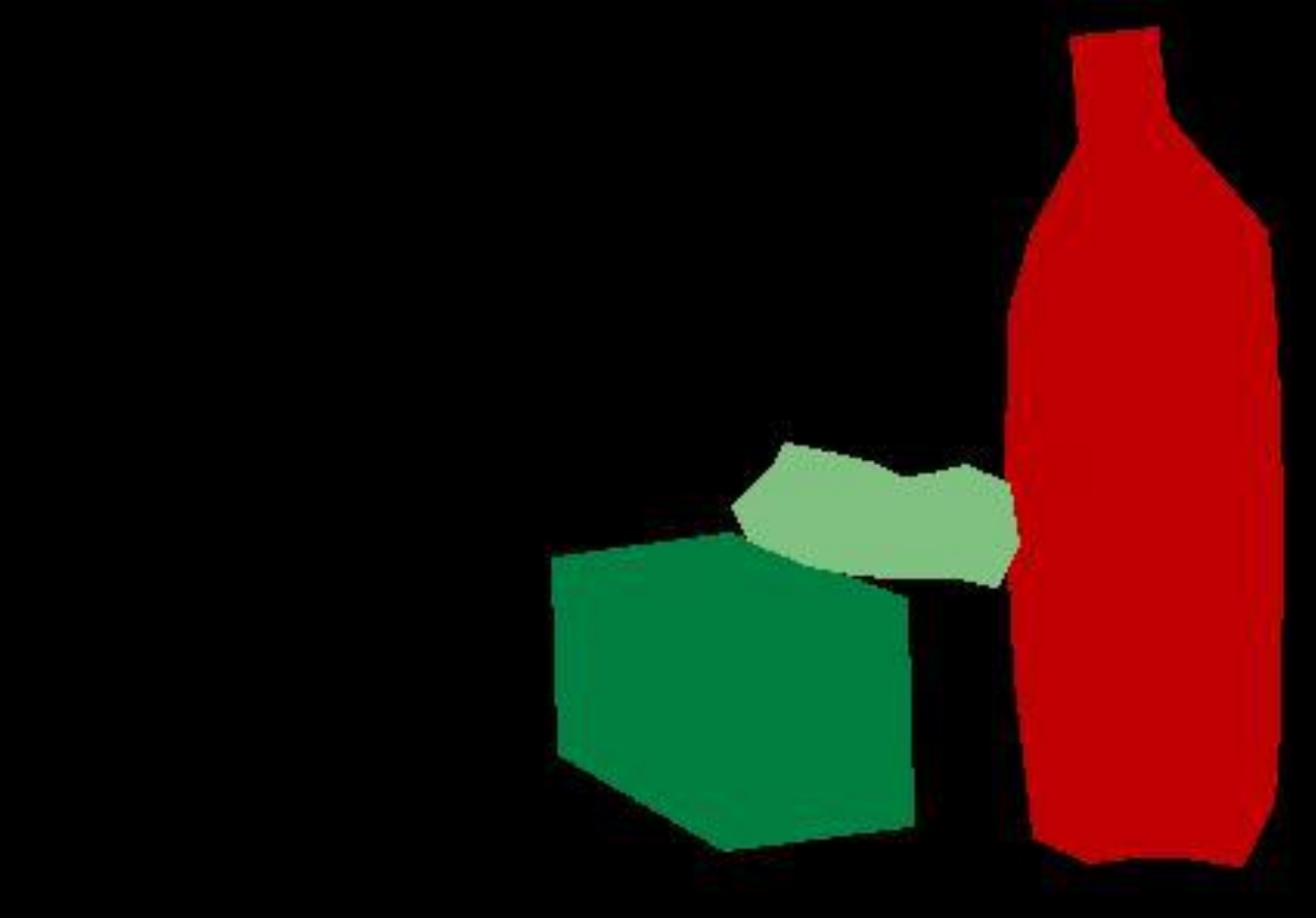}} \\
  scene 3 &
  \parbox[c]{\imgwidth}{\includegraphics[width=\imgwidth]{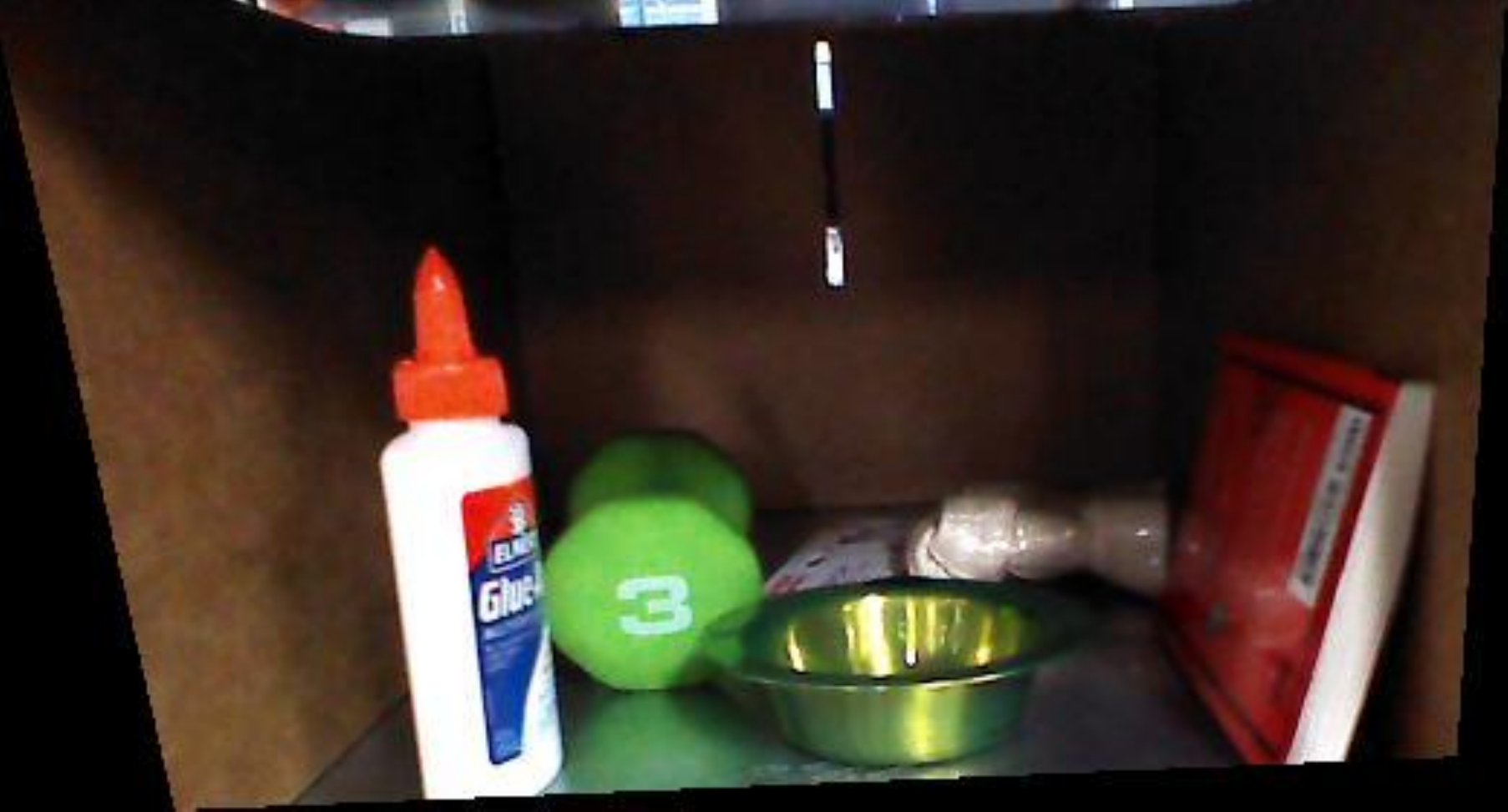}} &
  \parbox[c]{\imgwidth}{\includegraphics[width=\imgwidth]{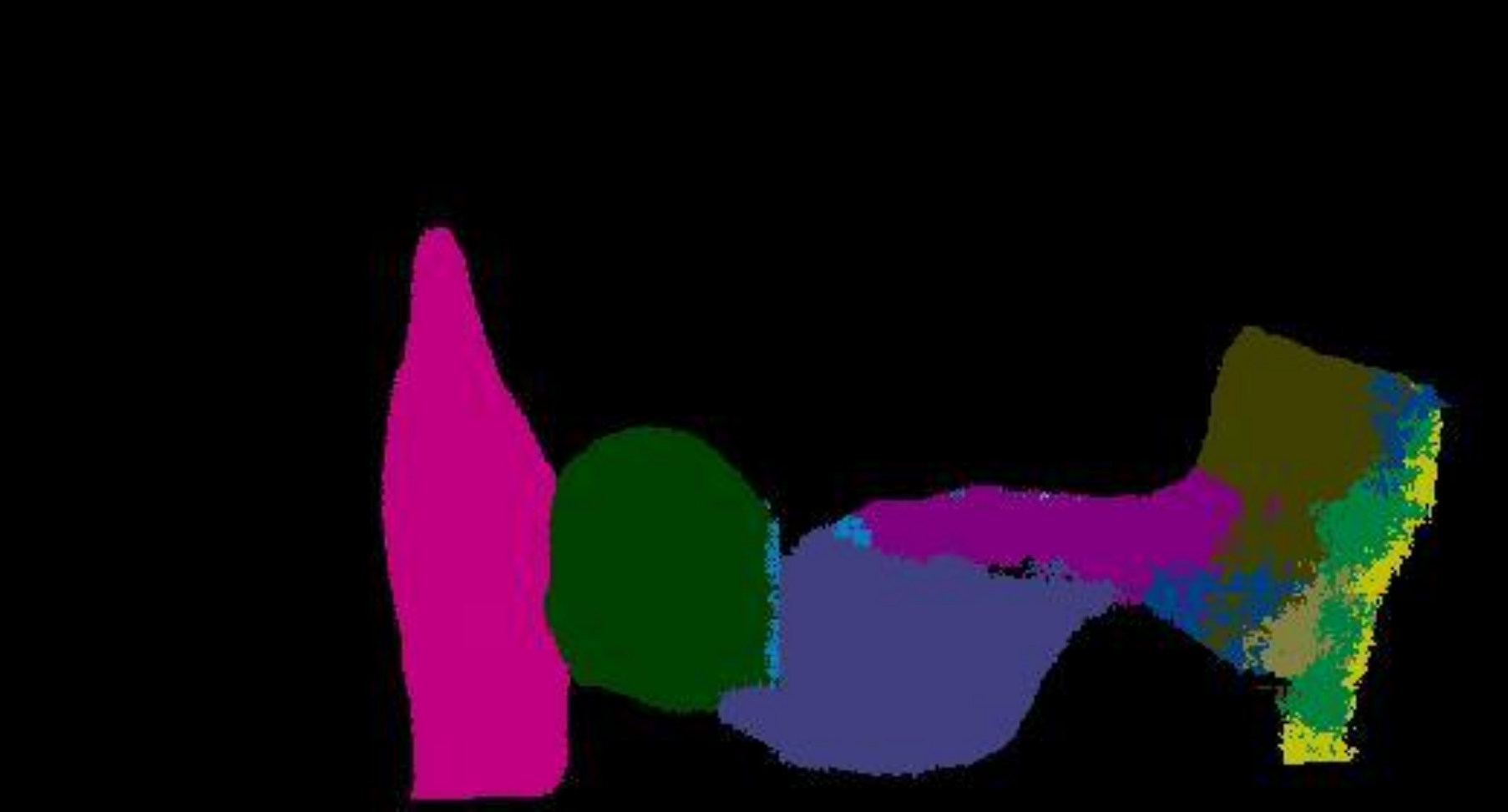}} &
  \parbox[c]{\imgwidth}{\includegraphics[width=\imgwidth]{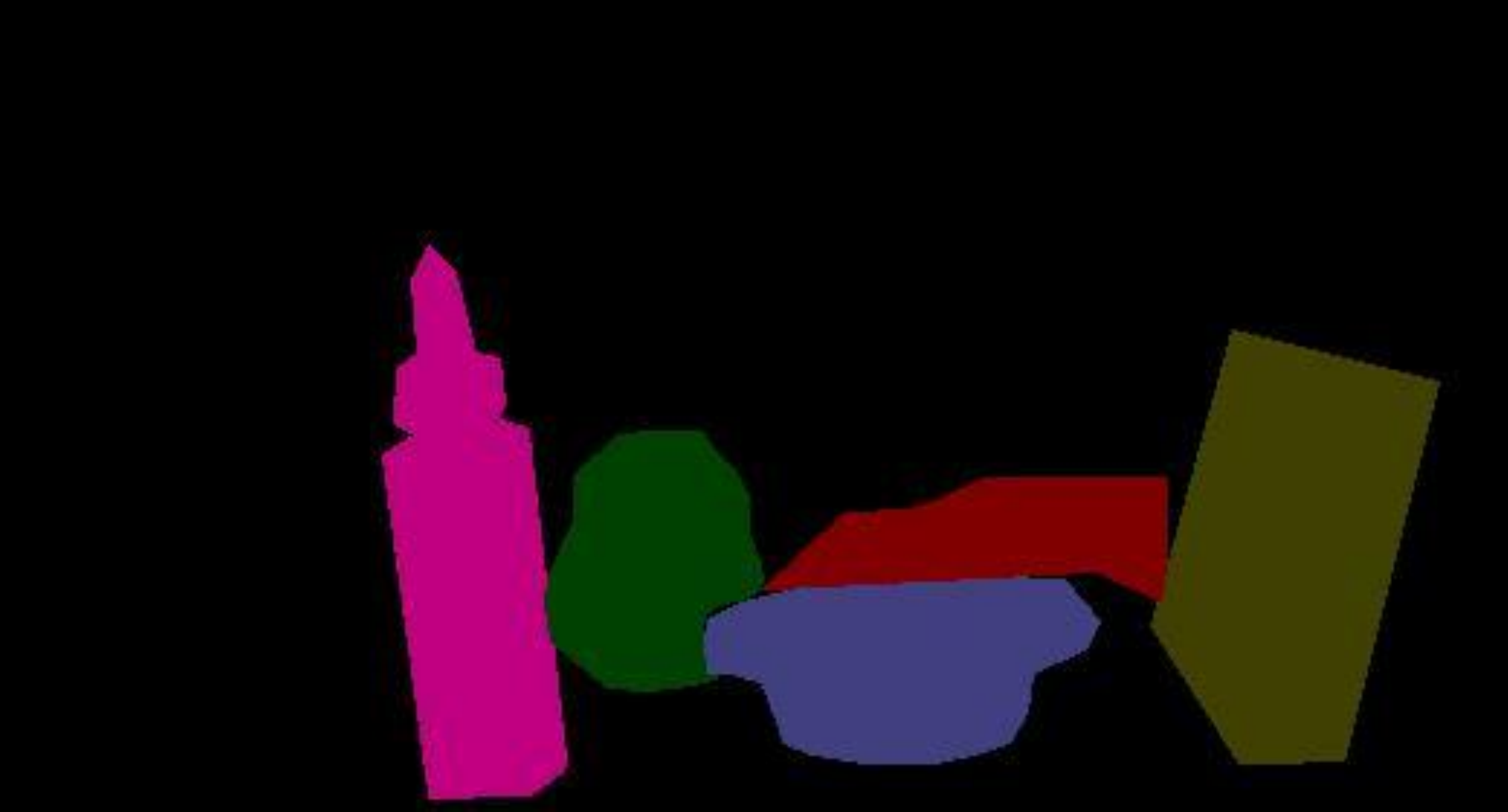}} \\
  \hline
  label colors &
  \multicolumn{3}{c}{\parbox[c]{0.75\columnwidth}{\includegraphics[width=0.75\columnwidth]{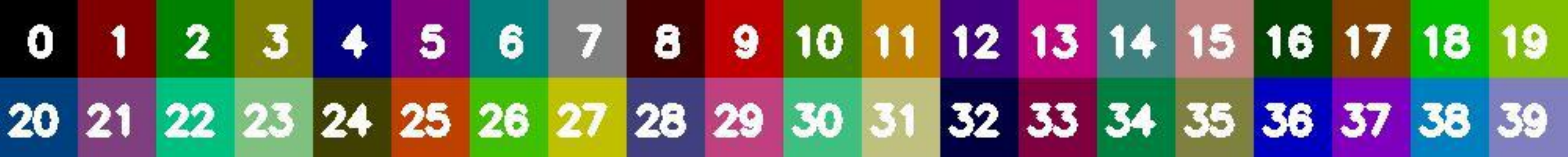}}}
  \end{tabular}
\end{table}

% -------------------------------------------------------------------------------------------------
\subsection{Evaluation of 3D Segmentation}

\begin{figure}[htbp]
  \centering
  \includegraphics[width=0.98\columnwidth]{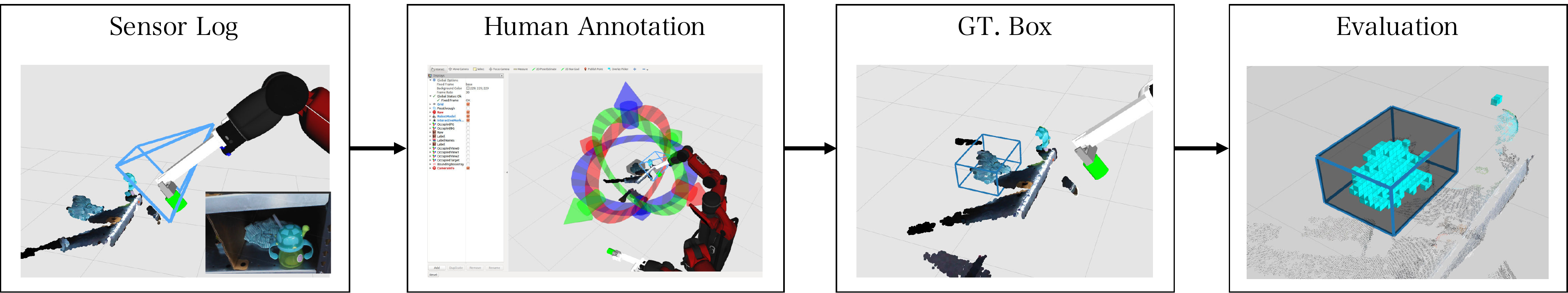}
  \caption{\textbf{Segmentation evaluation pipeline.}
    \small{
      The ground truth (gt.) box is annotated by human, and overlap between voxels and the box
      is computed as the accuracy.
    }
  }
  \figlab{segmentation_evaluation}
\end{figure}

\subsubsection{Ground Truth Annotation}
We evaluate our segmentation method with human annotation for ground truth
as shown in \figref{segmentation_evaluation}.
The sensor log, camera and robotic joint states,
are collected with rosbag \footnote{http://wiki.ros.org/rosbag},
and ground truth (gt.) box is annotated using interactive markers \cite{gossow2011interactive}
on rviz \footnote{http://wiki.ros.org/rviz}.

\subsubsection{Evaluation Metric}
The metric is the three-dimensional intersect-over-union $IU_{3d}$
between the set of generated object voxels $V_{pred}$
and the box $V_{gt}$ with \eqnref{iu}, where the volume of voxels and box is calculated
as $v_{pred} = {\cal V}(V_{pred}), v_{gt} = {\cal V}(V_{gt})$.
\begin{eqnarray} \eqnlab{iu}
  V_{tp} = V_{pred} \cap V_{gt}, \quad v_{tp} = {\cal V}(V_{tp}) \nonumber \\
  v_{fp} = v_{pred} - v_{tp}, \quad v_{fn} = v_{gt} - v_{tp} \nonumber\\
  IU_{3d} = \frac{v_{tp}}{v_{tp} + v_{fp} + v_{fn}}  \qquad
\end{eqnarray}
This metric $IU_{3d}$ has value in range $[0, 1]$,
so we show value multiplied with 100 in following.

\begin{figure}[htbp]
  \centering
  \subfloat[View 1]{
    \includegraphics[width=0.3\columnwidth]{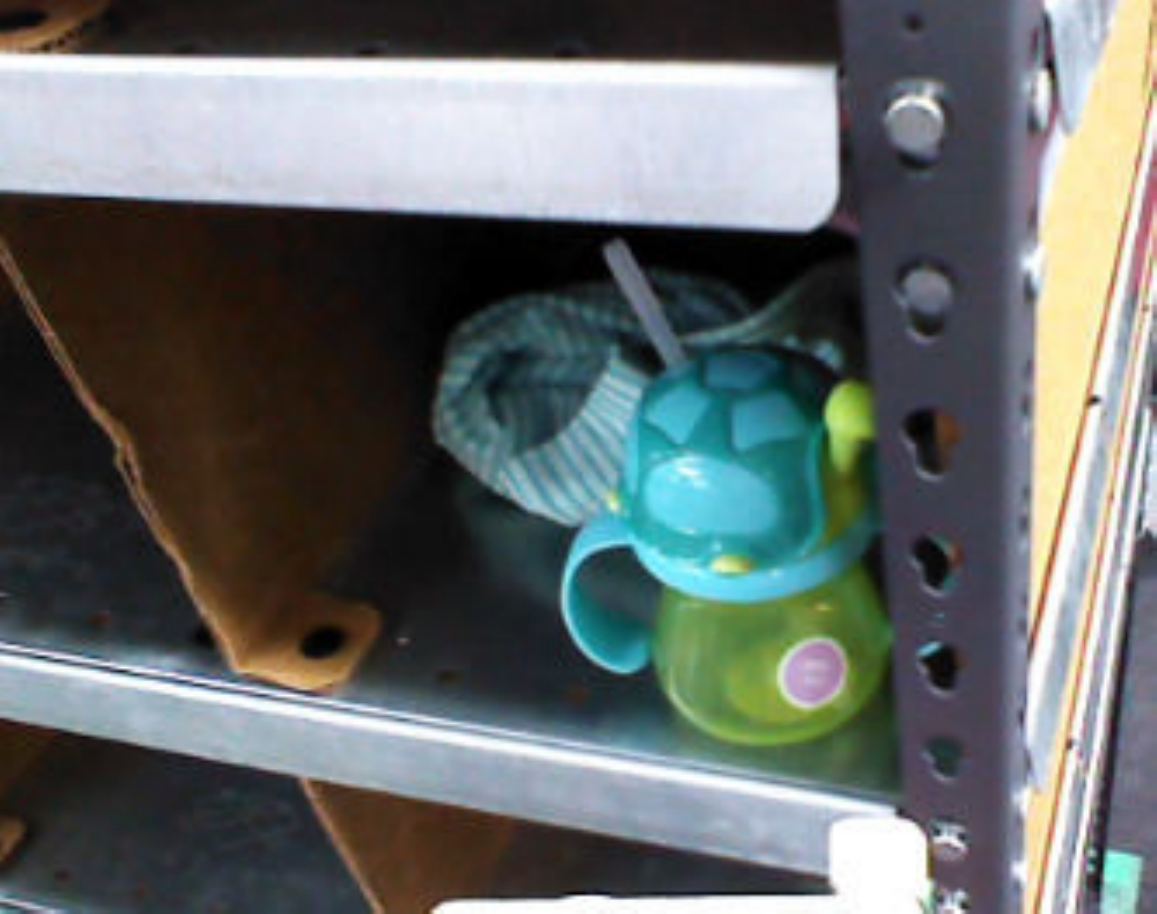}
    \figlab{eval_view1}
  }
  \subfloat[View 2]{
    \includegraphics[width=0.3\columnwidth]{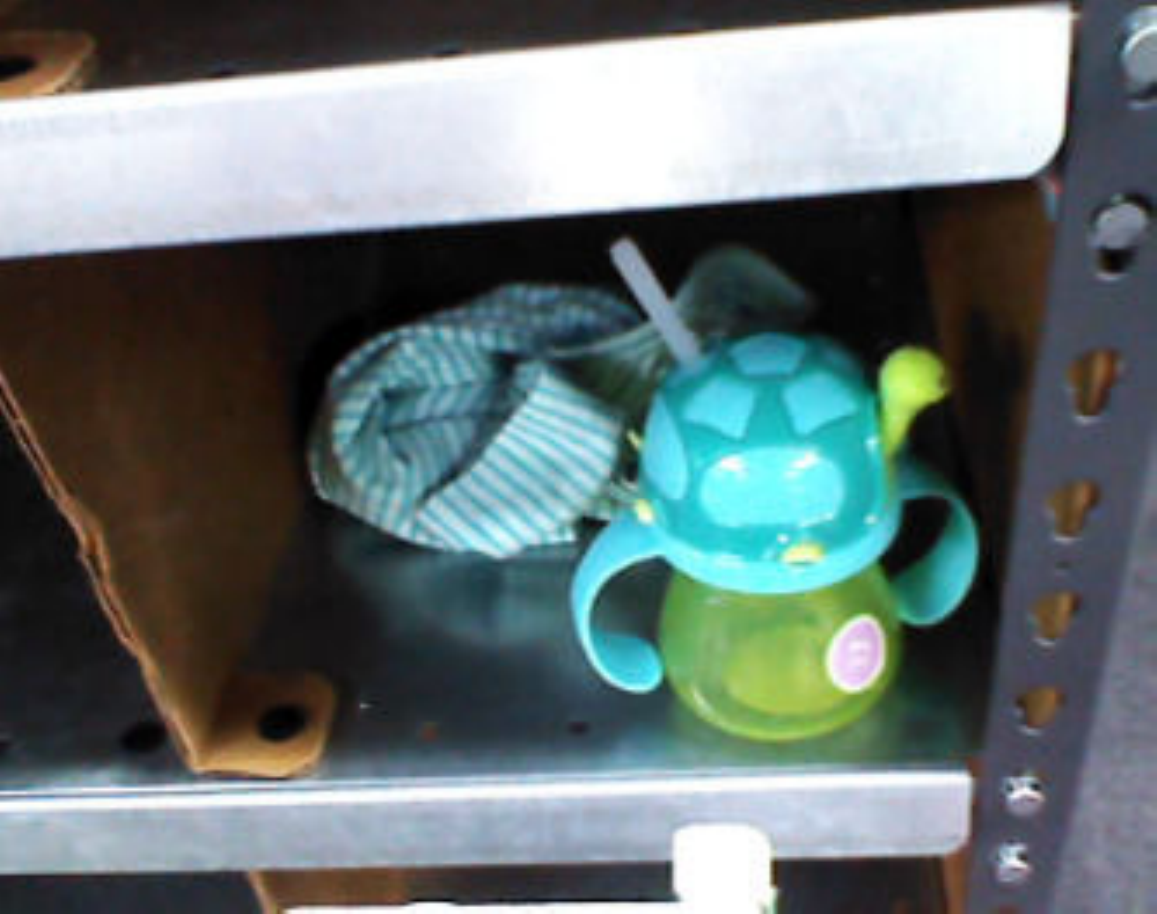}
    \figlab{eval_view2}
  }
  \subfloat[View 3]{
    \includegraphics[width=0.3\columnwidth]{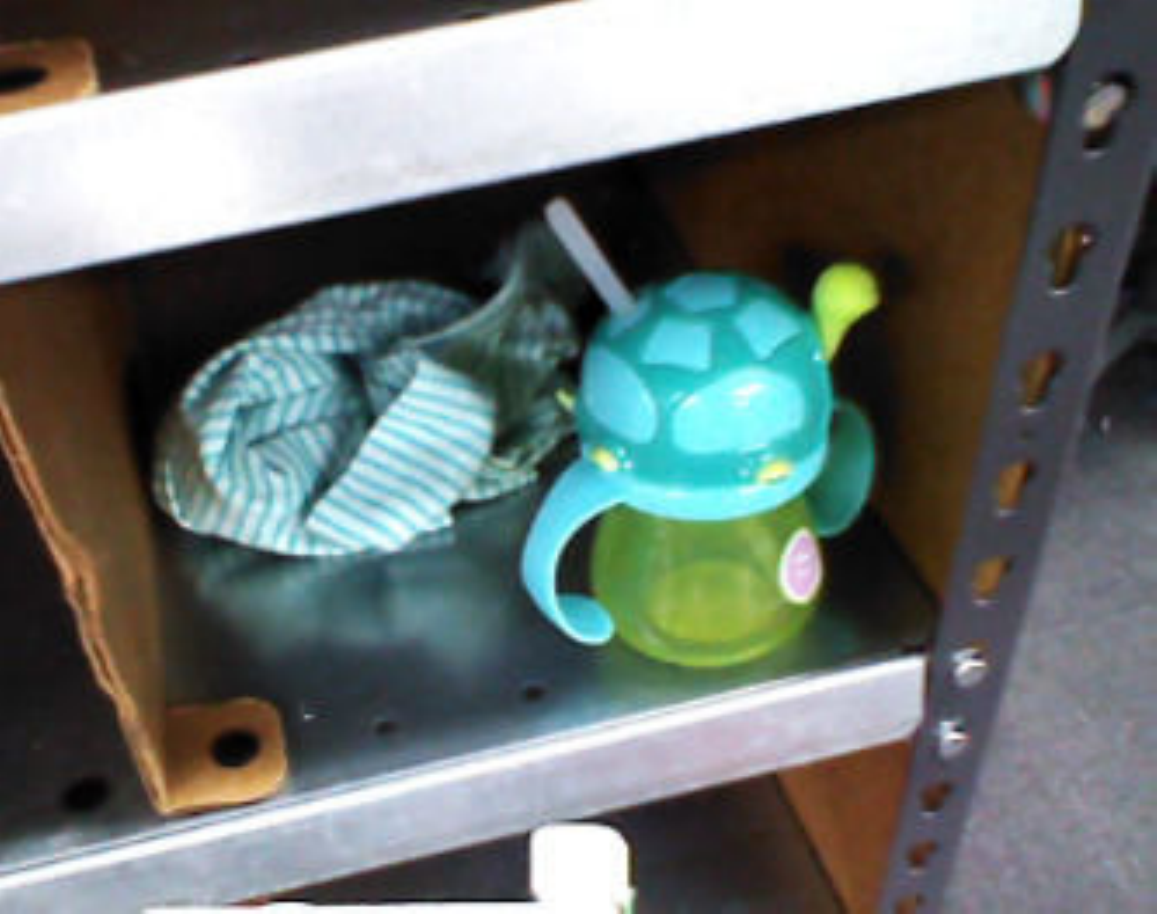}
    \figlab{eval_view3}
  }
  \caption{\textbf{Three views used in the evaluation.}}
  \figlab{three_views}
\end{figure}

\subsubsection{Result}
We generate a looking-around motion with 3 different views shown in \figref{three_views},
and the evaluation is conducted in situations with occlusions by object itself and the others.
The result of projection-based method for both mean and max in views,
is collected for each view and that based on LabelOctoMap is done after the views.
\tabref{iu_result} shows the evaluation result with all 39 objects shown in \figref{objects},
and it shows our method (LabelOctoMap) exceeds conventional method (Projection)
for 29 objects, and is close for 4 (label 7,14,27,34).
The mapping result is worse than projection for 2 objects (25,26),
which are both toothbrush with color difference.
From this result, it is conceivable that our method is not efficient
to segment objects which looks very similar and can have competitive probabilities,
because the probabilities become relatively lower in that case than that when segmenting other objects.
% because probabilities of them can be competitive and relatively lower value than
% segmenting other object classes.
Results by both methods are similarly bad for 3 objects (5,30,31),
and it can be said that this is mainly caused by lower accuracy of 2D segmentation (5,31)
and insensible depth information of the objects because of its holes in the surface (30).
These results are summarized in \tabref{segmentation_evaluation_result},
and accuracy with our method exceeds that with projection-based method for both mean and max in views.
These results show our mapping-based method is efficient to segment object three-dimensionally
compared to the conventional projection-based method,
and the validity of our proposed octomap which segments multiple objects in a multi-view action.

% mean IU for single view:  0.0468772293968
% mean of max IUs of three single views:  0.0726847850138
% mean IU for LabelOctoMap:  0.0831144640966
% ratio of LabelOctoMap exceeds max IU of three single views:  0.619047619048
% ratio of LabelOctoMap exceeds mean IU of three single views:  0.952380952381
\definecolor{Gray}{gray}{0.85}
\begin{table}[htbp]
  \centering
  \caption{\textbf{Summary of segmentation results.}
    \small{
      They are the average of results ($IU_{3d}$) in \tabref{iu_result}.
    }
  }
  \tablab{segmentation_evaluation_result}
  \begin{tabular}{l|rr}
              methods  & mean $IU_{3d}$      \\
  \hline
  Projection (Mean) & 6.40           \\
  Projection (Max)  & 8.98           \\
  \rowcolor{Gray}
  LabelOctoMap        & \textbf{12.57} \\
  \end{tabular}
\end{table}

% -------------------------------------------------------------------------------------------------
%
%
% -------------------------------------------------------------------------------------------------
\subsection{Multi-Object Manipulation with Segmentation using LabelOctoMap}

We evaluated real-time validity and applicability of our method
in a multi-object manipulation task, which have the similar configuration as APC2016.

\subsubsection{Task Configuration}

The manipulation task is conducted in a workspace shown in \figref{workspace},
and the information below is given to the robot:
\begin{itemize}
  \item Location and model of the shelf.
  \item Object set (\figref{objects}).
  \item Object names located each shelf bin.
  \item Name of the target object in each bin.
\end{itemize}
The purpose of the task is to pick the target object from the shelf bin and place it to the tote,
so robot does not have to manipulate the non-target objects.
However, since we located non-target objects with occluding the target object in all scenes,
it is hard to pick the target without manipulating the obstacle objects.
It is also allowed for the robot to move objects from a shelf bin to another,
but information of the bin to which the object is moved must be noticed by the robot,
because the information will be used another attempt of the task for the bin.

\begin{figure}[htbp]
  \centering
  \includegraphics[width=0.9\columnwidth]{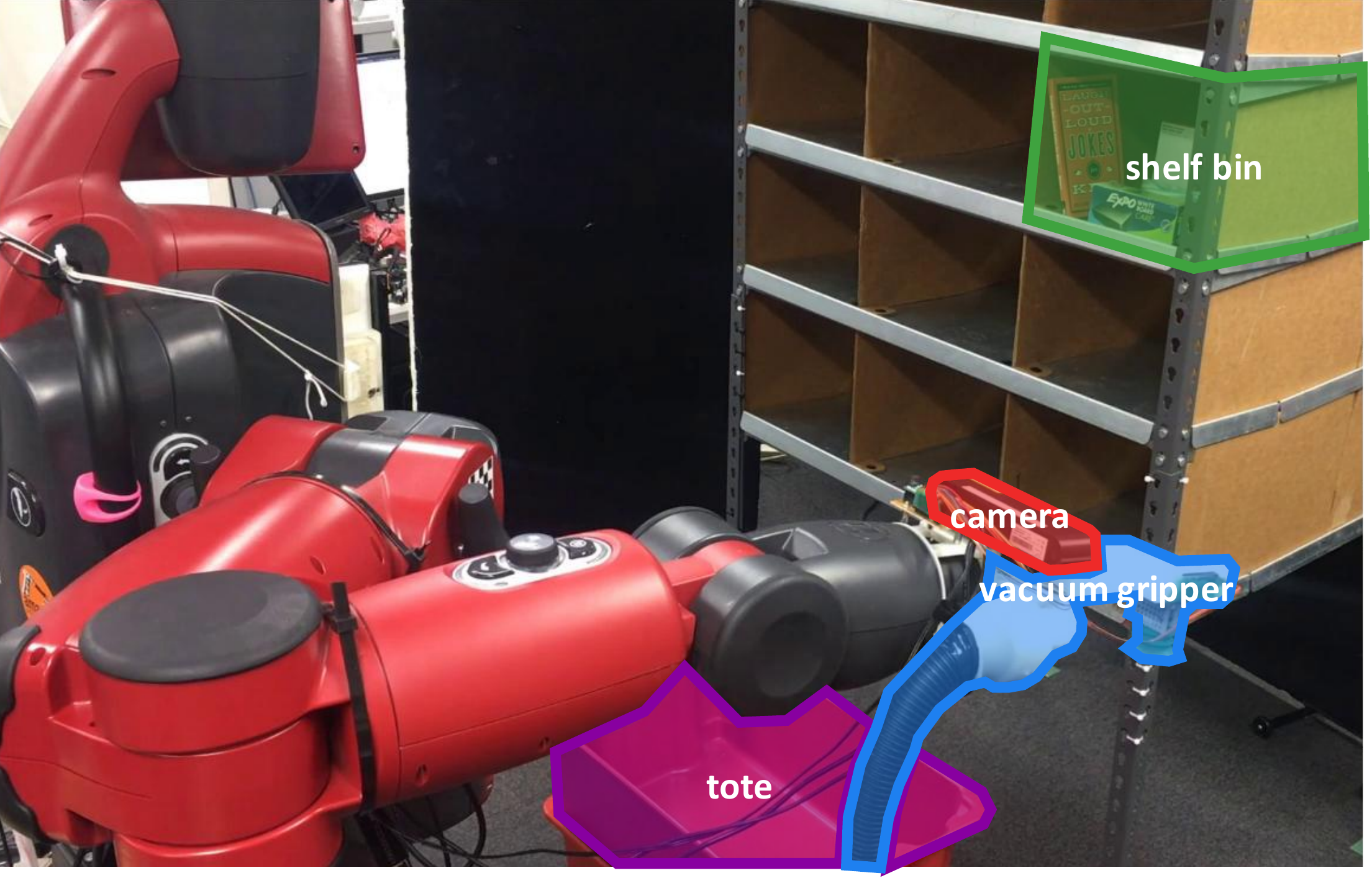}
  \caption{\textbf{Workspace of the manipulation task.}
    \small{The location of each shelf bin and tote are given to the robot.}
  }
  \figlab{workspace}
\end{figure}

\subsubsection{Picking System Configuration}
% Intro
The overview of the picking system is shown in \figref{picking_system}.
In detail, the task is conducted as follows:
\begin{enumerate}
  \item Generate multilabel object voxels by LabelOctoMap with multi-view action.
  \item Detect non-target objects in front of the target:
        centroids of each object-class voxels are computed, and robot decide to pick non-target object
        if its centroid is in front of the target object's.
  % \item Select what object-class to grasp: if non-target object box is detected in front of
  %       the target object box, robot decise to remove it, and target object box is selected if else.
  \item Remove outlier from voxels by euclidean clustering and fit bounding box.
  \item Plan how to grasp: position is decided from centroid of the voxels, and
        orientation is decided from location and dimension of the box.
  \item Remove or place the grasped object: non-target object is placed into empty bins,
        and target object is placed into the tote.
\end{enumerate}
This task sequence is executed until the robot picks the target object
and places into the designated place.
In this manipulation scenario,
it is assumed the location of the shelf bin and the name of target object are given.

\begin{figure}[htbp]
  \centering
  \includegraphics[width=0.98\columnwidth]{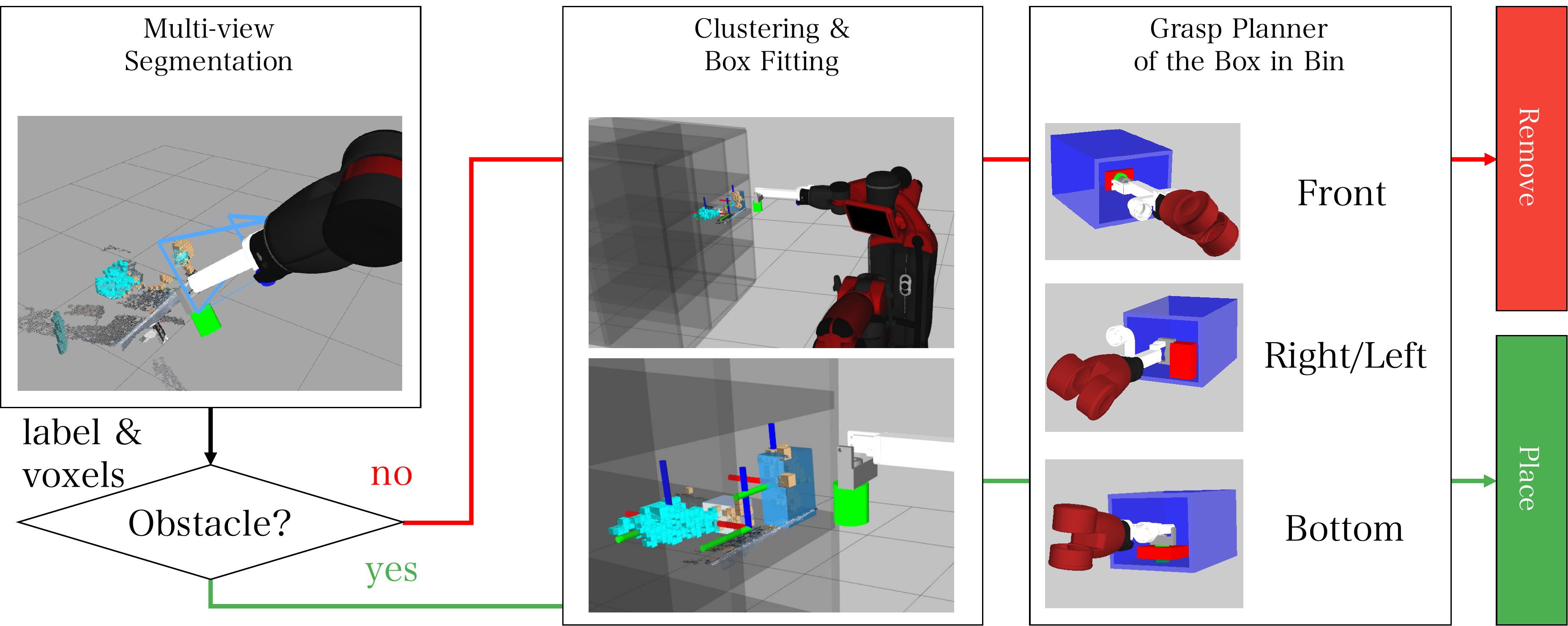}
  \caption{\textbf{Picking system.}
    \small{
      The picking system includes multi-object manipulation, in order to
      remove obstacle objects to pick a target object.
    }
  }
  \figlab{picking_system}
\end{figure}

% % Task Sequence
% The task is conducted as follows:
% firstly robot looks around the target bin to generate LabelOctoMap for objects in the bin,
% secondly picks objects that locates in front of the target object,
% thirdly place object into other shelf bin if picked object is not the target
% and in other cases into order bin.
% This task sequence is executed until the robot picks the target object
% and places into the order bin.
% With the looking around motion with RGB-D sensor on hand
% the label occupancy voxel map is generated,
% and the picking motion of the robot is generated based on the fitted box and centroid estimation
% with the generated map.

\subsubsection{Result}
\figref{pick_task_sequence} shows the sequential frames of task in environment with 1 target and 3 obstacle objects,
where the robot needs to remove the obstacle objects in order to pick the target object which is occluded.
The robot successfully moved the obstacle objects to other shelf bin,
and picked the target object and placed into the order bin.

\newcommand{\expimgscale}{0.62\columnwidth}
\begin{figure*}[t]
  \centering
  \subfloat[Look around]{
    \includegraphics[width=\expimgscale]{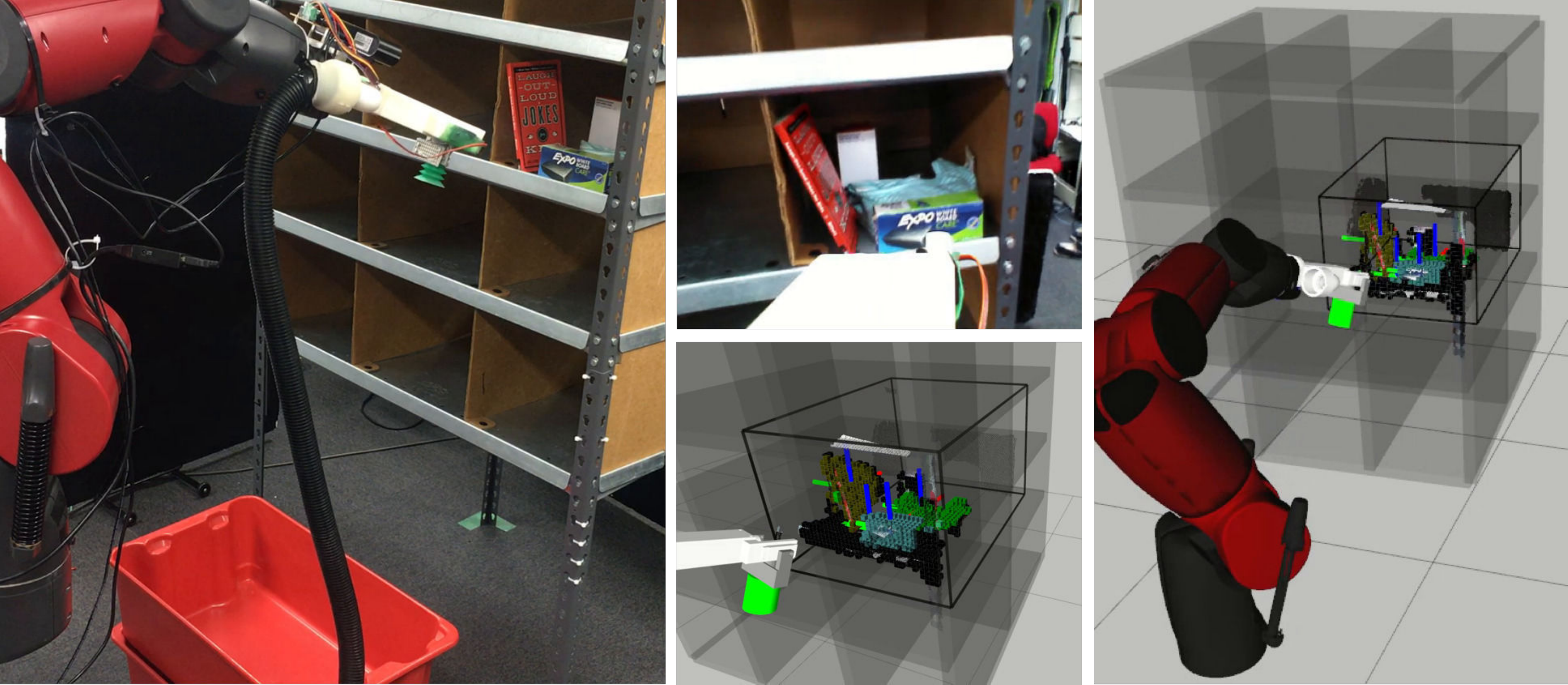}
    \figlab{exp00001}}
  \subfloat[Pick non target objects]{
    \includegraphics[width=\expimgscale]{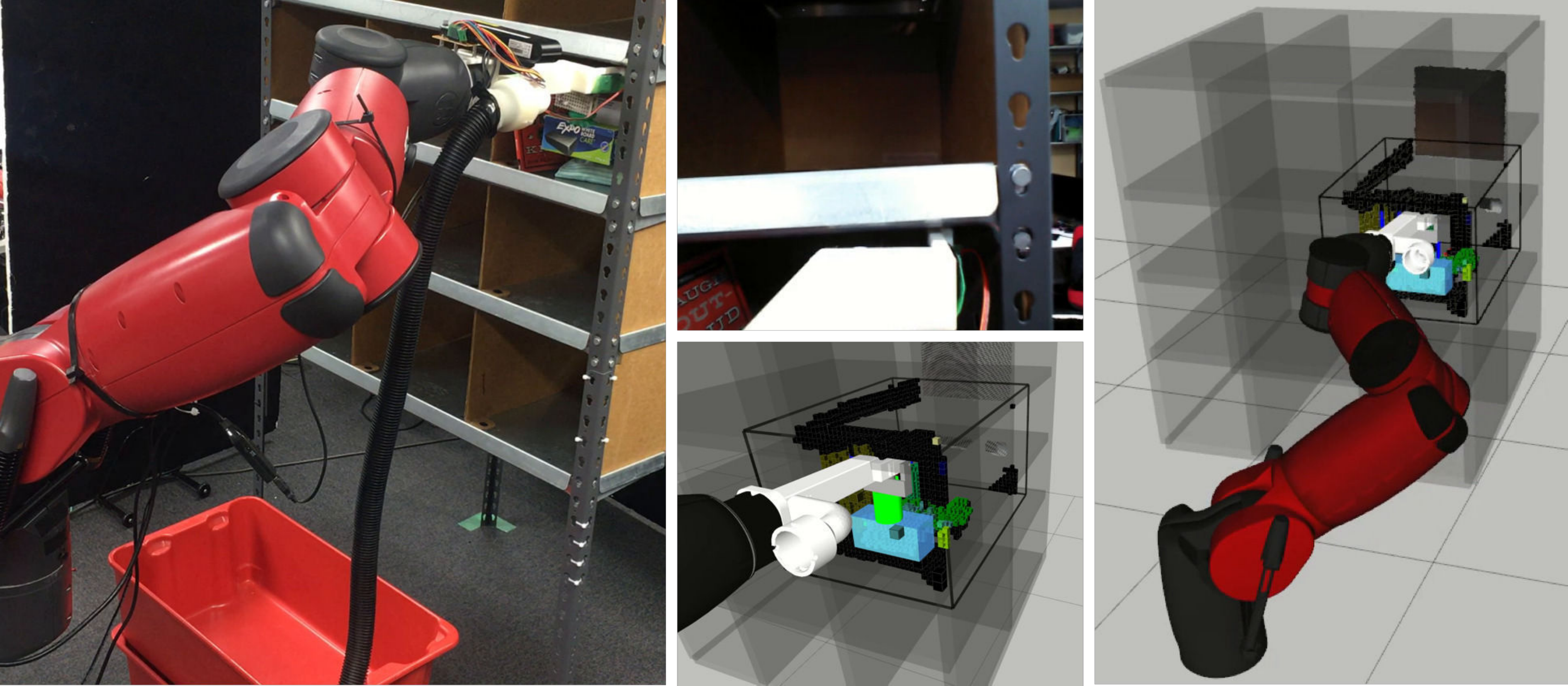}
    \figlab{exp00002}}
  \subfloat[Remove non target objects]{
    \includegraphics[width=\expimgscale]{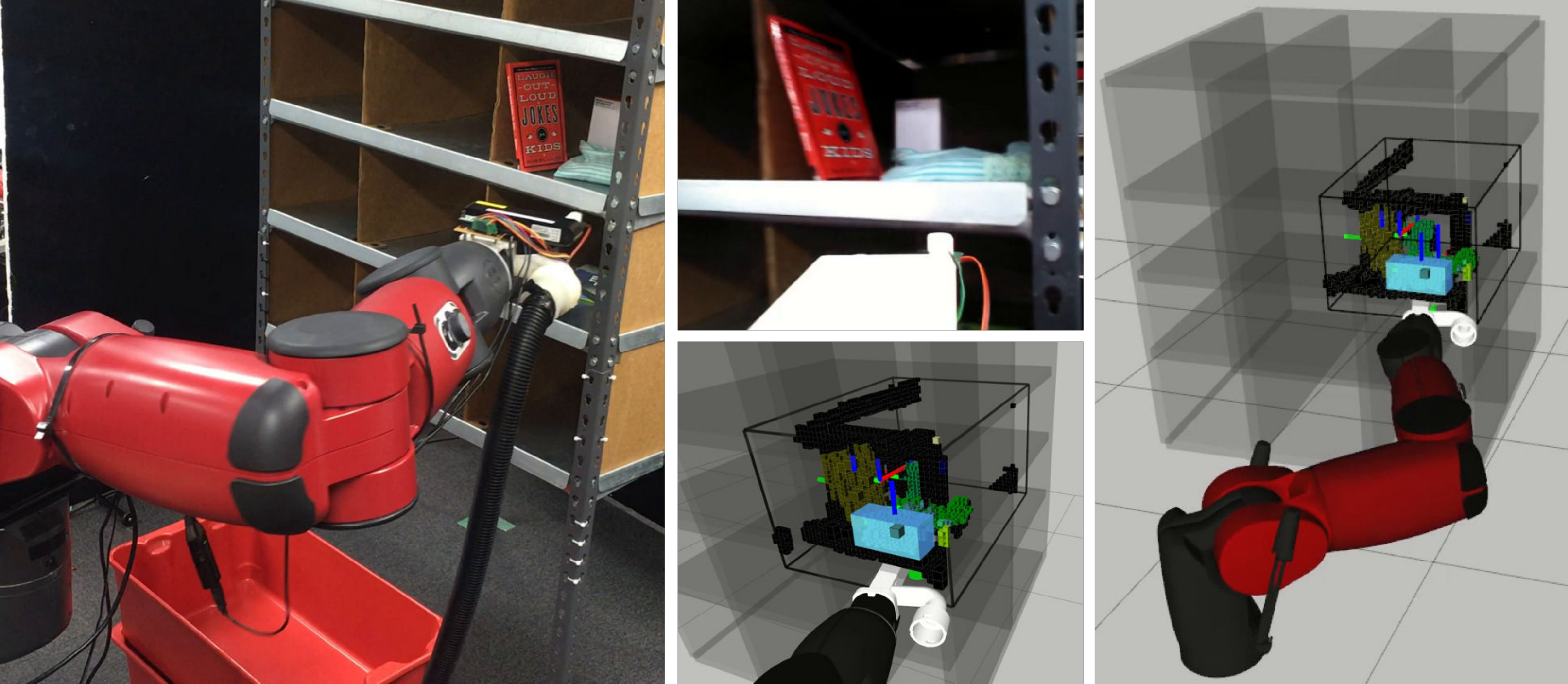}
    \figlab{exp00003}}

  \subfloat[Pick the target object]{
    \includegraphics[width=\expimgscale]{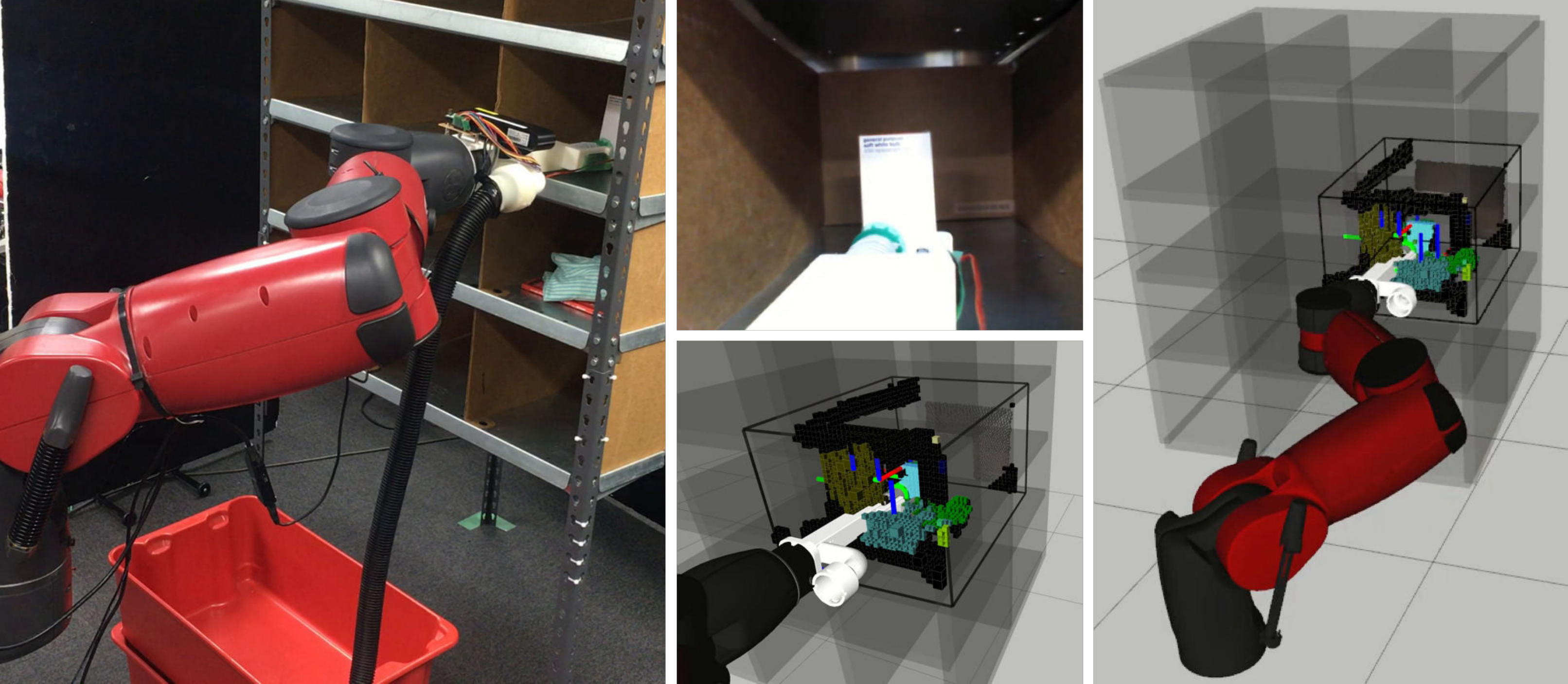}
    \figlab{exp00004}}
  \subfloat[Place the target object]{
    \includegraphics[width=\expimgscale]{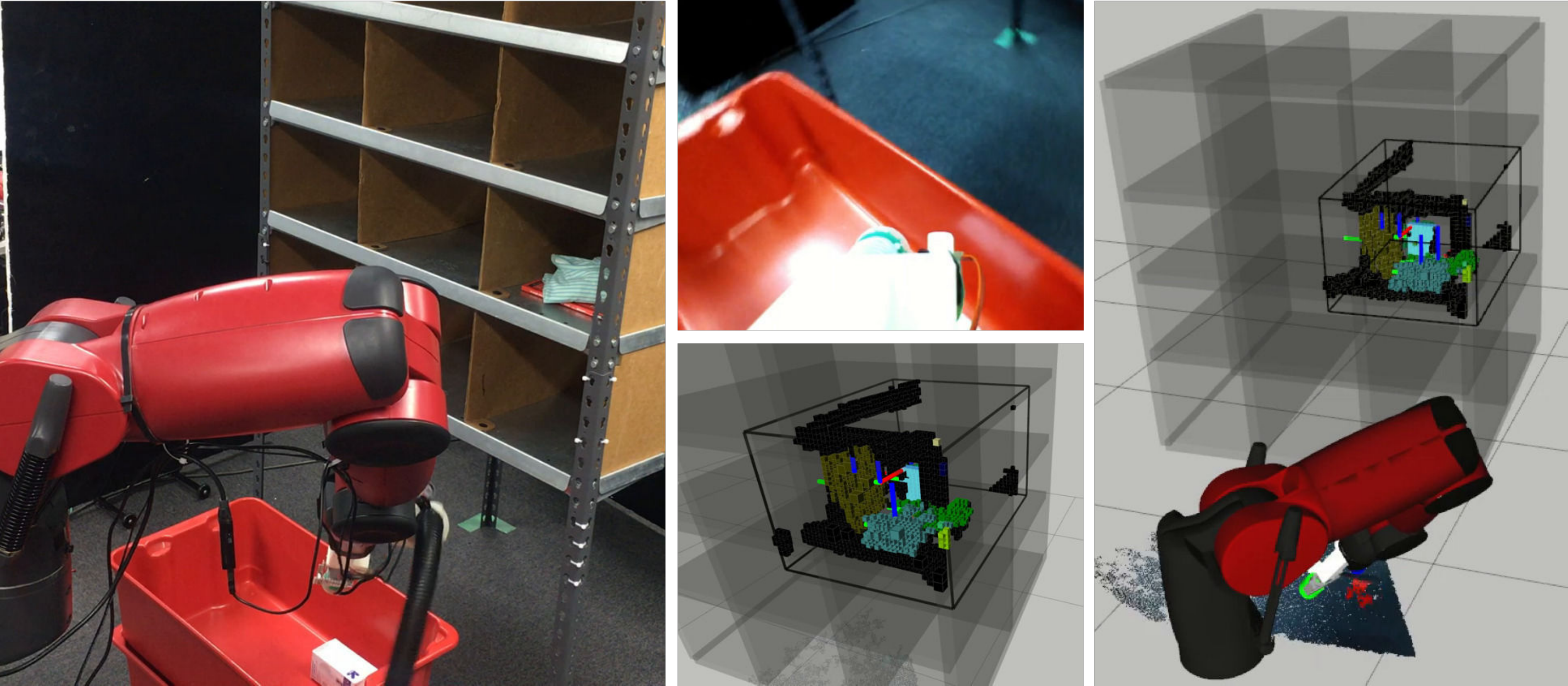}
    \figlab{exp00005}}
  \subfloat[Result]{
    \includegraphics[width=\expimgscale]{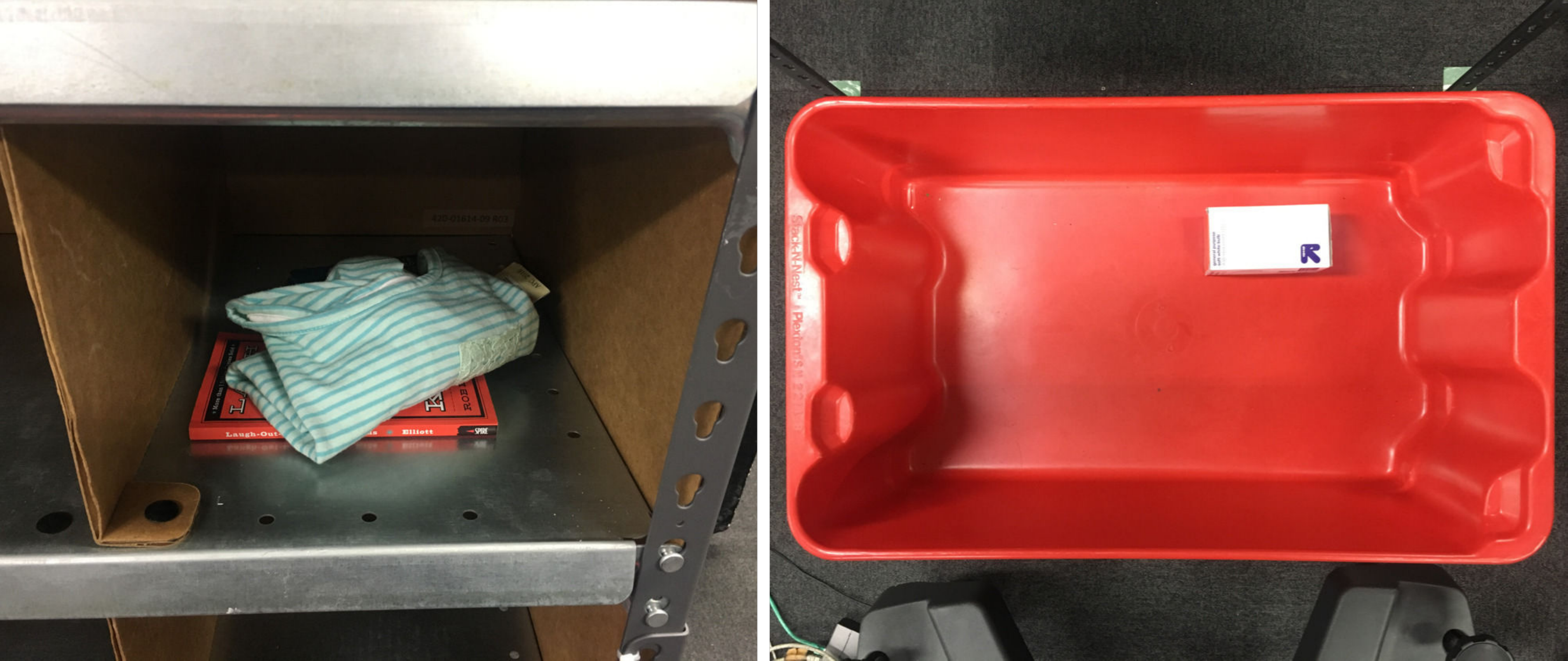}
    \figlab{exp00006}}
    \caption{\textbf{Sequential frames of the picking task.}
    \small{
      The frame shows the state of real robot (left), hand camera image (upper center),
      and visualization (lower center, right)
      in which colored voxels are the segmented objects and the axes represent the centroids.
    }
  }
  \figlab{pick_task_sequence}
\end{figure*}

We evaluated the efficiency of the picking system with proposed method
by testing in 61 times picking task execution with 1 obstacle and 1 target object.
The combination of 2 objects is randomly selected from 15 objects
(2,4,6,7,10,14,22,24,27,28,34-36,38,39), which can be grasped relatively easily
by the robot with a vacuum gripper.
The environment is set to put an obstacle object in front of the target object.
\tabref{picking_task_success_rate} shows the success rate, and
\tabref{recog_and_task_results} shows some examples with scene image and task result:
\textbf{recognize} is finding target and non-target,
\textbf{remove} is moving non-target, and \textbf{pick} is picking target.
The failure cases of \textbf{remove} and \textbf{pick} happen with gripper air leakage caused
by inappropriate grasp location,
and that of \textbf{recognize} happen with mistakenly detected non-target objects.
In the experiment, the task,
which includes multi-object manipulation in environment with occlusions,
is achieved (in $\sim$60\%), and this shows the applicability of our 3D segmentation method.

\begin{table}[htbp]
  \centering
  \caption{\textbf{Success rate of the picking task.}}
  \tablab{picking_task_success_rate}
  \begin{tabular}{l|rr}
    {}          & success rate   & success rate (total) \\
    \hline
    recognize   & 86.9\% (53/61) & 86.9\% (53/61)       \\
    remove      & 88.7\% (47/53) & 77.0\% (47/61)       \\
    pick        & 78.7\% (37/47) & 60.7\% (37/61)       \\
  \end{tabular}
\end{table}

\renewcommand{\imgwidth}{2.0cm}
\newcommand{\rotateclock}[1]{\rotatebox[origin=c]{90}{#1}}
\newcommand{\tabimg}[1]{\parbox[c]{\imgwidth}{\includegraphics[width=\imgwidth]{#1}}}
\newcommand{\tabimgnp}{\parbox[c]{\imgwidth}{\includegraphics[width=2.0cm,height=1.3cm]{nowprinting}}}
\begin{table}[htbp]
  \centering
  \caption{\textbf{Recognition and task results.}
    \small{It shows scenes with different task results: recognize, remove, and pick.}
  }
  \tablab{recog_and_task_results}
  \begin{tabular}{l|cc|ccc}
    {}                    & scene                                & segmentation                         & \thead{recog- \\ nize} & remove      & pick~        \\
    \hline
    \rotateclock{scene 1} & \tabimg{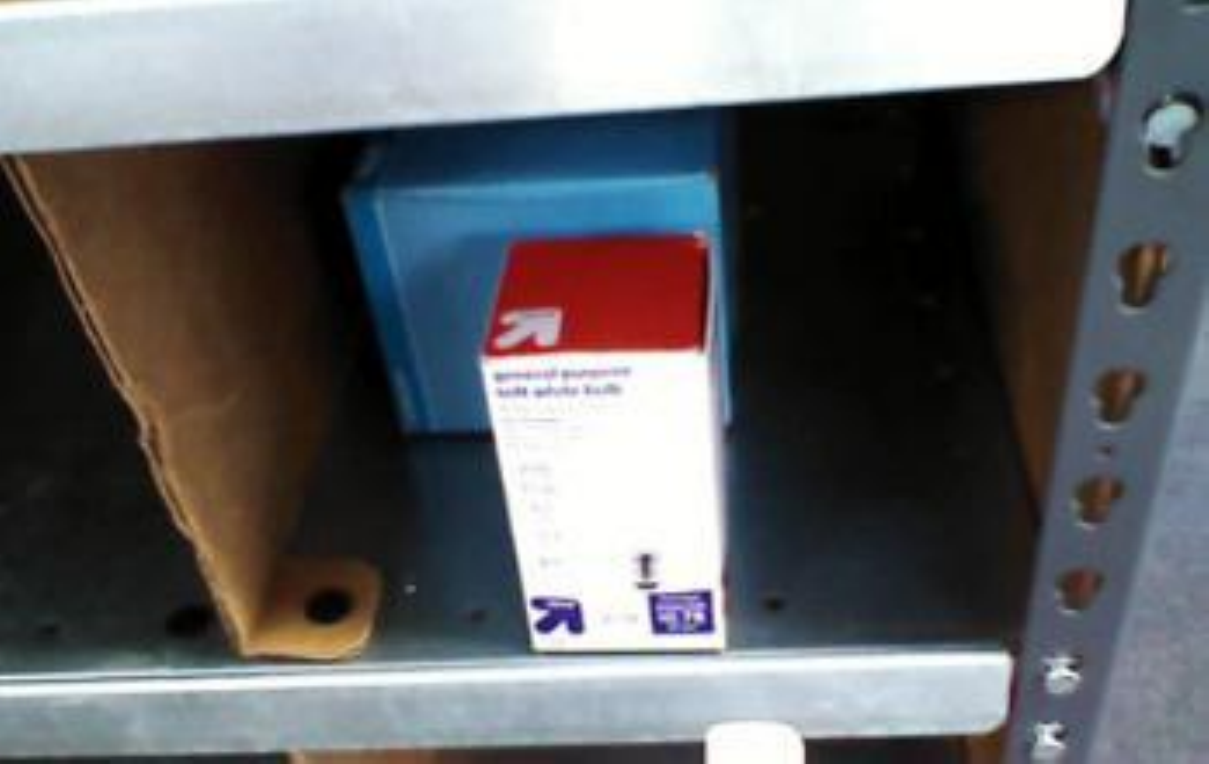}   & \tabimg{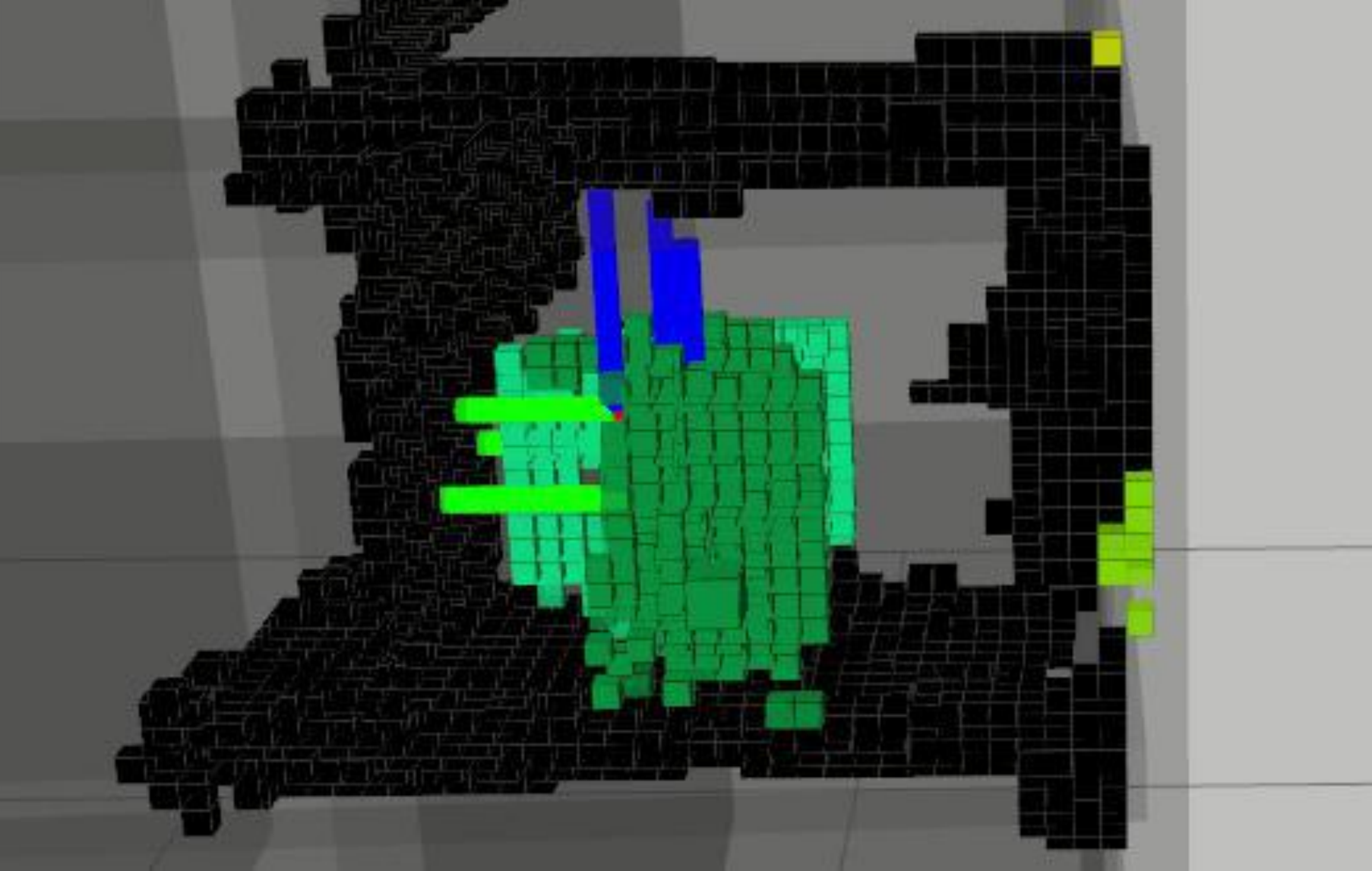}   & \cmark                 & \cmark      & \cmark       \\
    \rotateclock{scene 2} & \tabimg{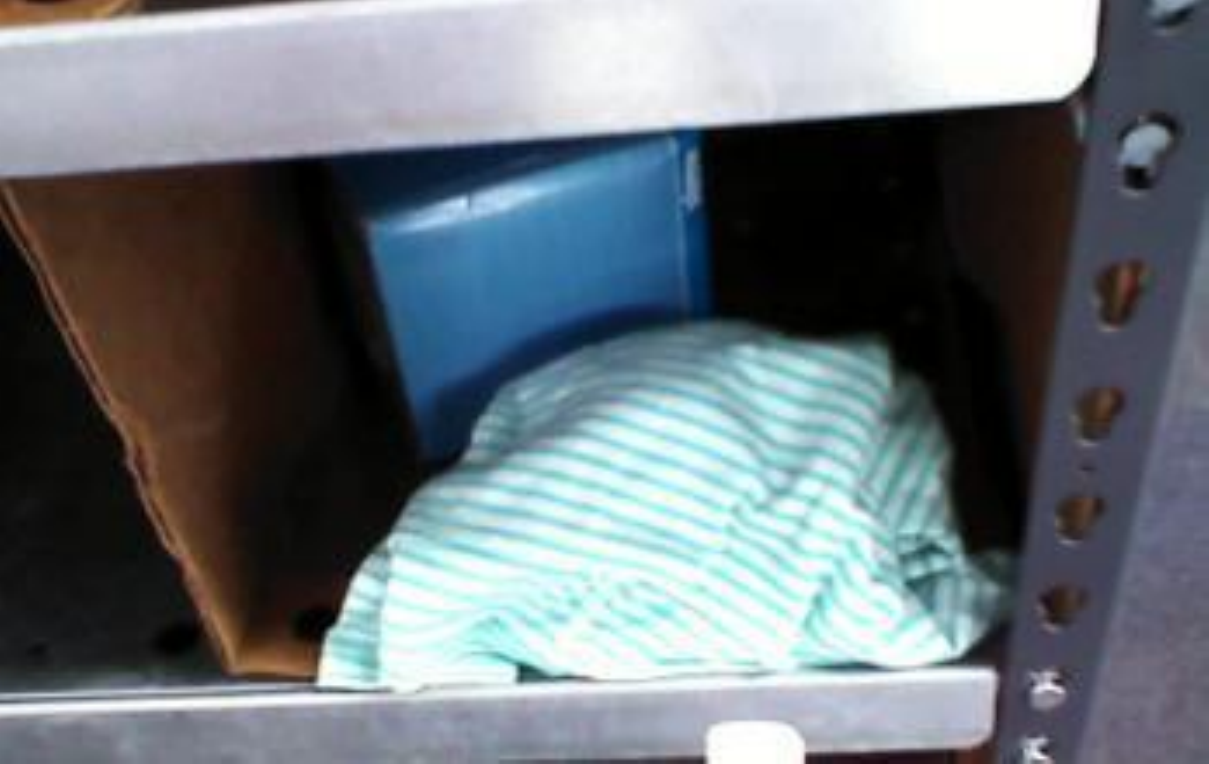}  & \tabimg{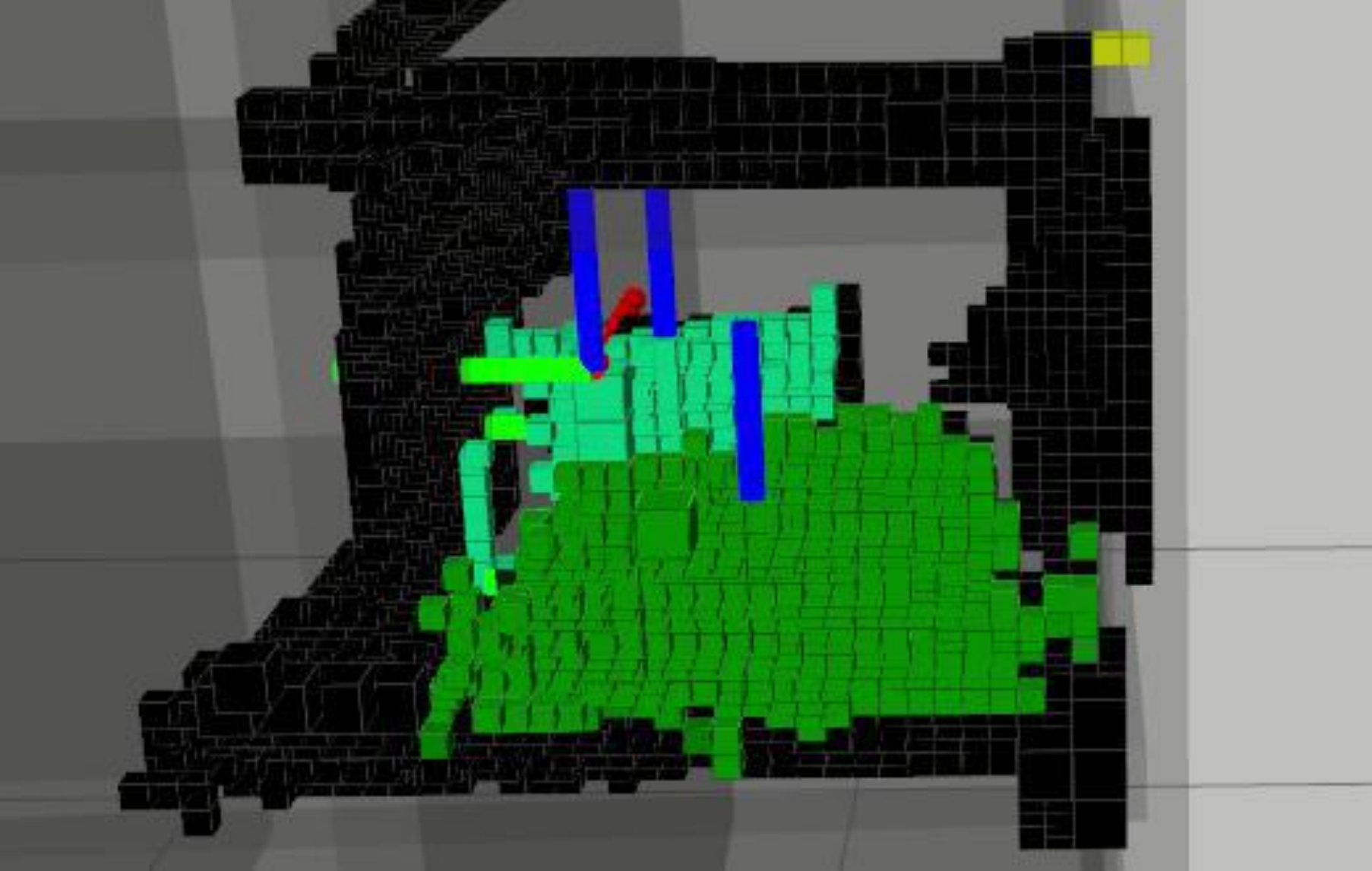}  & \cmark                 & \cmark      & \xmark       \\
    \rotateclock{scene 3} & \tabimg{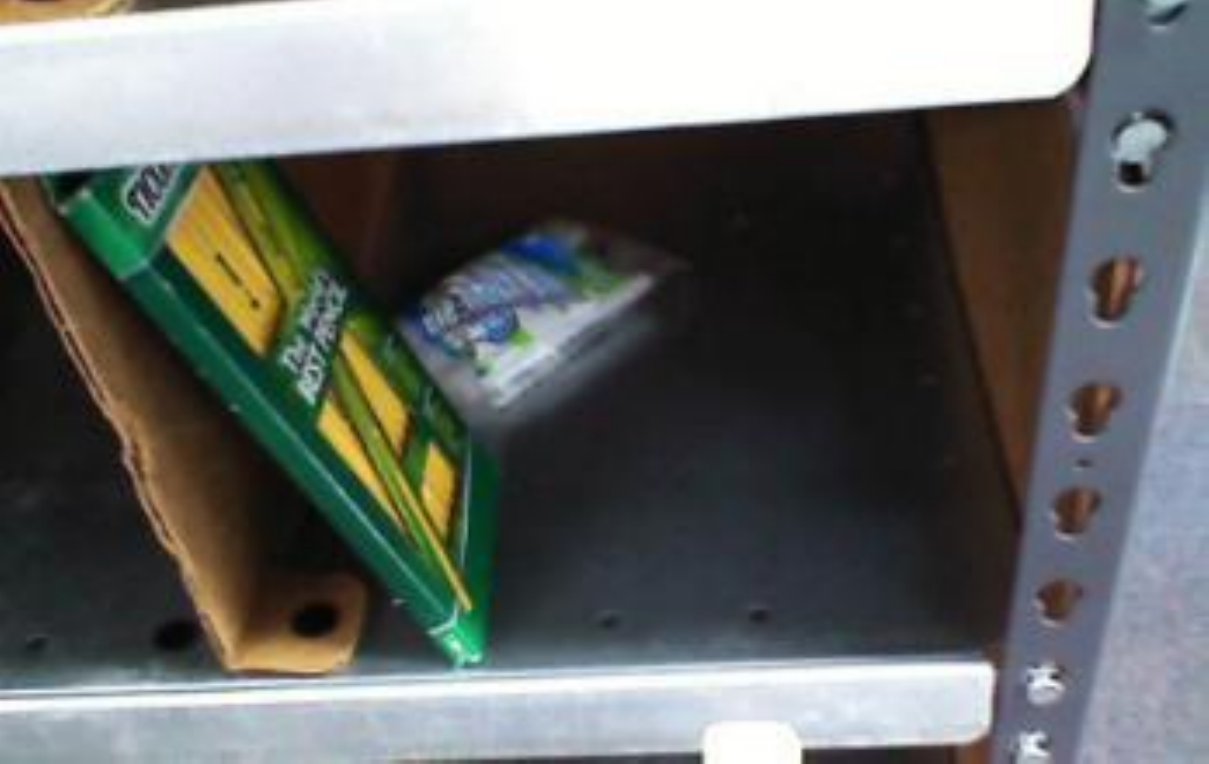} & \tabimg{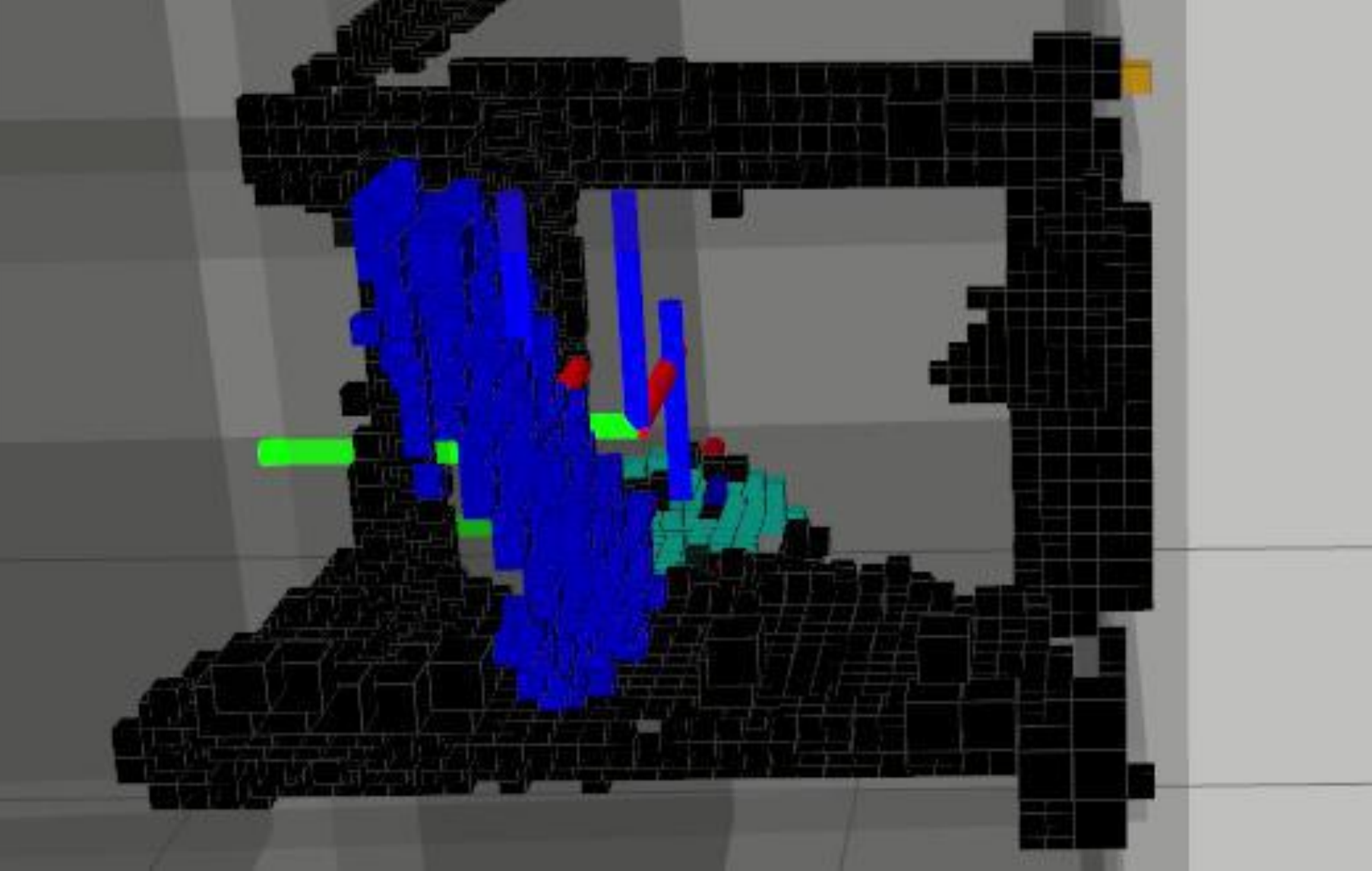} & \cmark                 & \xmark      & \textemdash  \\
    \rotateclock{scene 4} & \tabimg{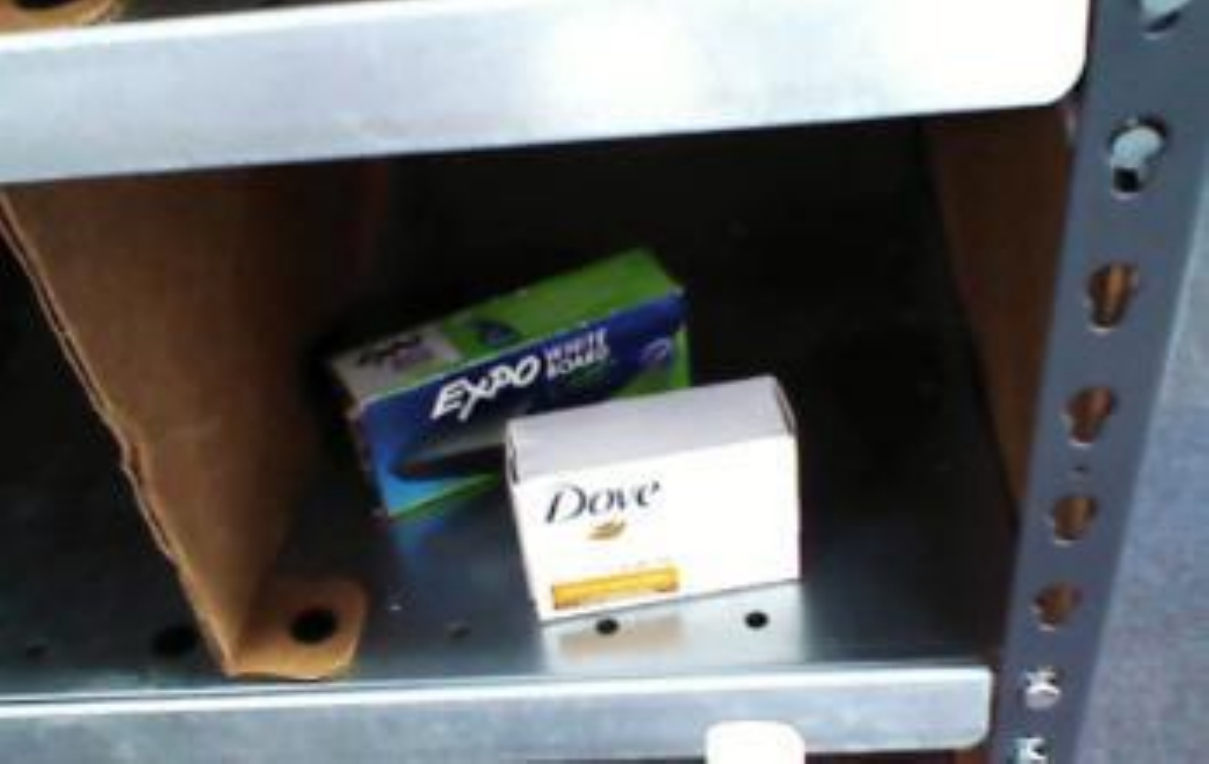} & \tabimg{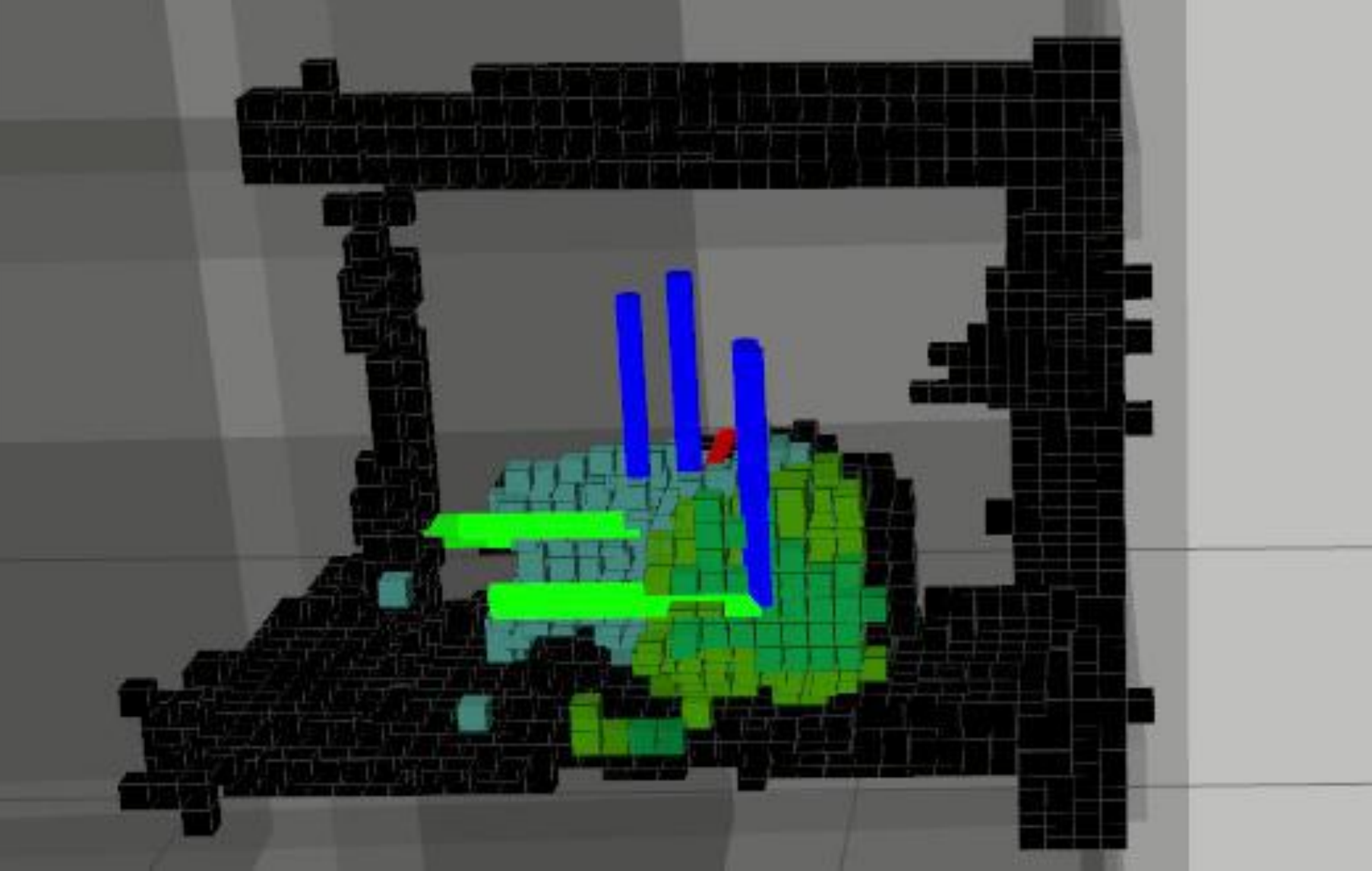} & \xmark                 & \textemdash & \textemdash  \\
  \end{tabular}
\end{table}

% Result

% \begin{figure}[htbp]
%   \centering
%   \includegraphics[width=0.48\textwidth]{pick_backward_22}
%   \caption{Manipulation task in an environment where an obstacle (book) is
%            located in front of the target object (tissue box).
%     \newline
%     \footnotesize{
%       In the first image, robot is segmenting objects in the shelf bin,
%       and in the second it is picking the book,
%       in the third it is picking the tissue box.
%     }
%   }
%   \figlab{pick_tissue_box}
% \end{figure}
% -------------------------------------------------------------------------------------------------

\renewcommand{\imgwidth}{1.4cm}
\newcolumntype{a}{>{\columncolor{Gray}}r}
\begin{table*}[htbp]
  \centering
  \caption{\textbf{Segmentation results of the 39 objects used at APC2016.}
    \small{
      The blue box in images indicates the target object for segmentation,
      and bold text in table highlights the maximum, underline does competitive,
      and no decoration means all low accuracies (less than 1) in each row.
      The correnpondence between voxel color and label value is same
      as that shown in \tabref{fcn_qualitative_result}.
    }
  }
  \tablab{iu_result}
  \begin{minipage}{0.49\textwidth}
  \centering
  \begin{tabular}{c|rra|cc}
  {}          & \multicolumn{3}{c}{$IU_{3d}$}                                                                                                     &   \multirow{2}{*}{Scene}                &   \multirow{2}{*}{\thead{Result \\ (LabelOctoMap)}}           \\
  {}          &          \thead{Projec- \\ tion \\ (Mean)} &           \thead{Projec- \\ tion \\ (Max)} &      \thead{Label- \\ Octo- \\ Map} &                                                                          \\
  label       &                                              &                                              &                                     &                                    &                                     \\
  \hline
  1           &                                         7.81 &                                        10.74 &                      \textbf{16.11} &   \tabimg{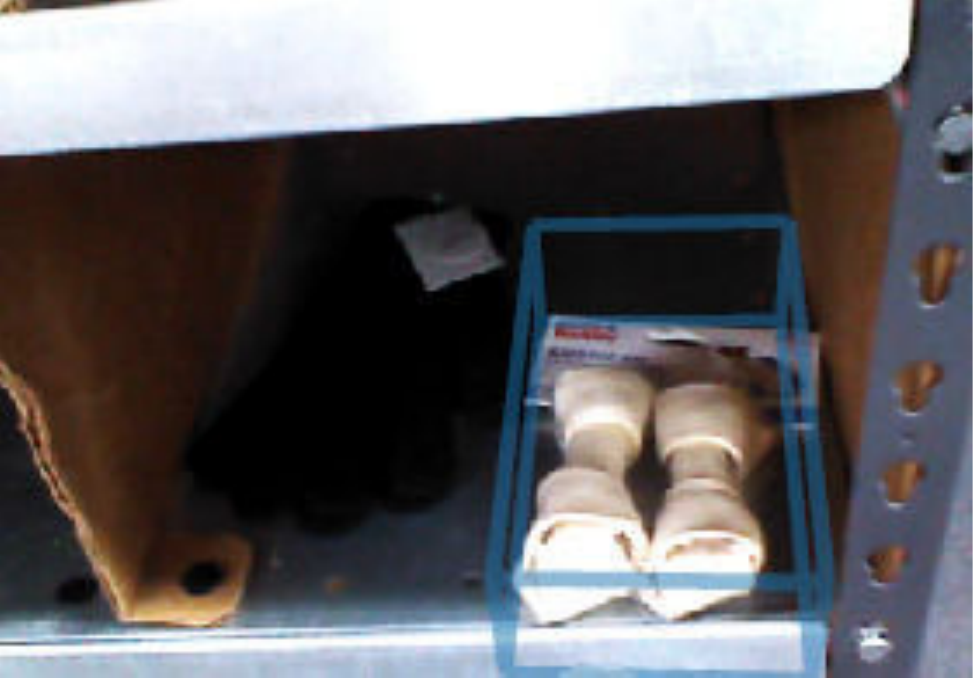} &   \tabimg{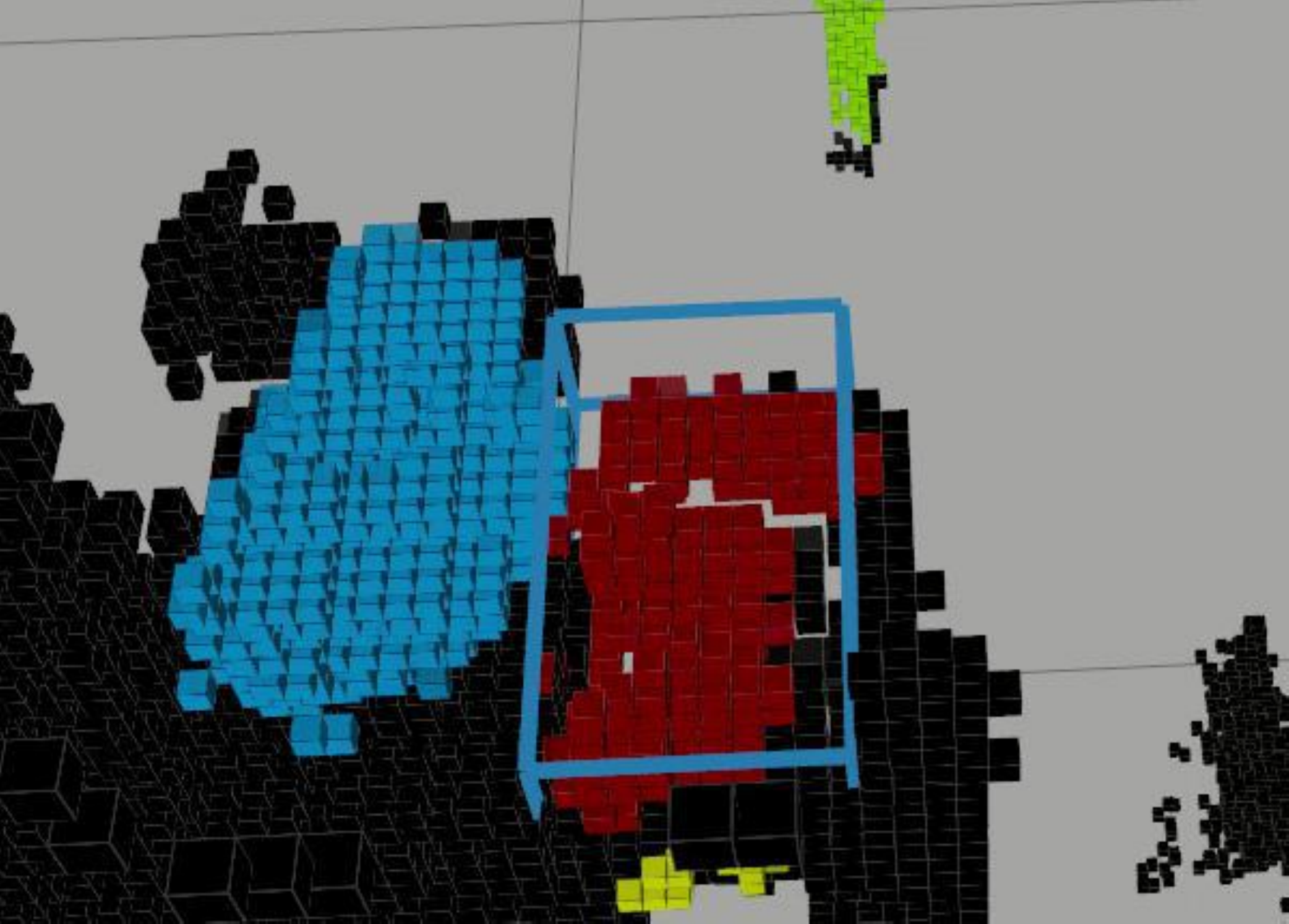} \\
  2           &                                         5.35 &                                         7.93 &                      \textbf{12.99} &   \tabimg{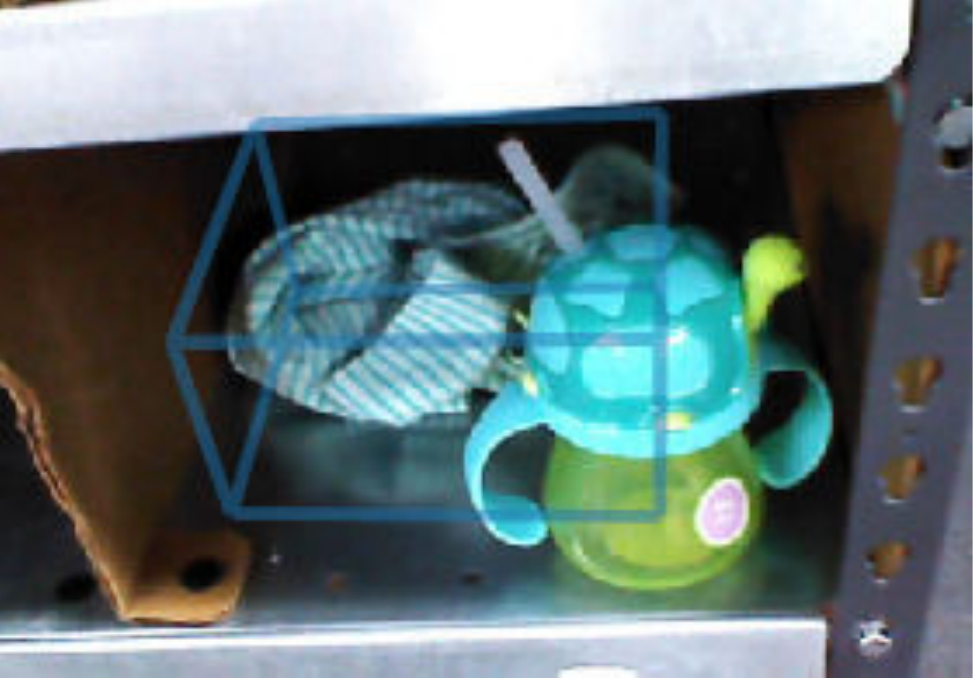} &   \tabimg{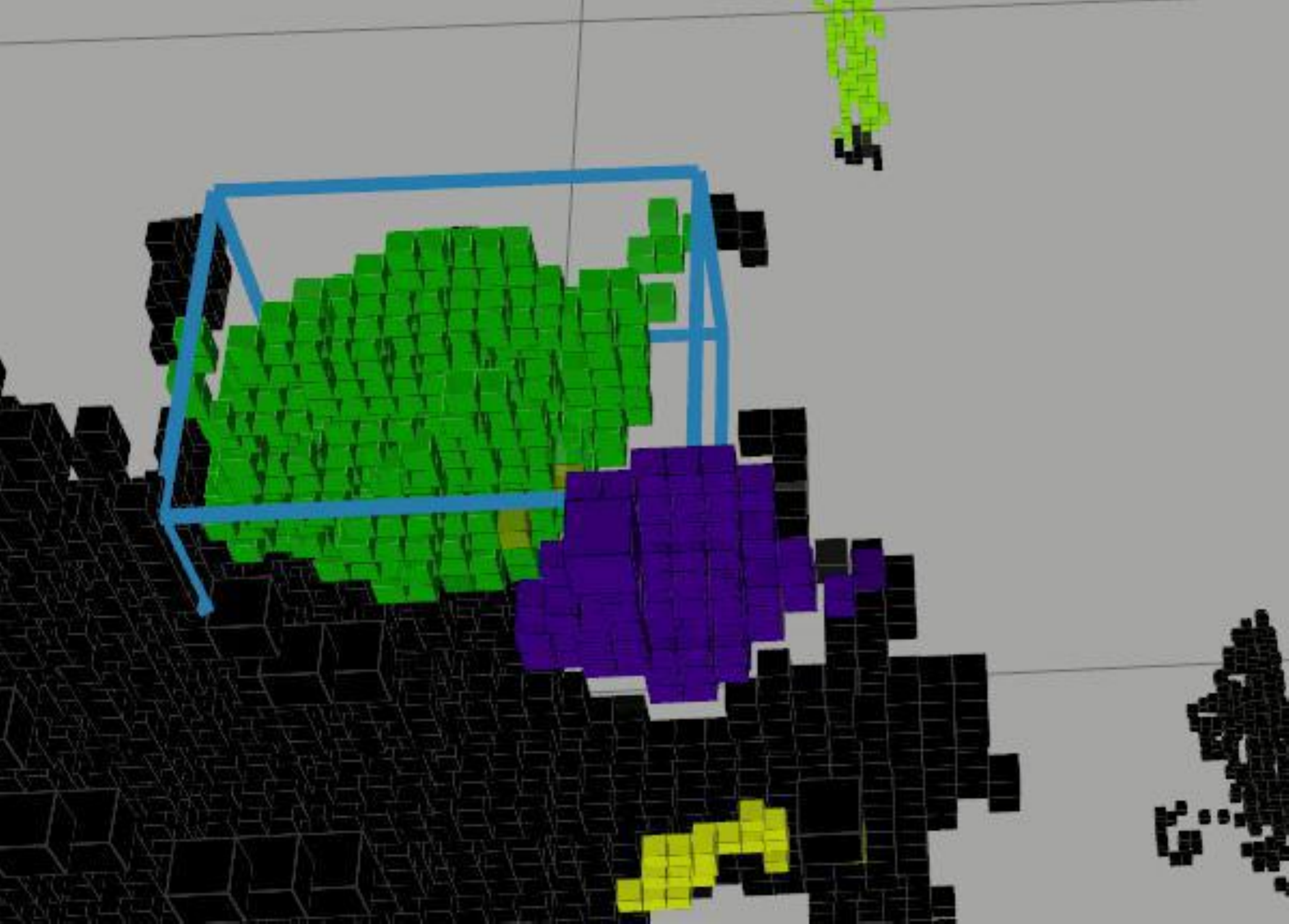} \\
  3           &                                         8.68 &                                        10.05 &                      \textbf{15.94} &   \tabimg{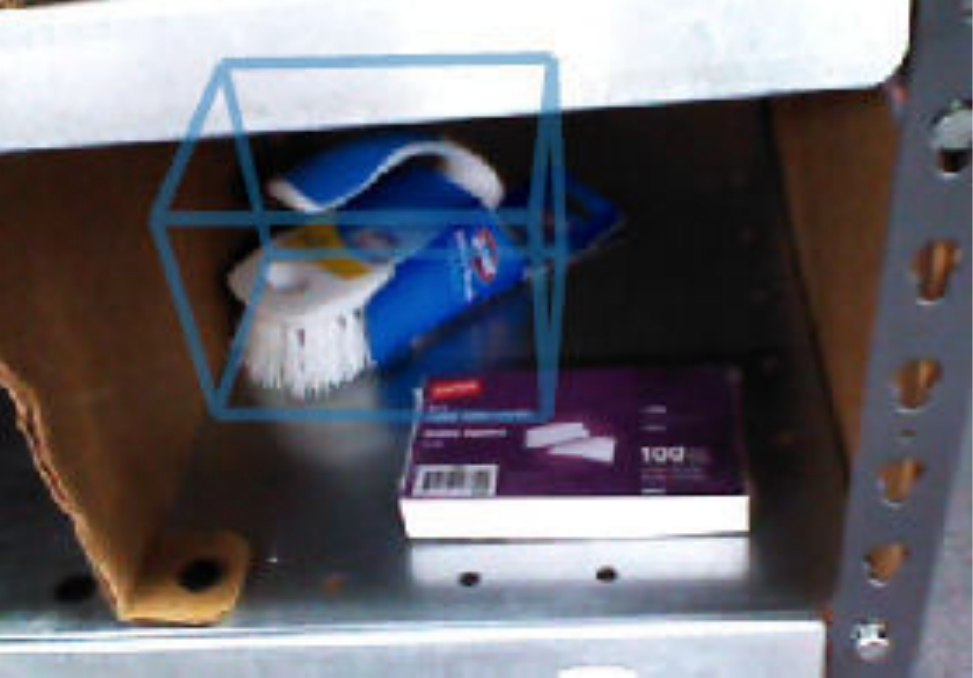} &   \tabimg{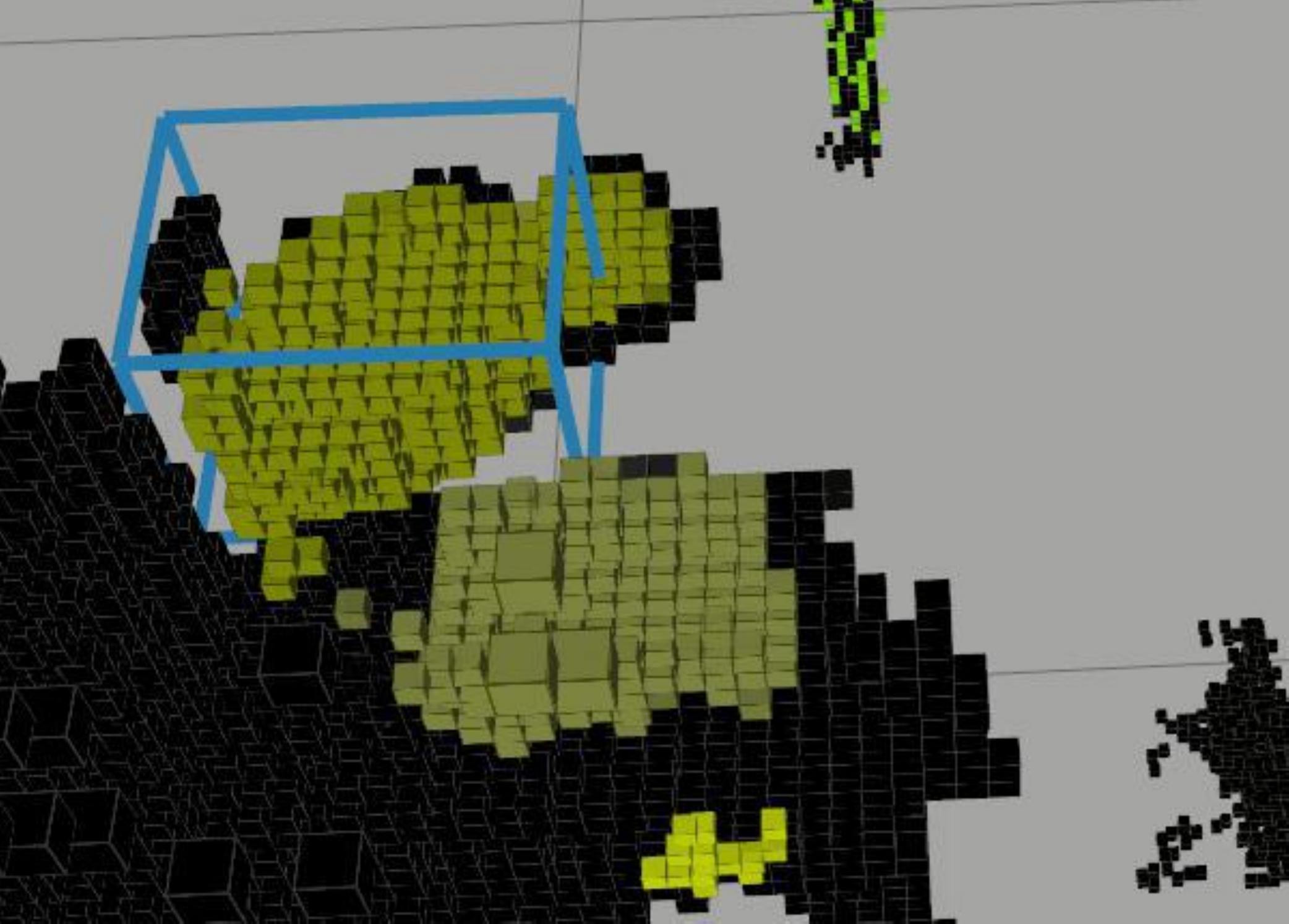} \\
  4           &                                         2.10 &                                         4.91 &                       \textbf{5.39} &   \tabimg{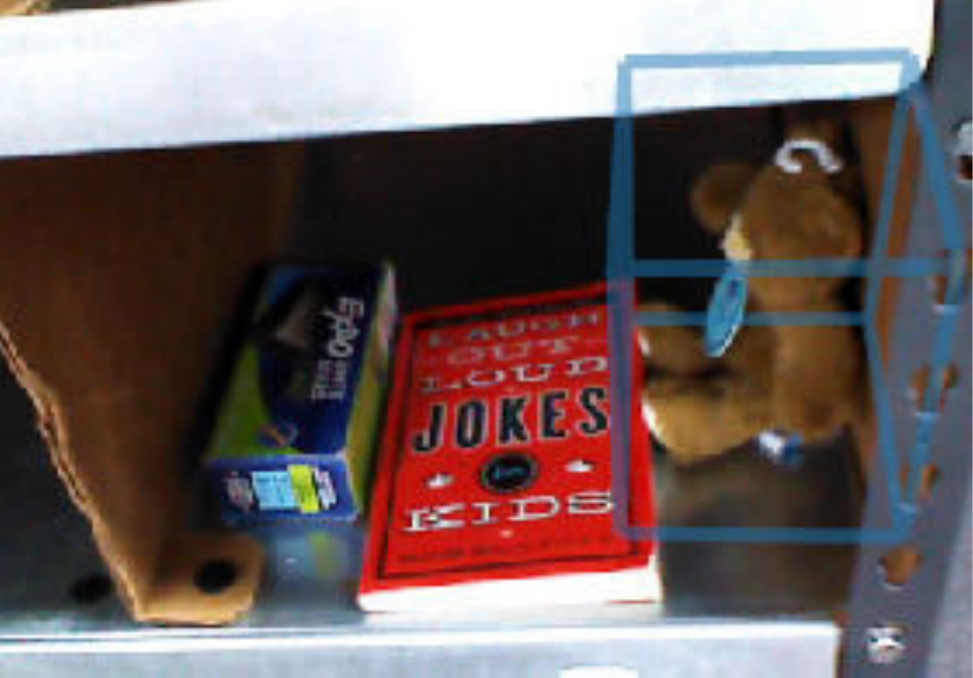} &   \tabimg{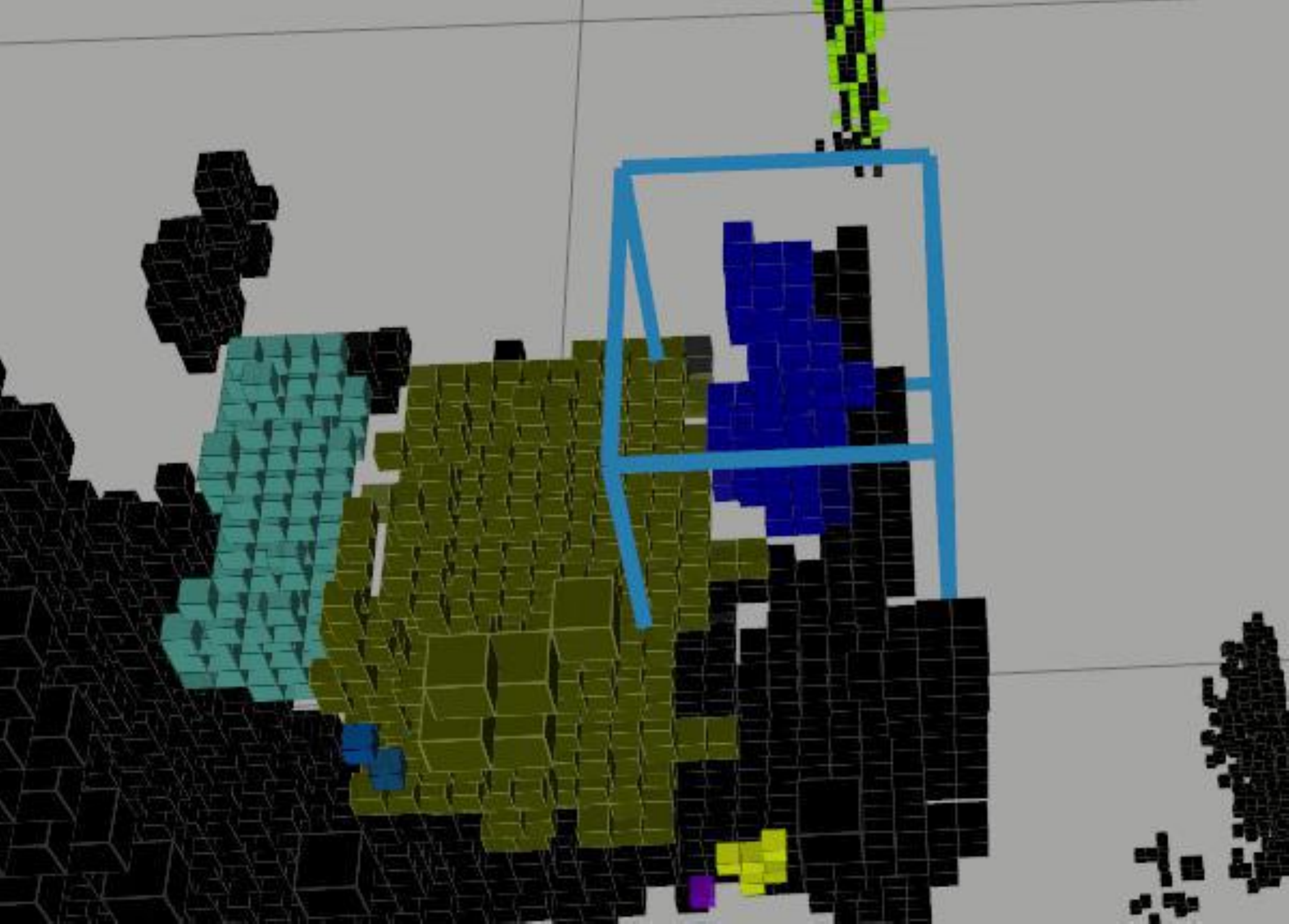} \\
  5           &                                         0.07 &                                         0.36 &                                0.05 &   \tabimg{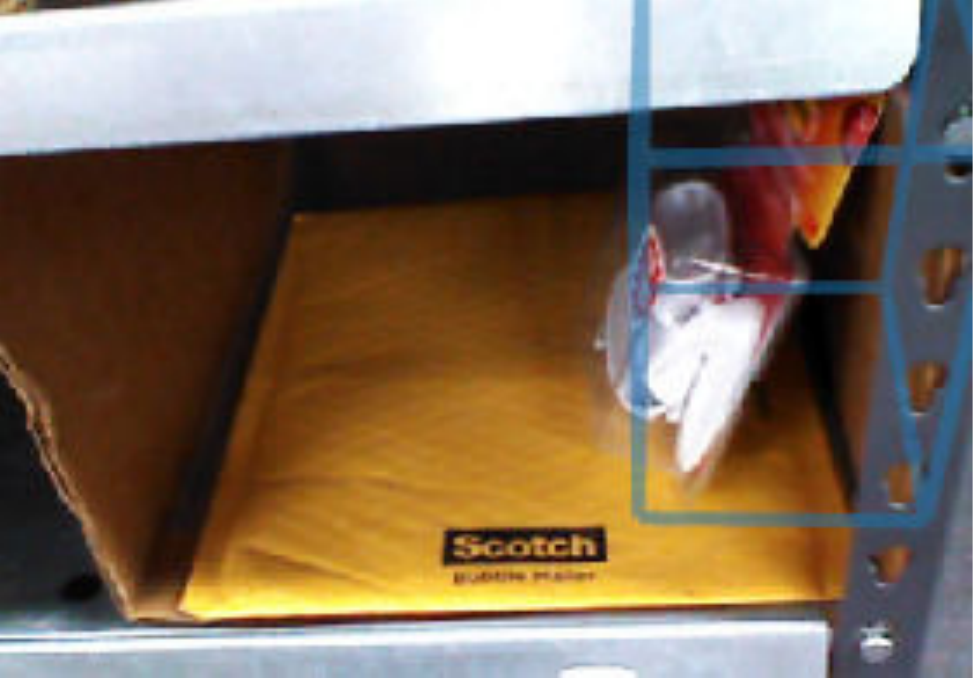} &   \tabimg{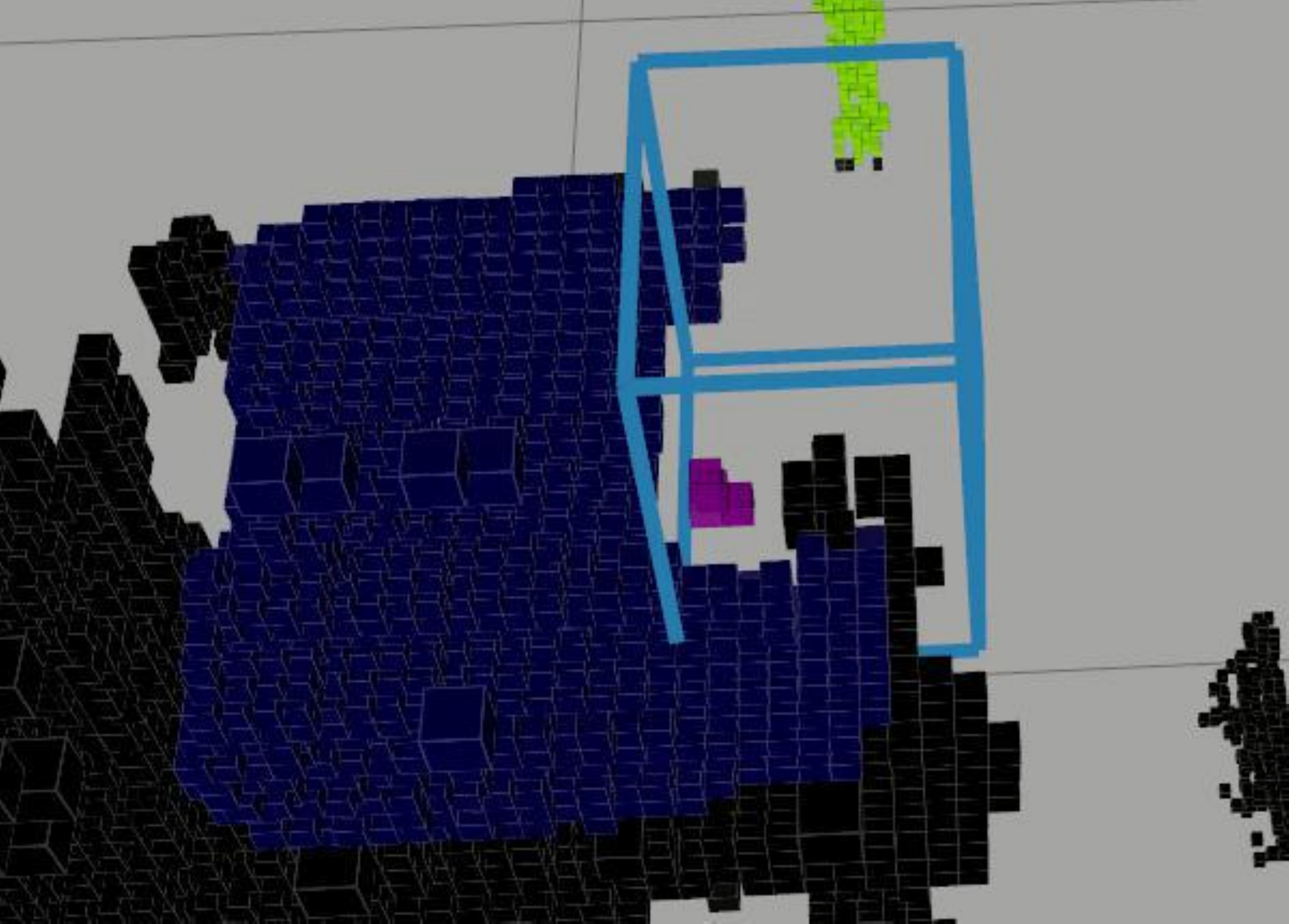} \\
  6           &                                        11.14 &                                        14.02 &                      \textbf{32.78} &   \tabimg{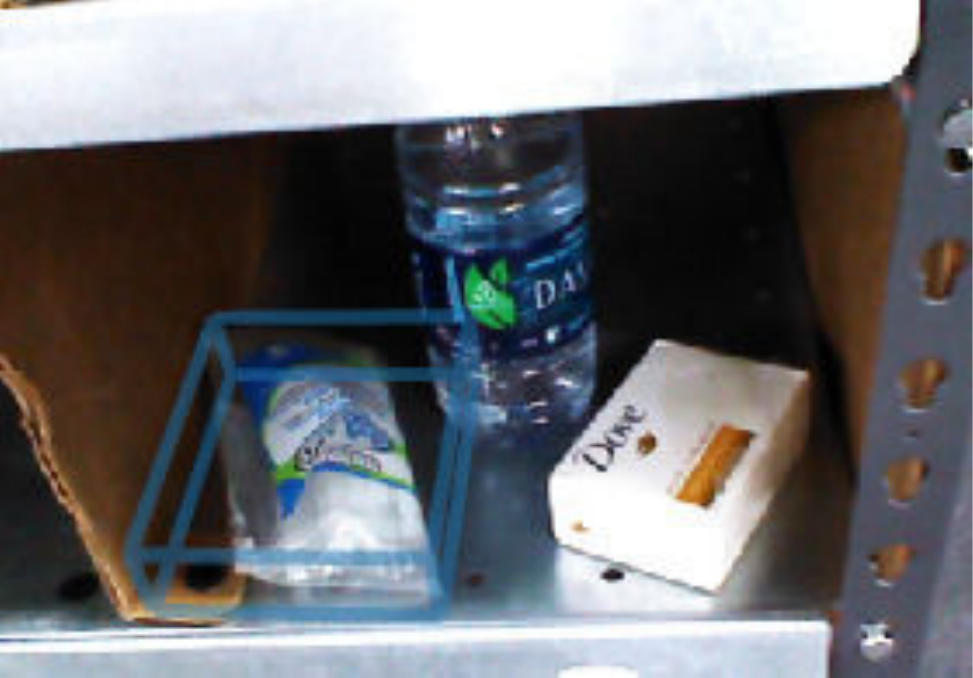} &   \tabimg{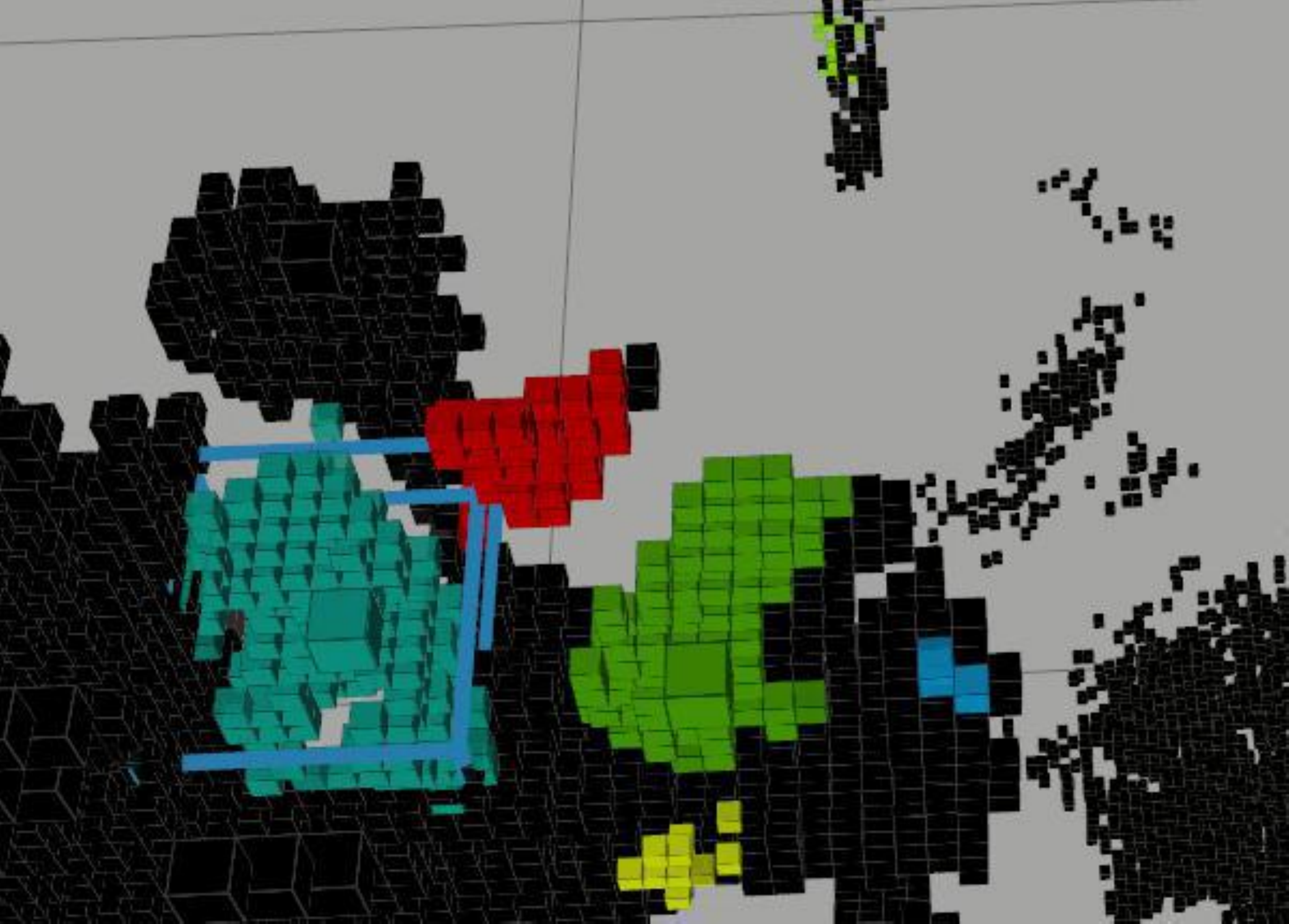} \\
  7           &                                         6.76 &                            \underline{18.30} &                   \underline{18.28} &   \tabimg{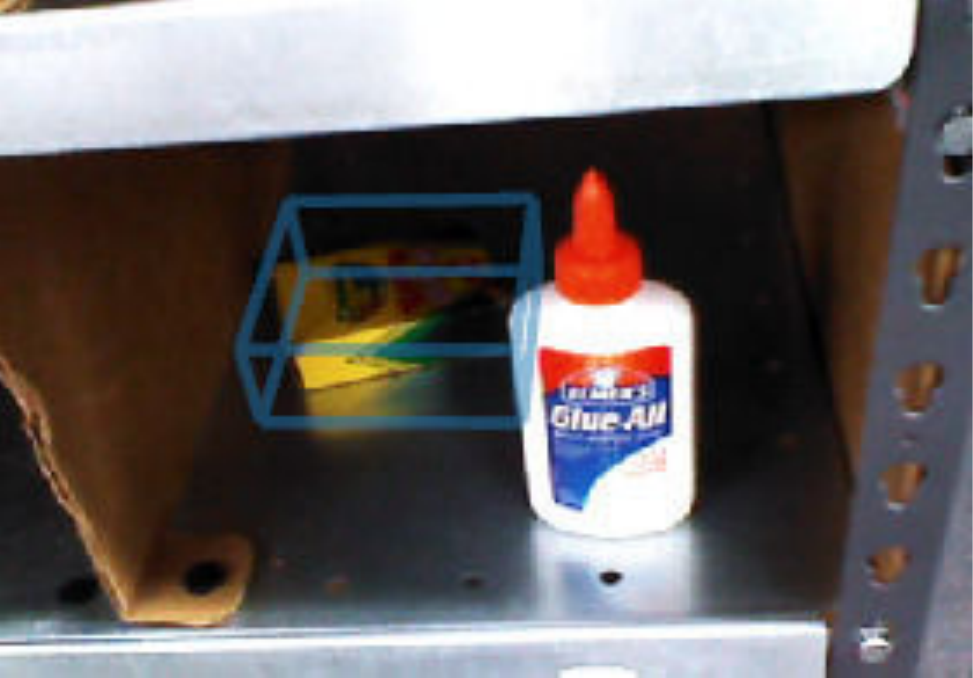} &   \tabimg{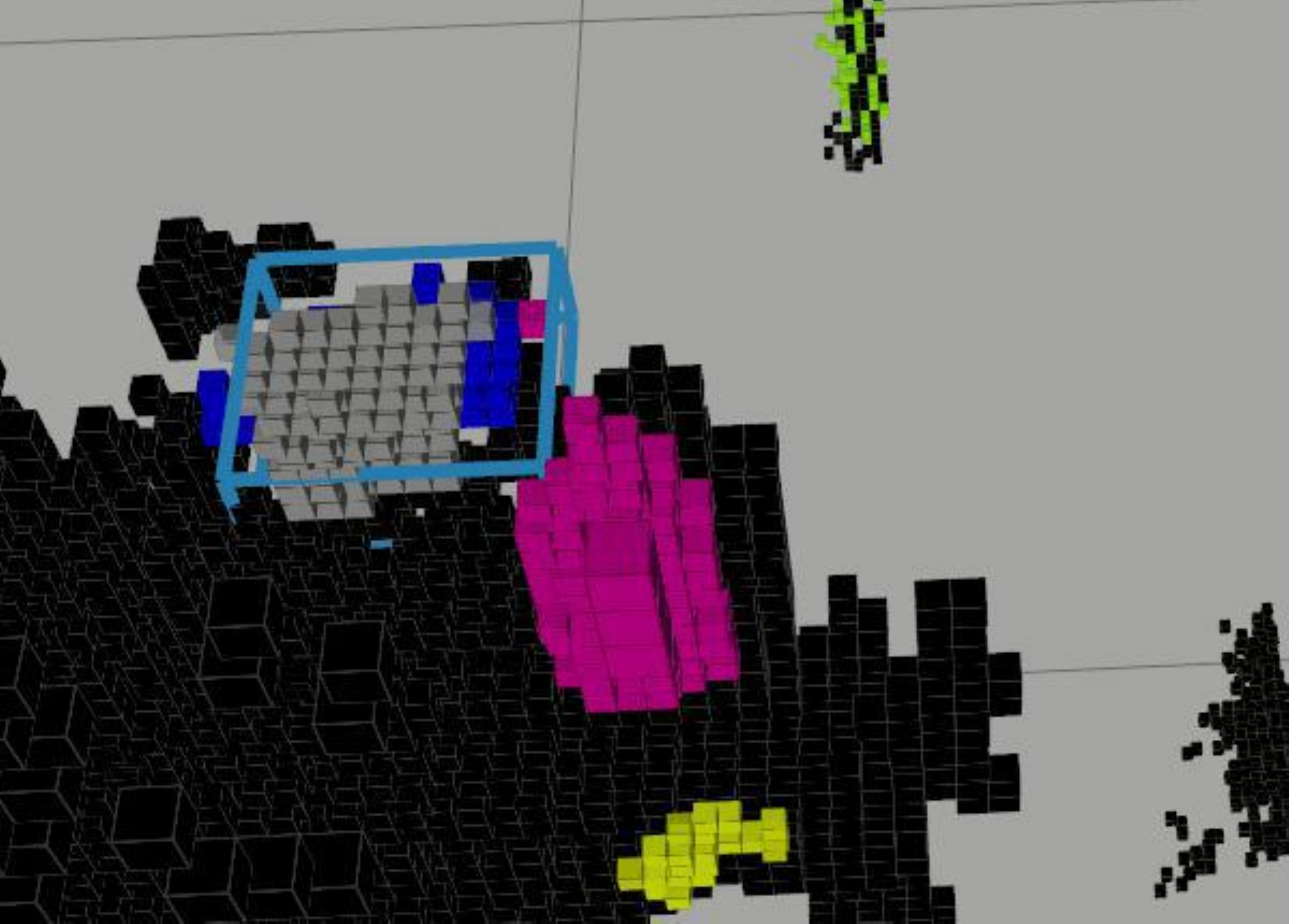} \\
  8           &                                         3.50 &                                         5.58 &                      \textbf{10.61} &   \tabimg{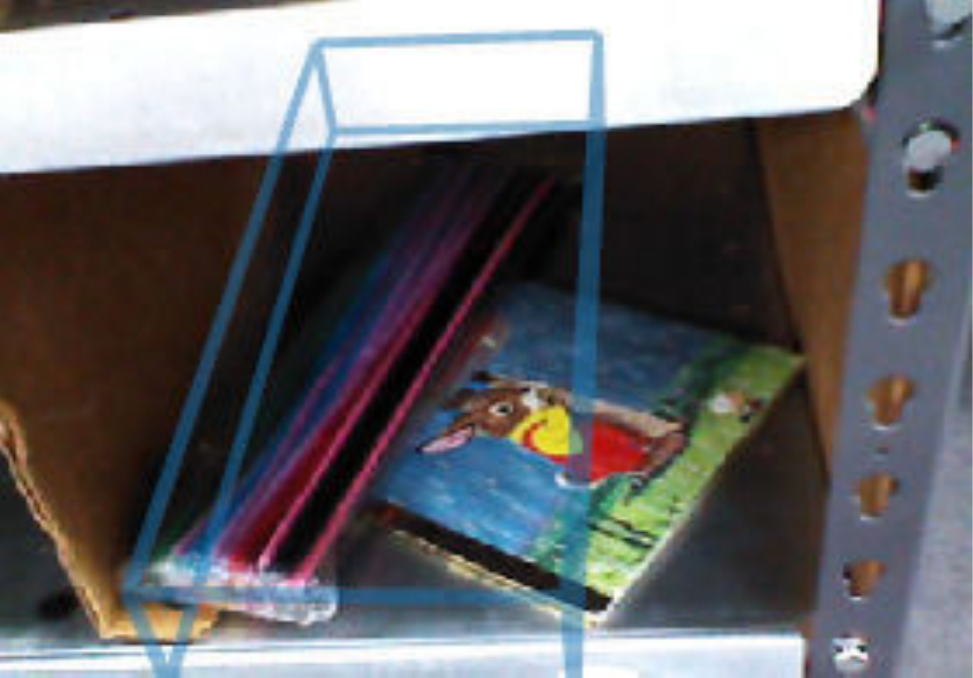} &   \tabimg{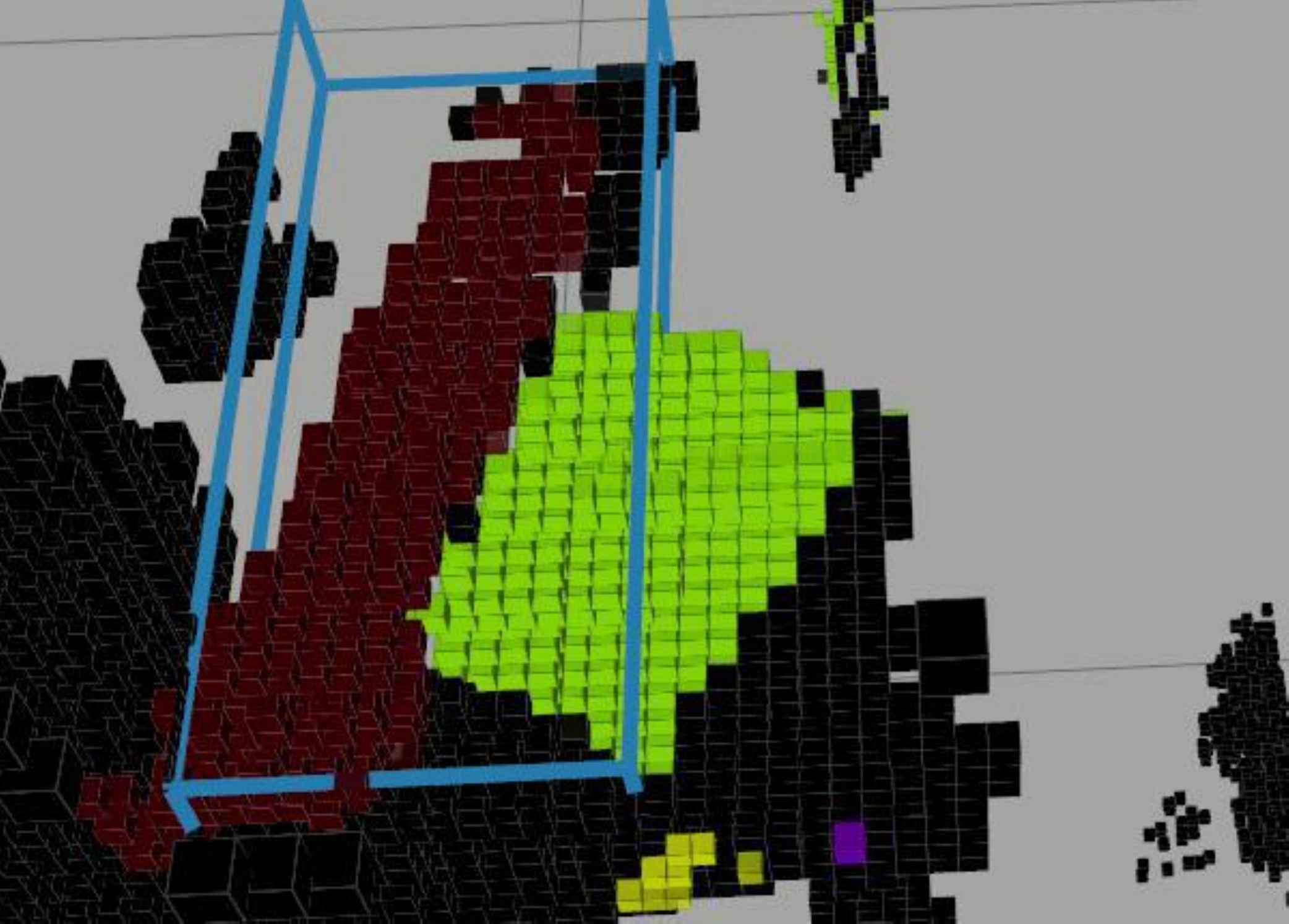} \\
  9           &                                         1.16 &                                         2.40 &                       \textbf{4.83} &   \tabimg{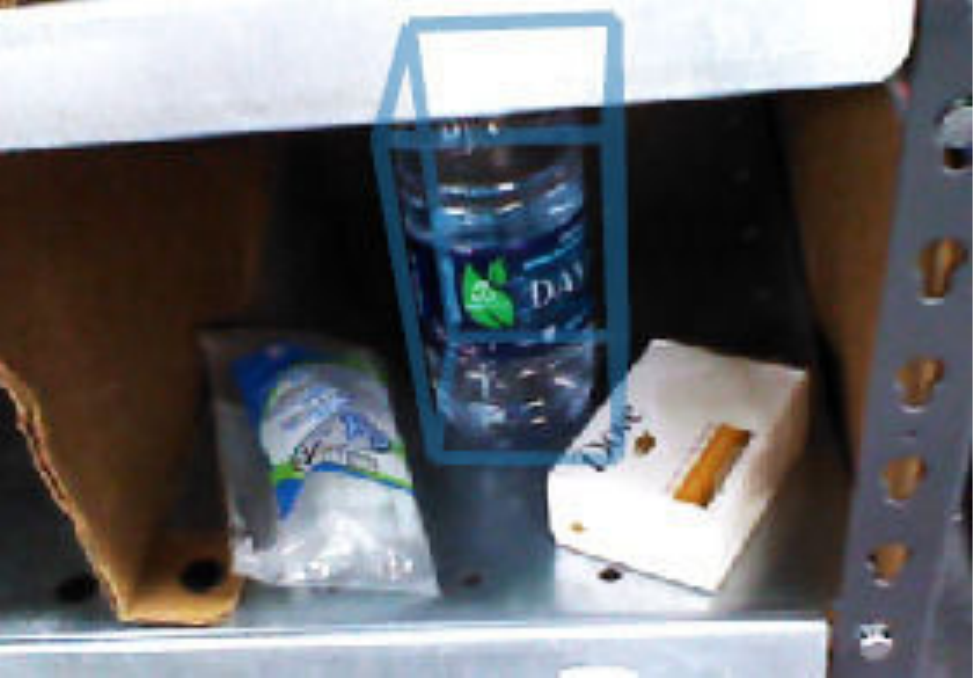} &   \tabimg{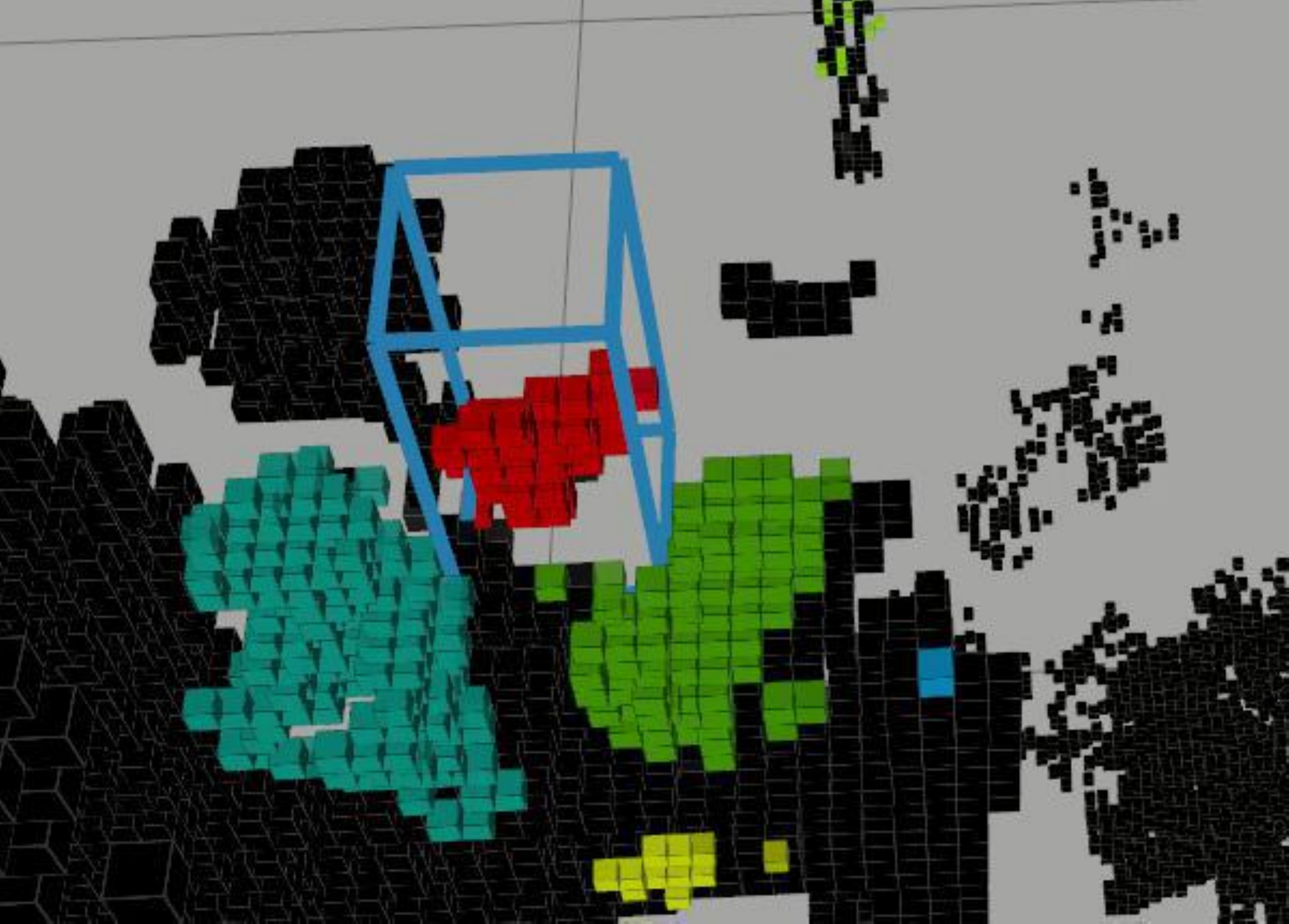} \\
  10          &                                         9.44 &                                        12.56 &                      \textbf{17.77} &  \tabimg{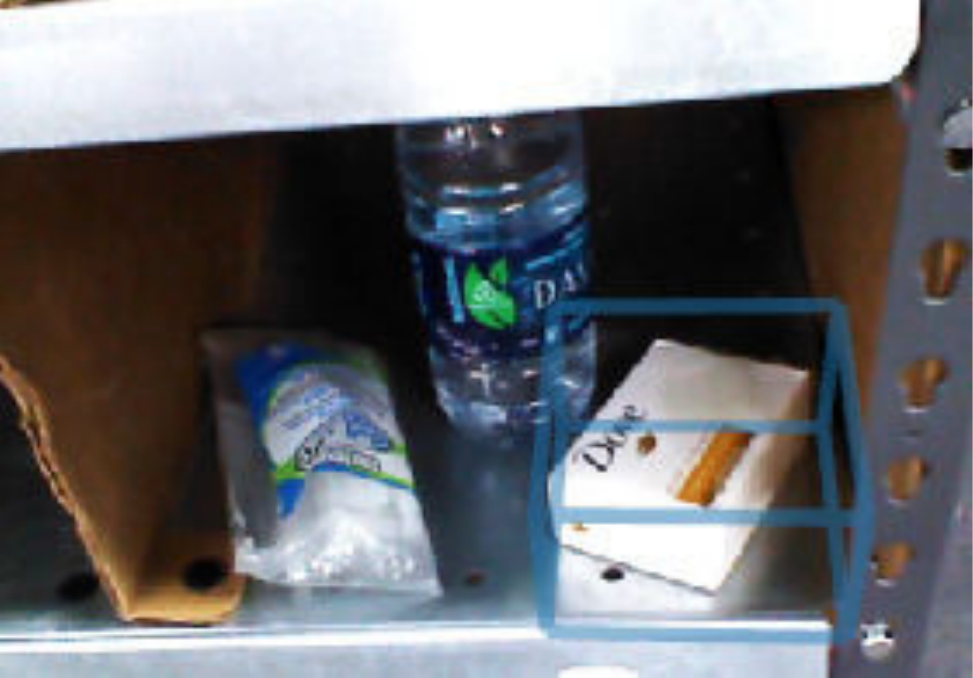} &  \tabimg{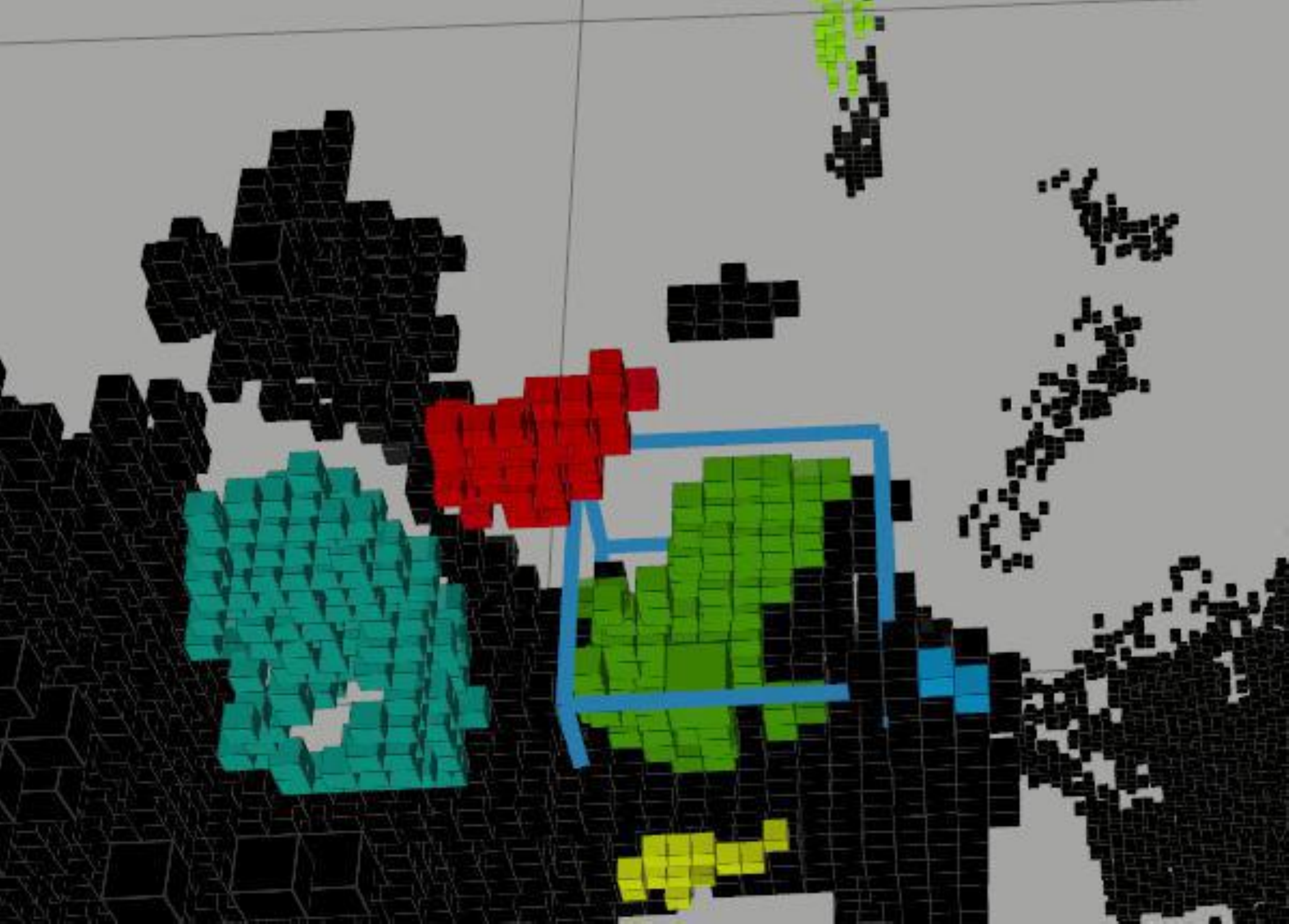} \\
  11          &                                         4.59 &                                        10.86 &                      \textbf{14.16} &  \tabimg{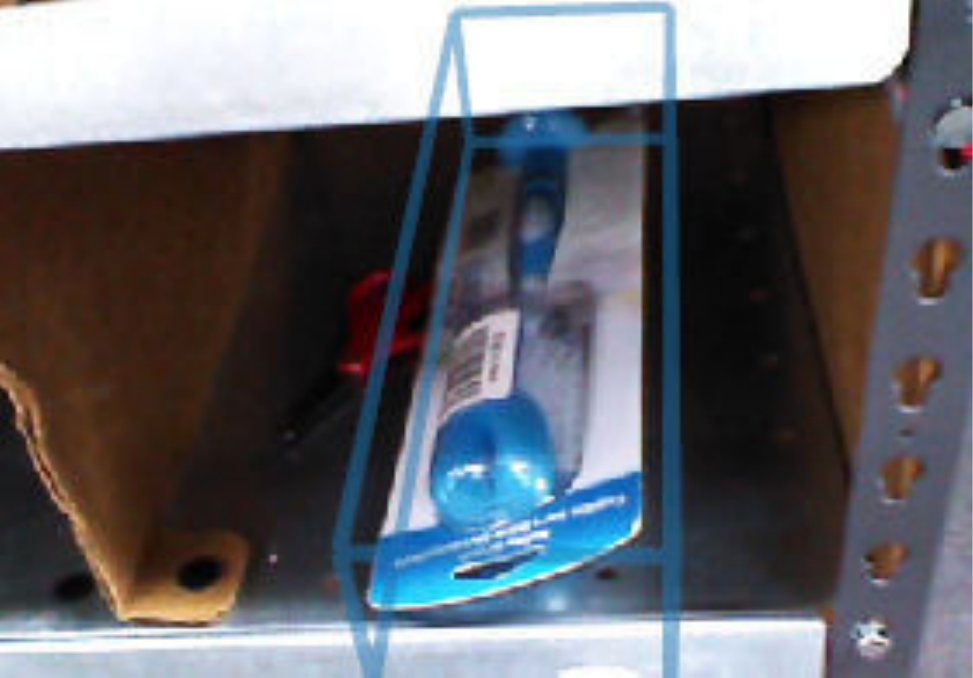} &  \tabimg{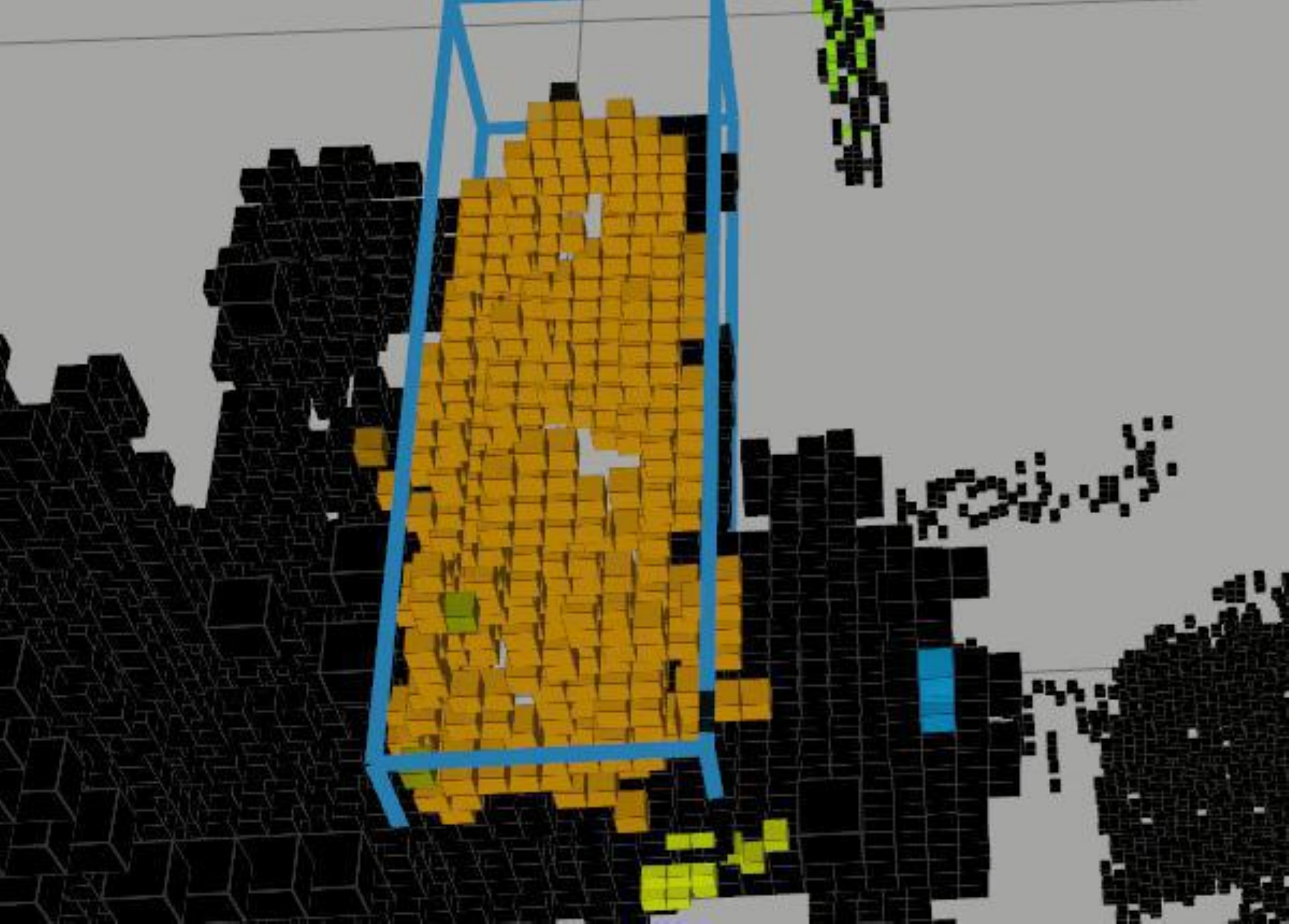} \\
  12          &                                         3.16 &                                         3.92 &                       \textbf{7.10} &  \tabimg{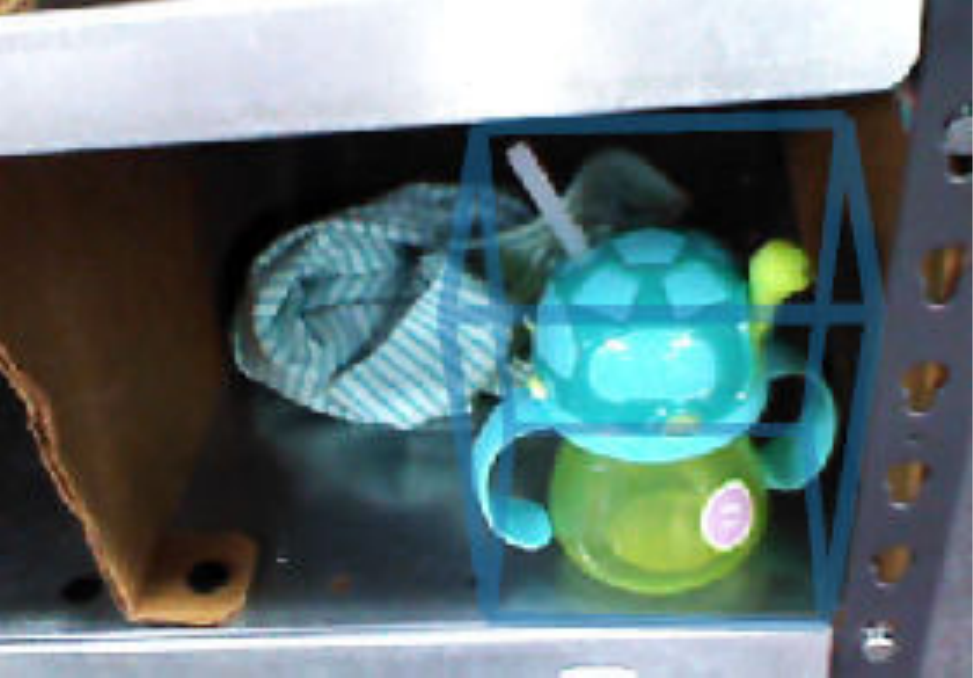} &  \tabimg{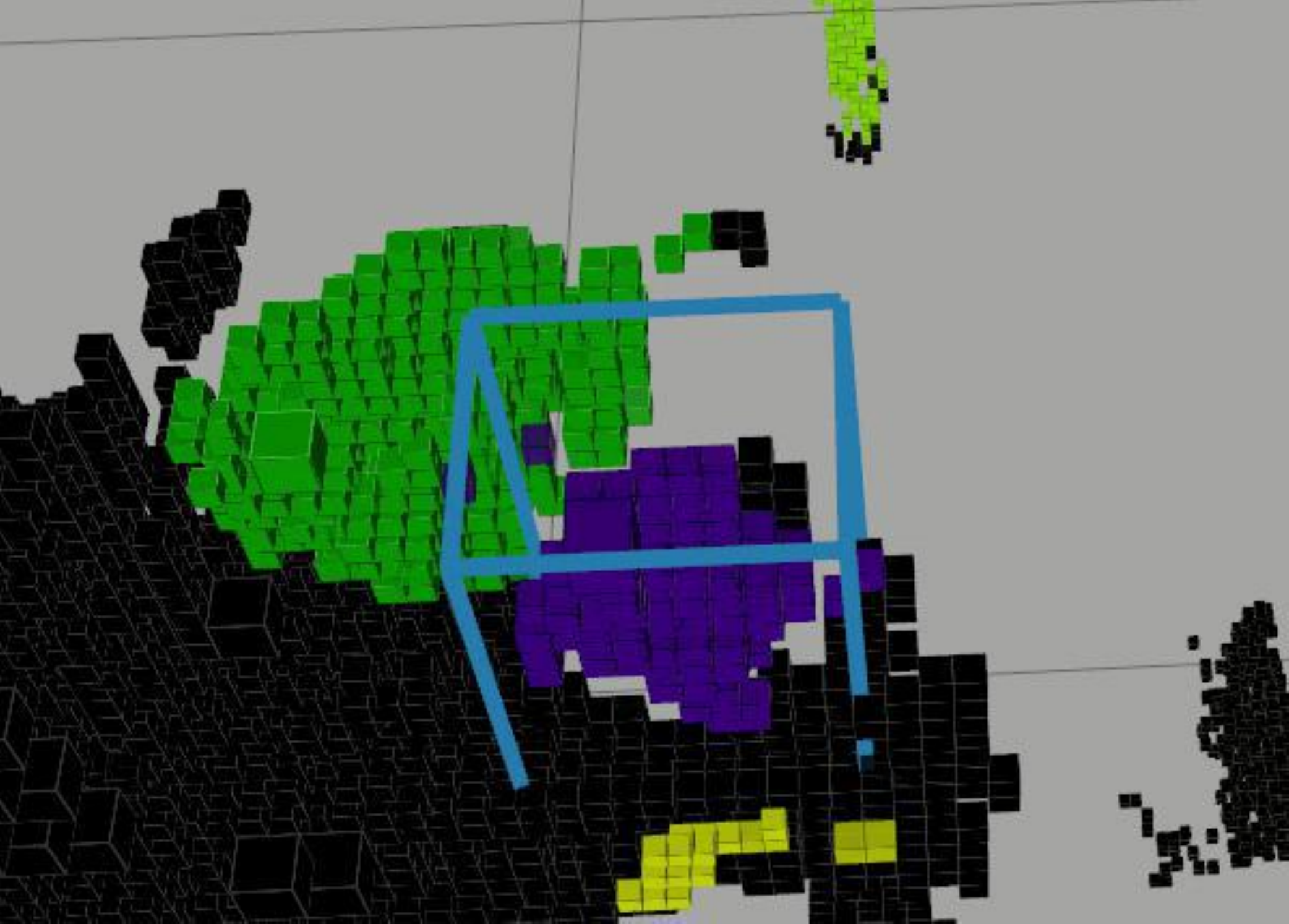} \\
  13          &                                        12.26 &                                        14.21 &                      \textbf{21.36} &  \tabimg{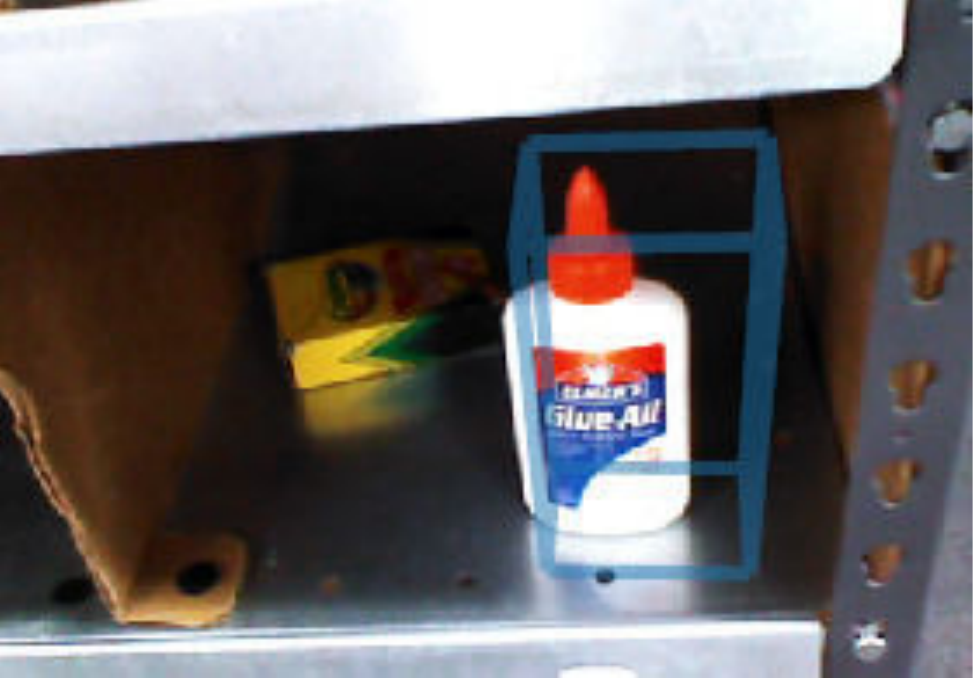} &  \tabimg{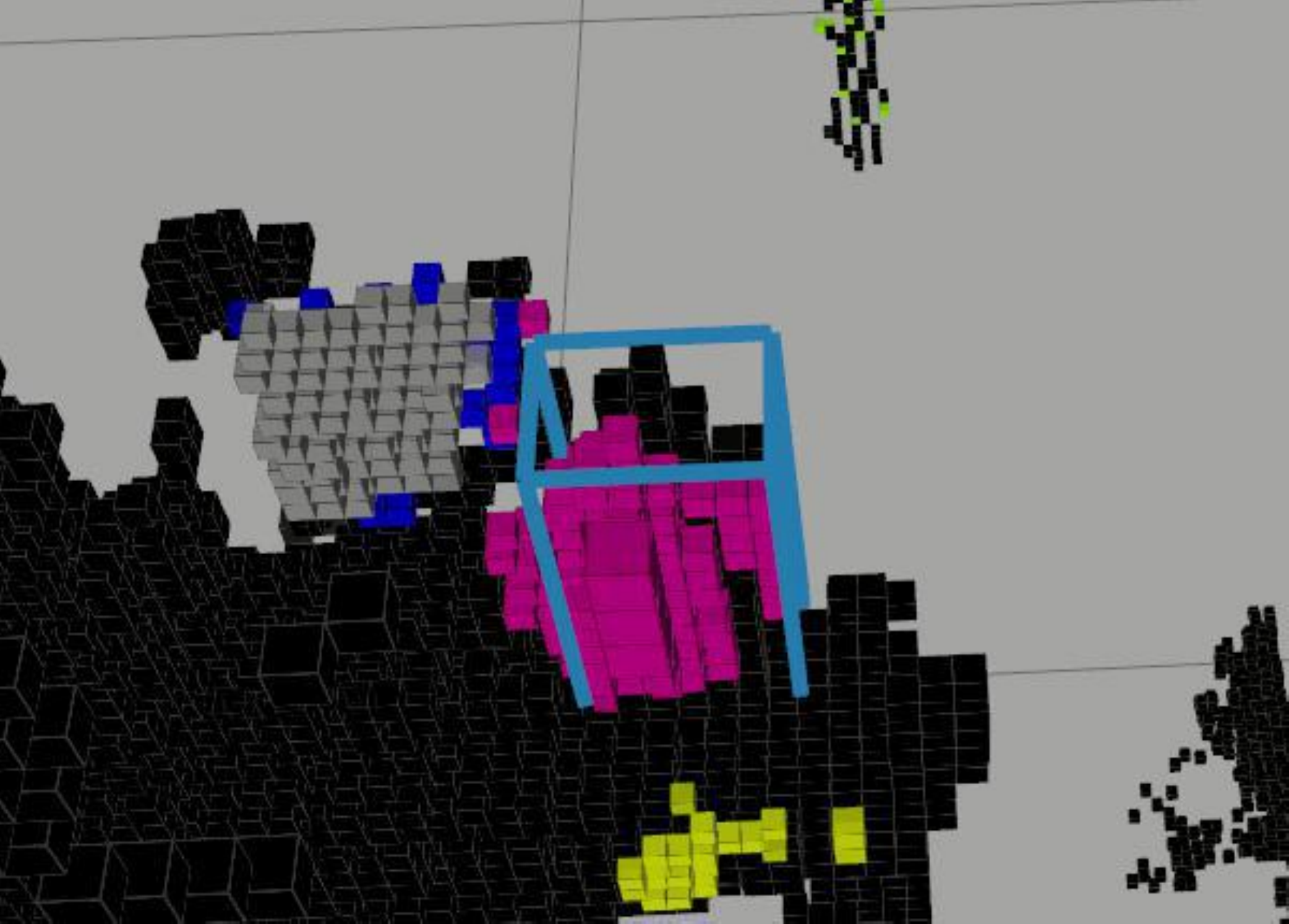} \\
  14          &                                        10.19 &                            \underline{16.68} &                   \underline{16.41} &  \tabimg{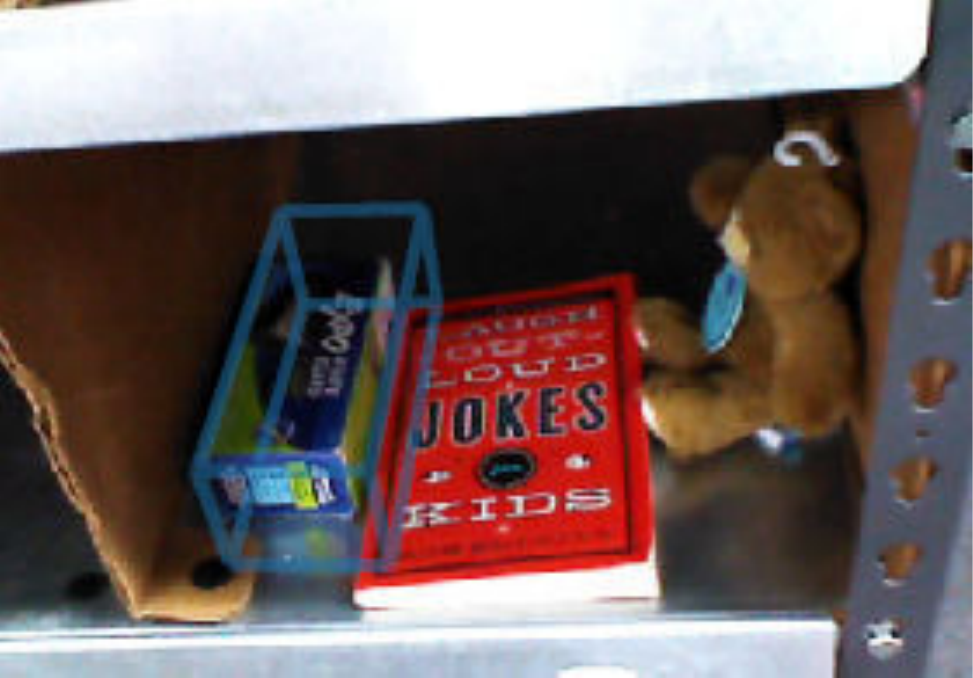} &  \tabimg{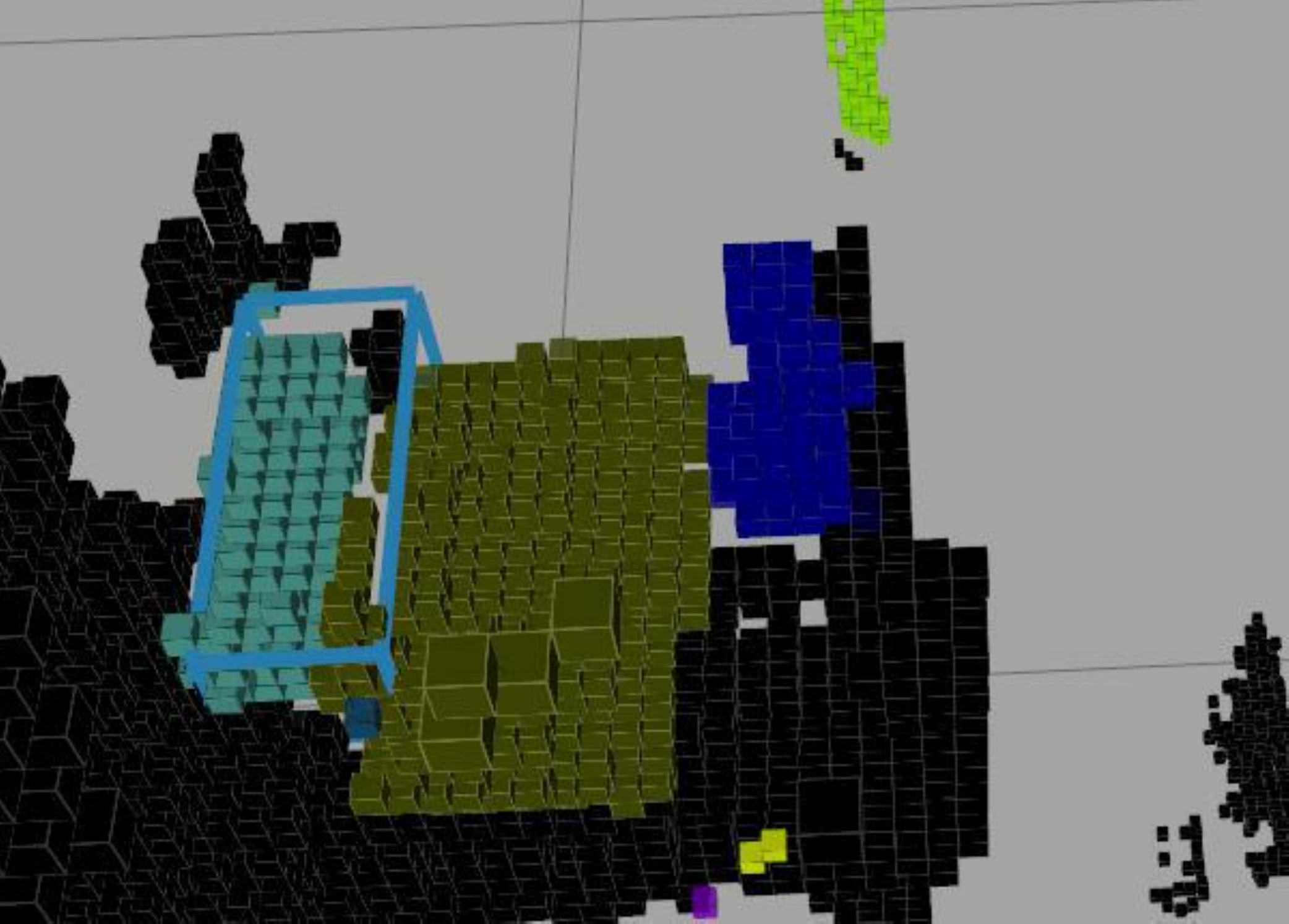} \\
  15          &                                         0.00 &                                         0.00 &                                0.00 &  \tabimg{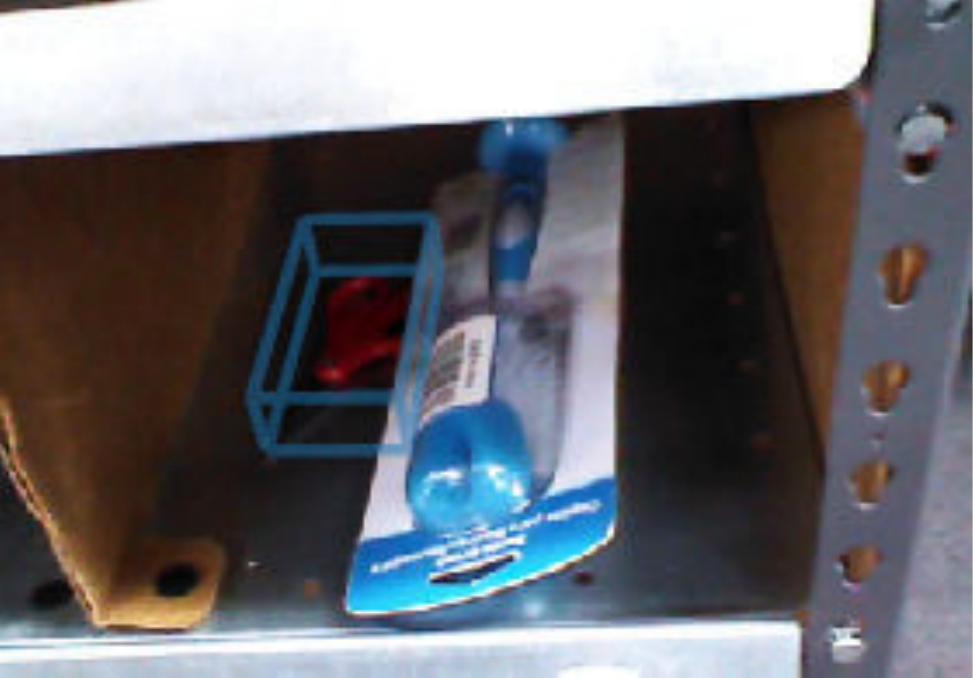} &  \tabimg{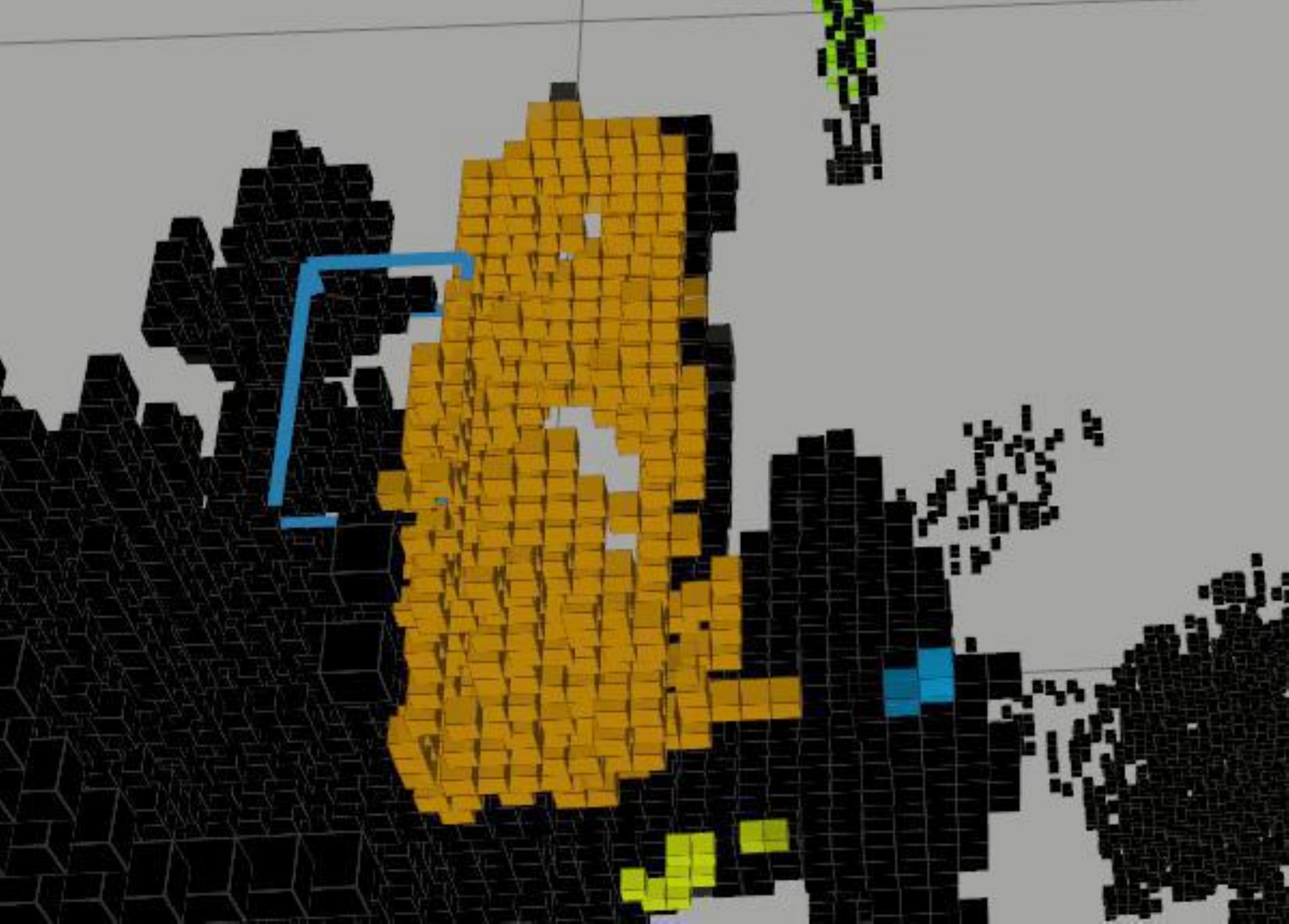} \\
  16          &                                         5.54 &                                         8.78 &                      \textbf{12.41} &  \tabimg{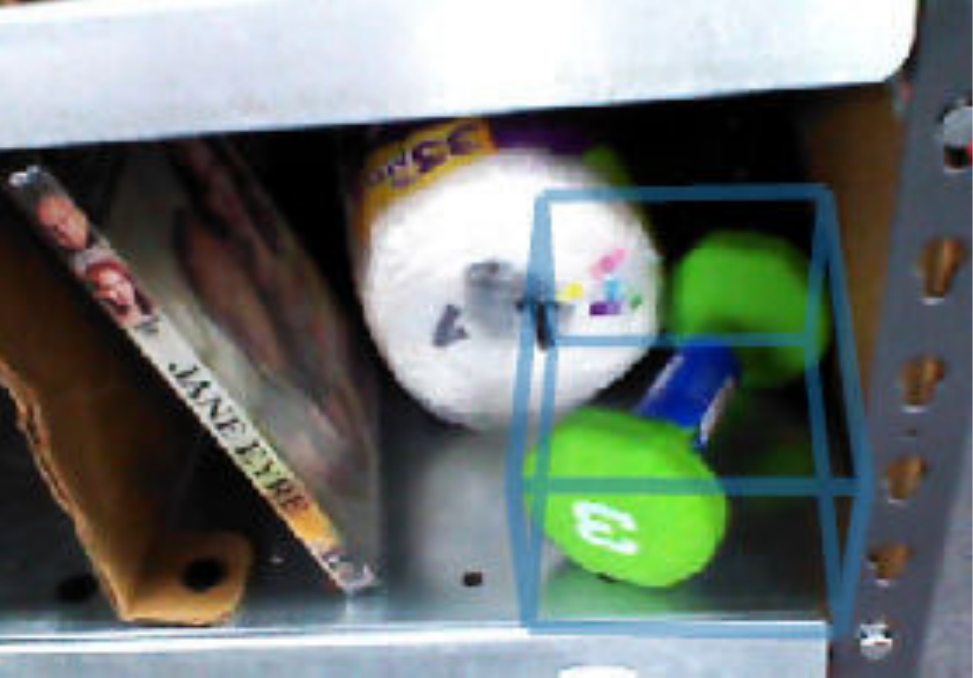} &  \tabimg{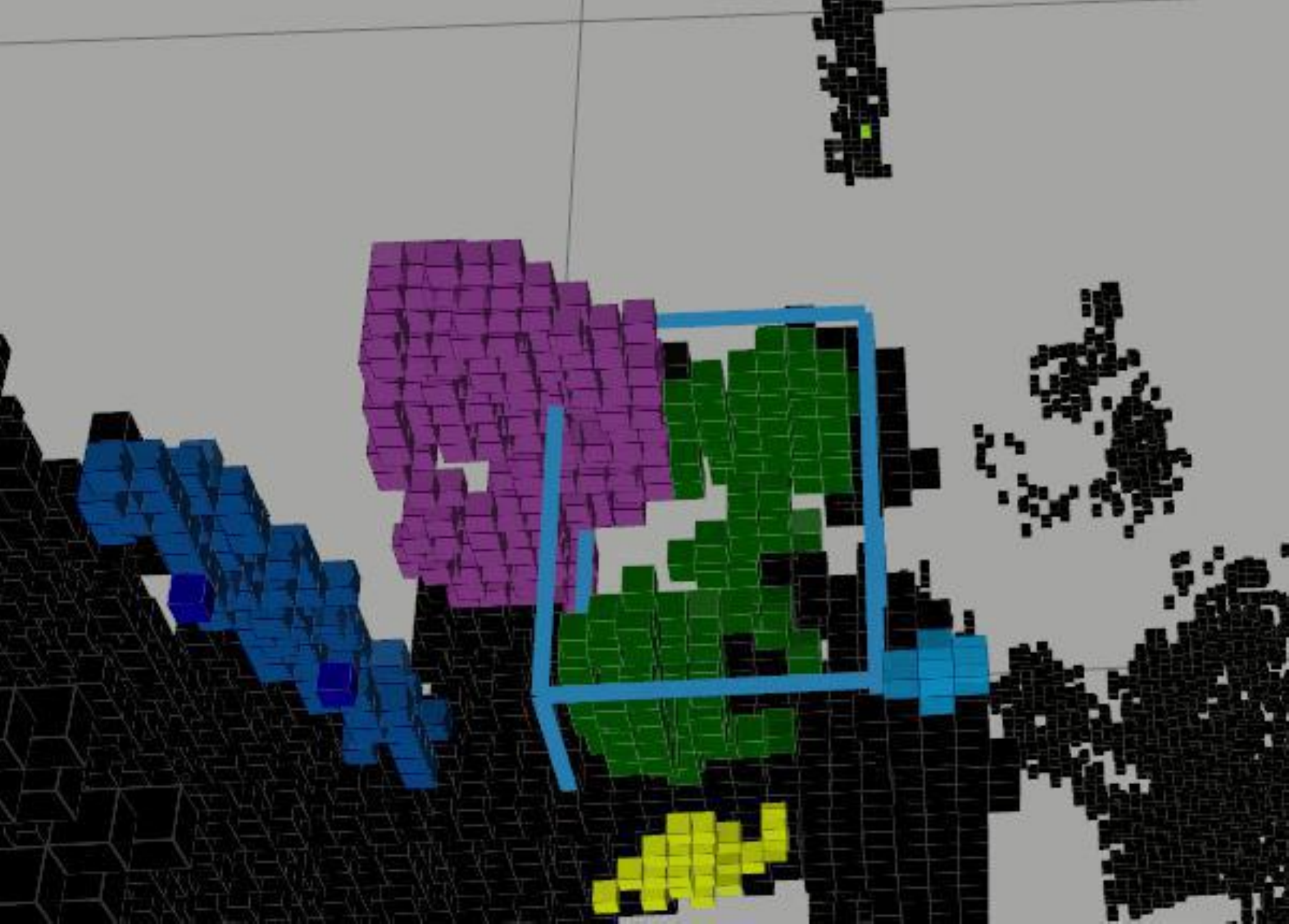} \\
  17          &                                         5.92 &                                         6.83 &                      \textbf{13.53} &  \tabimg{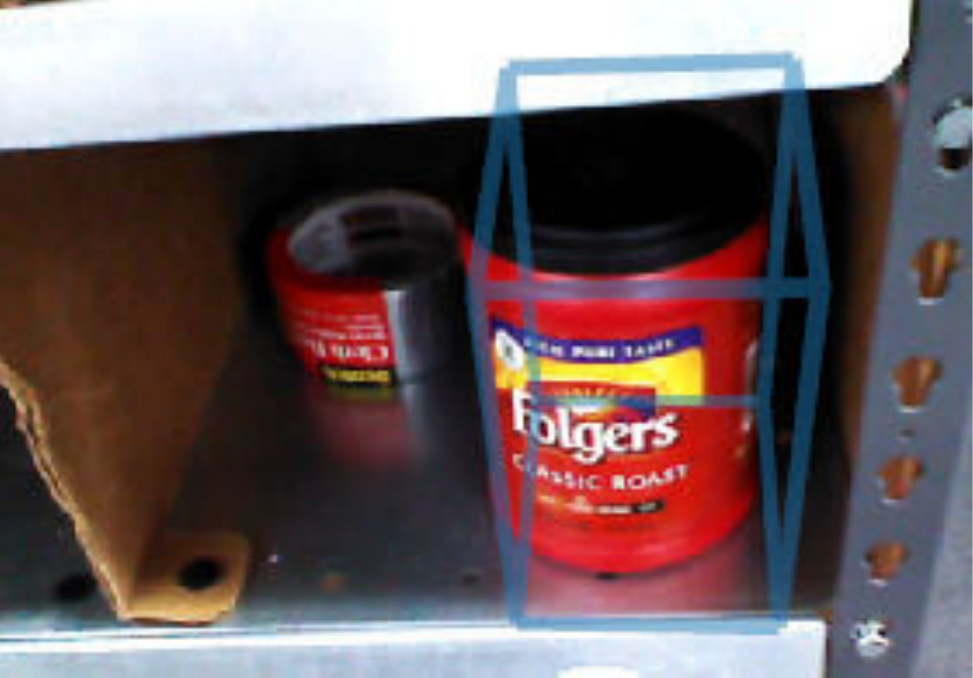} &  \tabimg{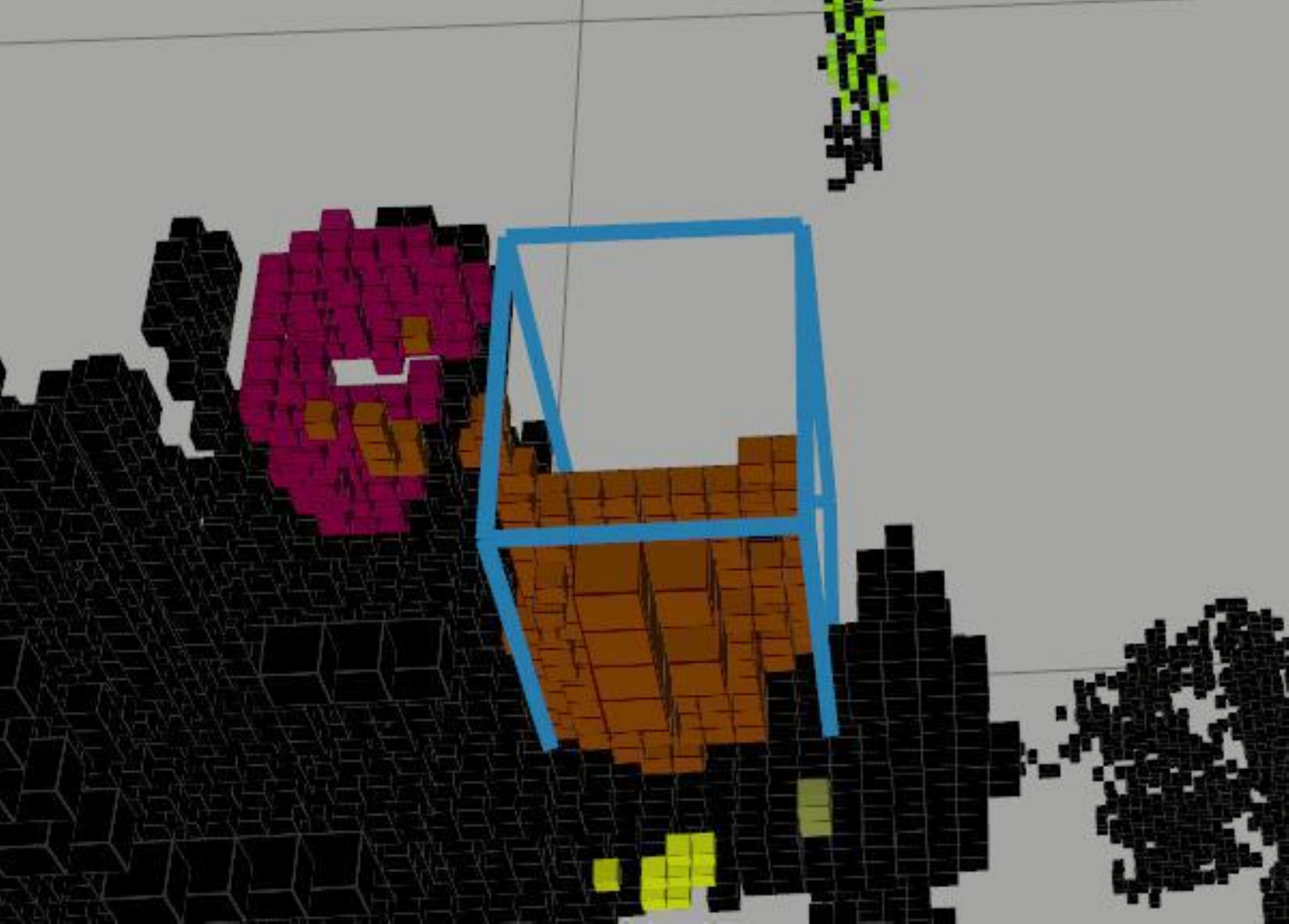} \\
  18          &                                         3.80 &                                         4.41 &                       \textbf{7.34} &  \tabimg{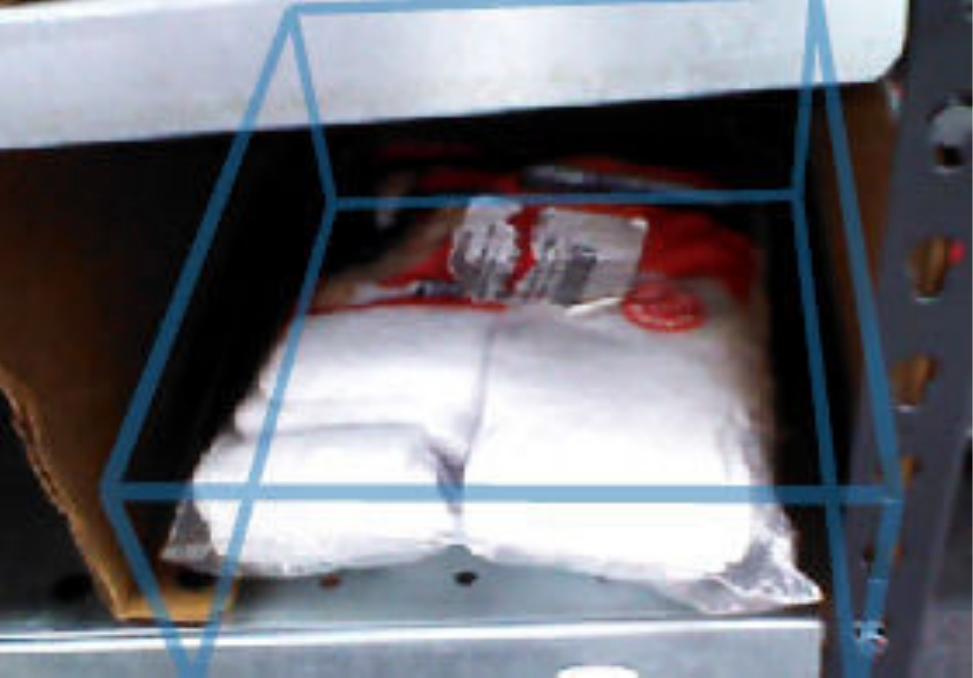} &  \tabimg{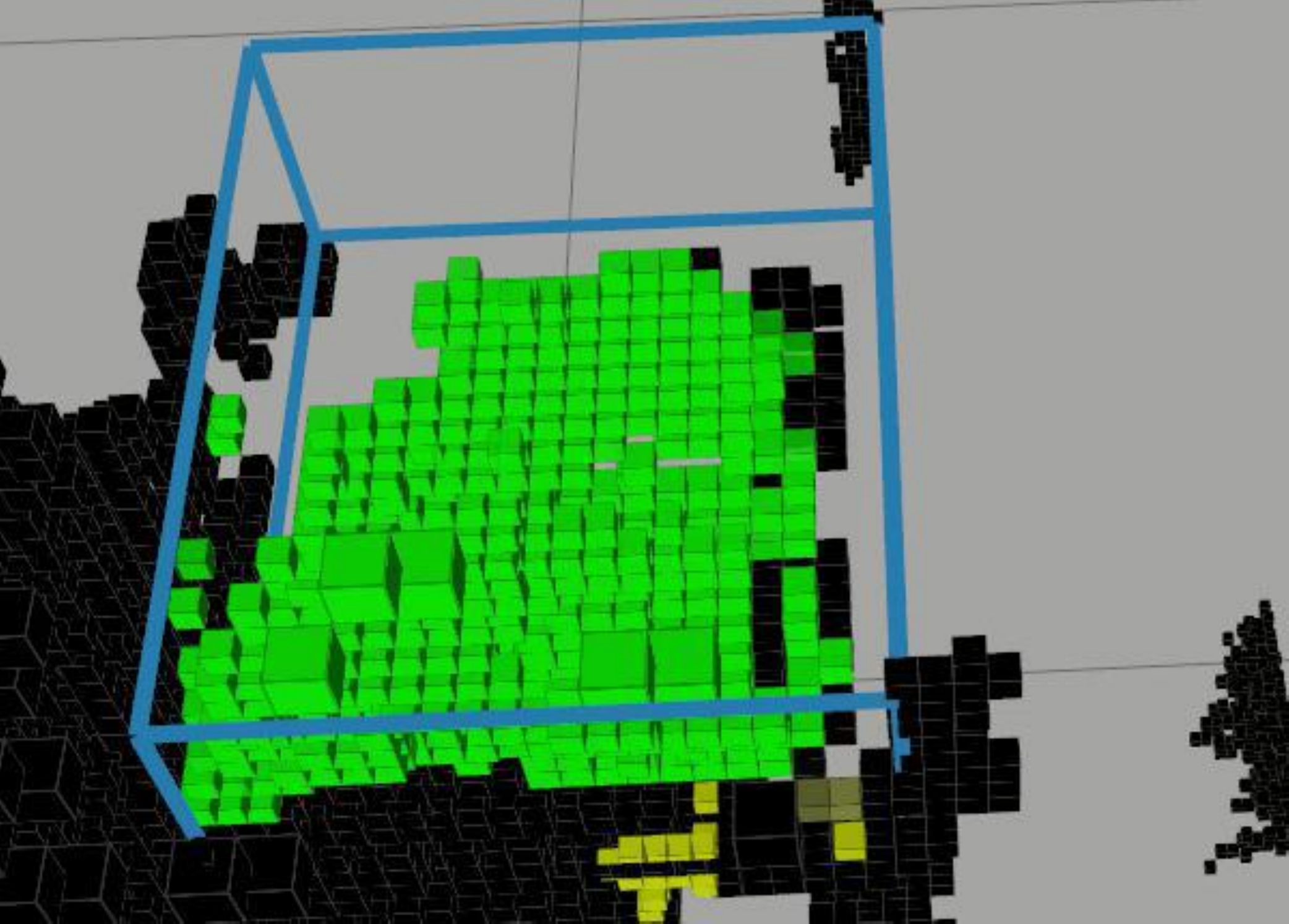} \\
  19          &                                         7.08 &                                         8.47 &                      \textbf{11.62} &  \tabimg{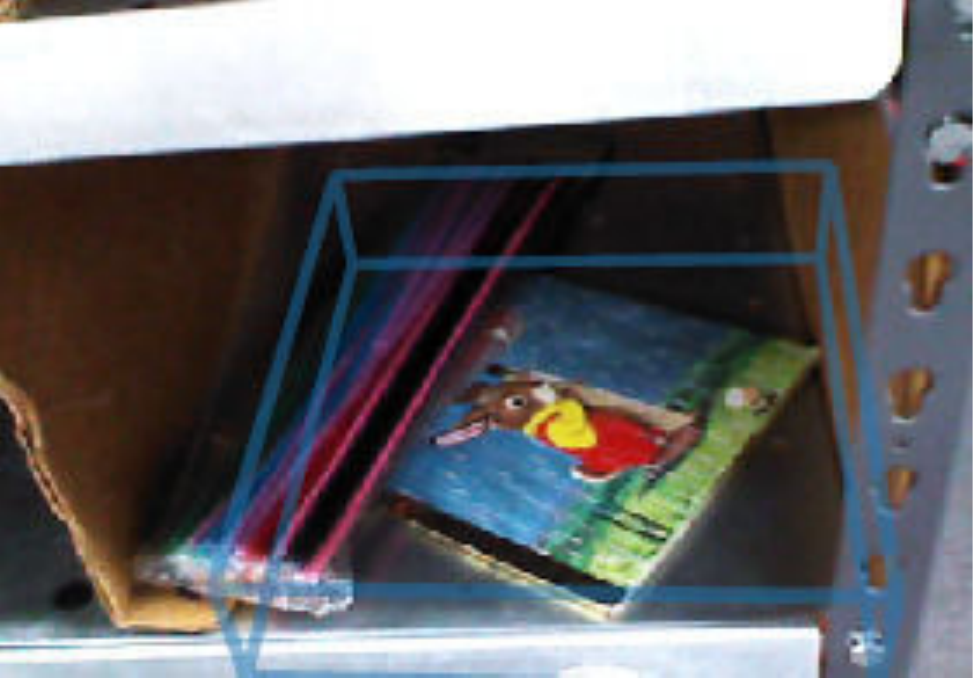} &  \tabimg{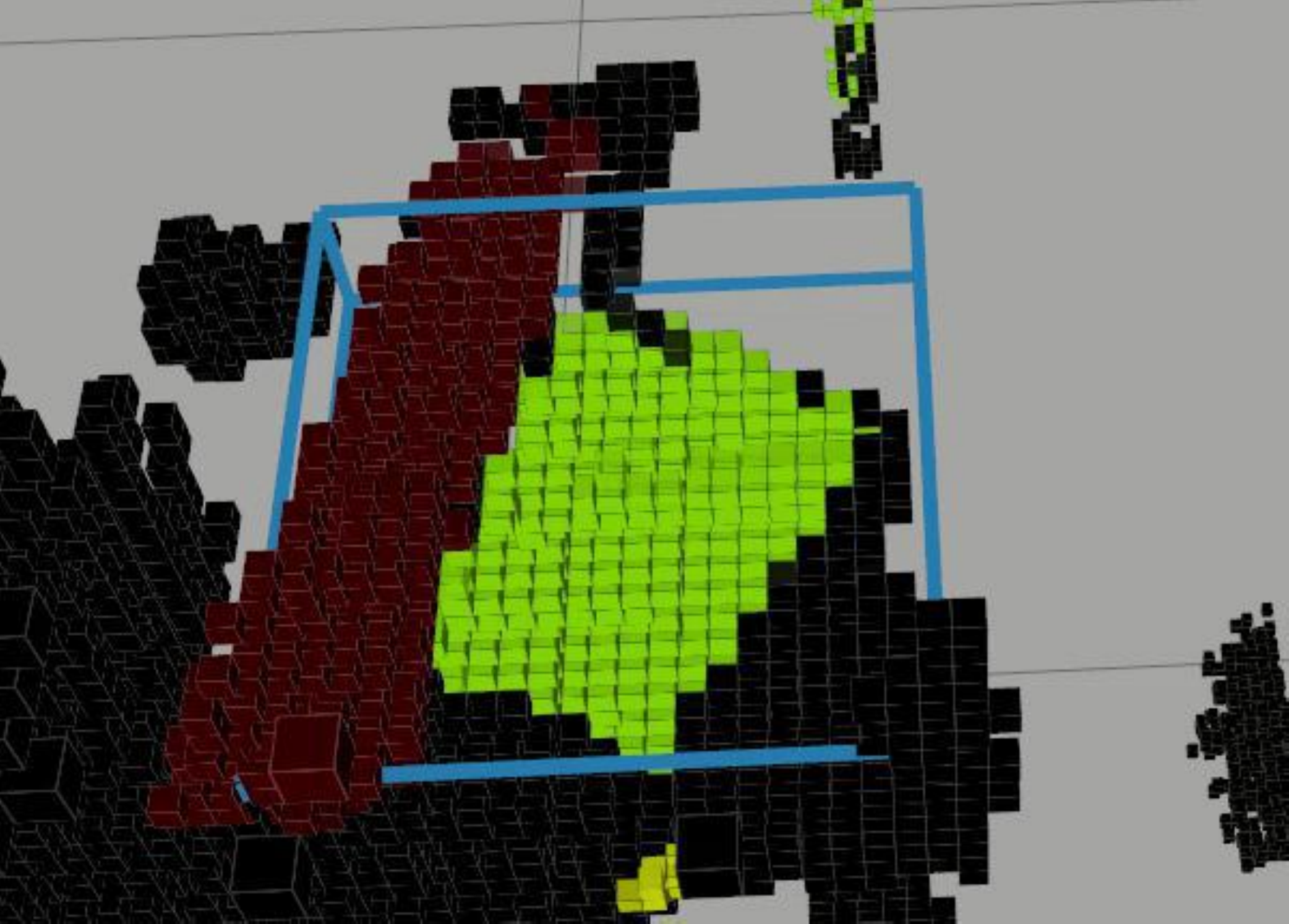} \\
  20          &                                         1.45 &                                         2.59 &                       \textbf{3.65} &  \tabimg{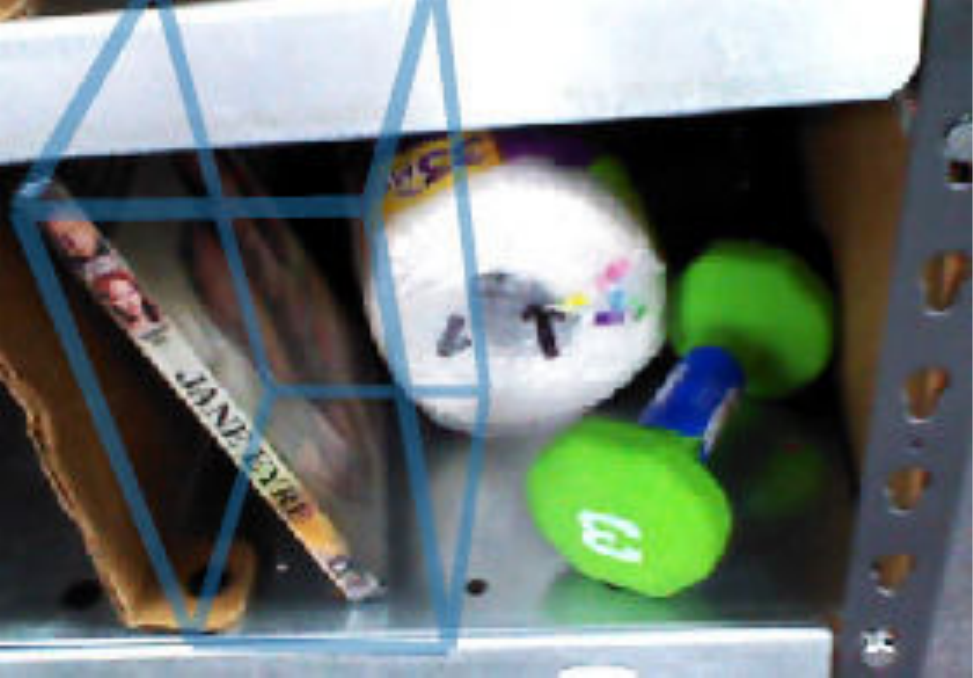} &  \tabimg{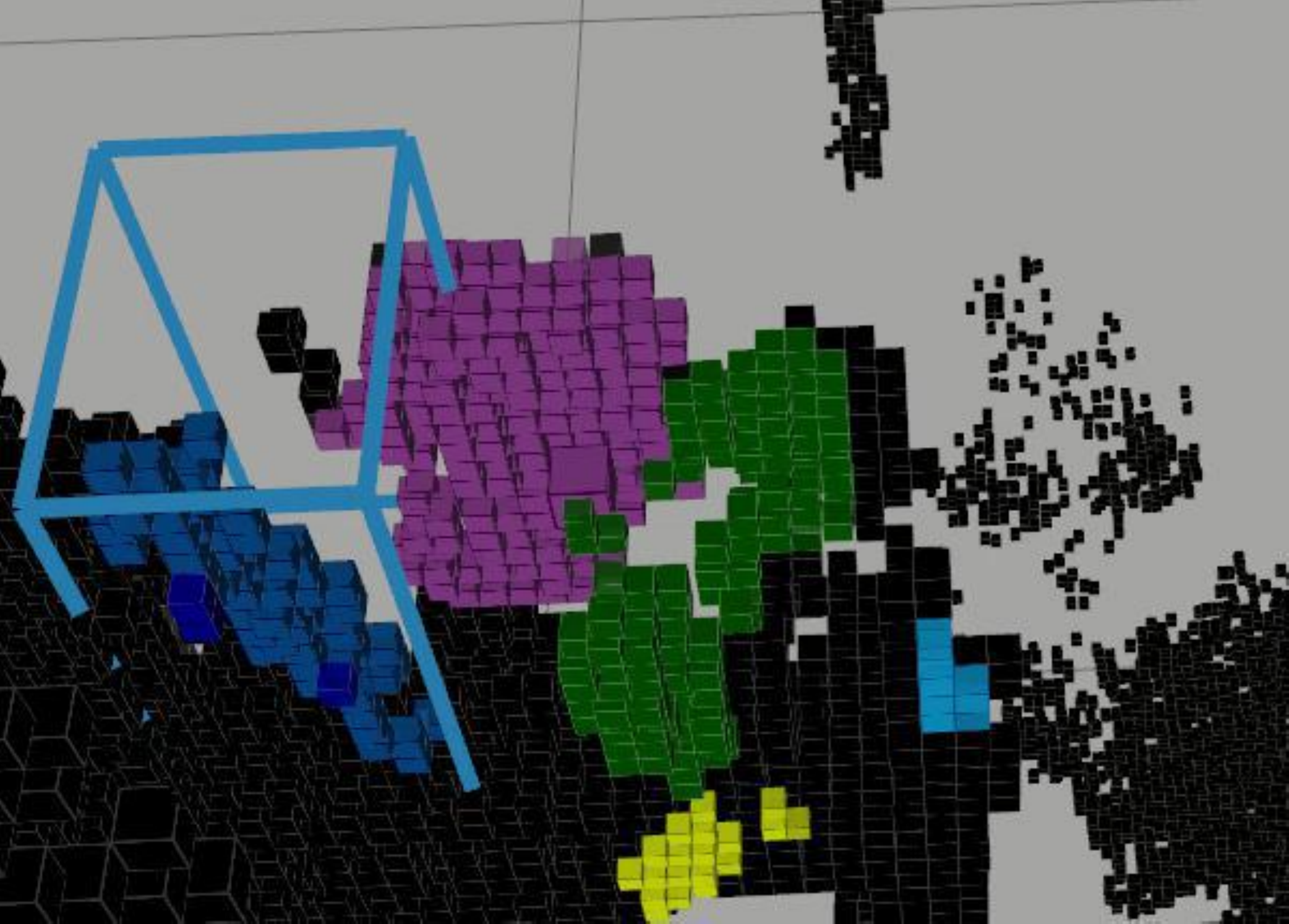} \\
  \end{tabular}
  \end{minipage}
  \begin{minipage}{0.49\textwidth}
  \centering
  \begin{tabular}{c|rra|cc}
  {}          & \multicolumn{3}{c}{$IU_{3d}$}                                                                                                     &   \multirow{2}{*}{Scene}           &   \multirow{2}{*}{\thead{Result \\ (LabelOctoMap)}}           \\
  {}          &          \thead{Projec- \\ tion \\ (Mean)} &           \thead{Projec- \\ tion \\ (Max)} &      \thead{Label- \\ Octo- \\ Map} &                                                                          \\
  label       &                                              &                                              &                                     &                                    &                                     \\
  \hline
  21          &                                         1.58 &                                         2.09 &                       \textbf{3.78} &  \tabimg{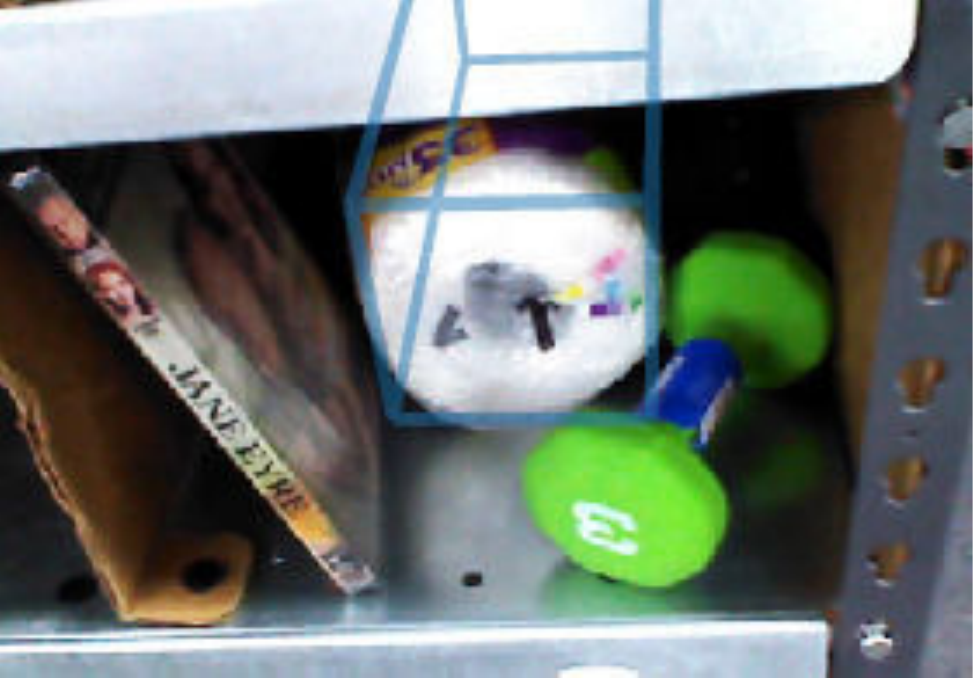} &  \tabimg{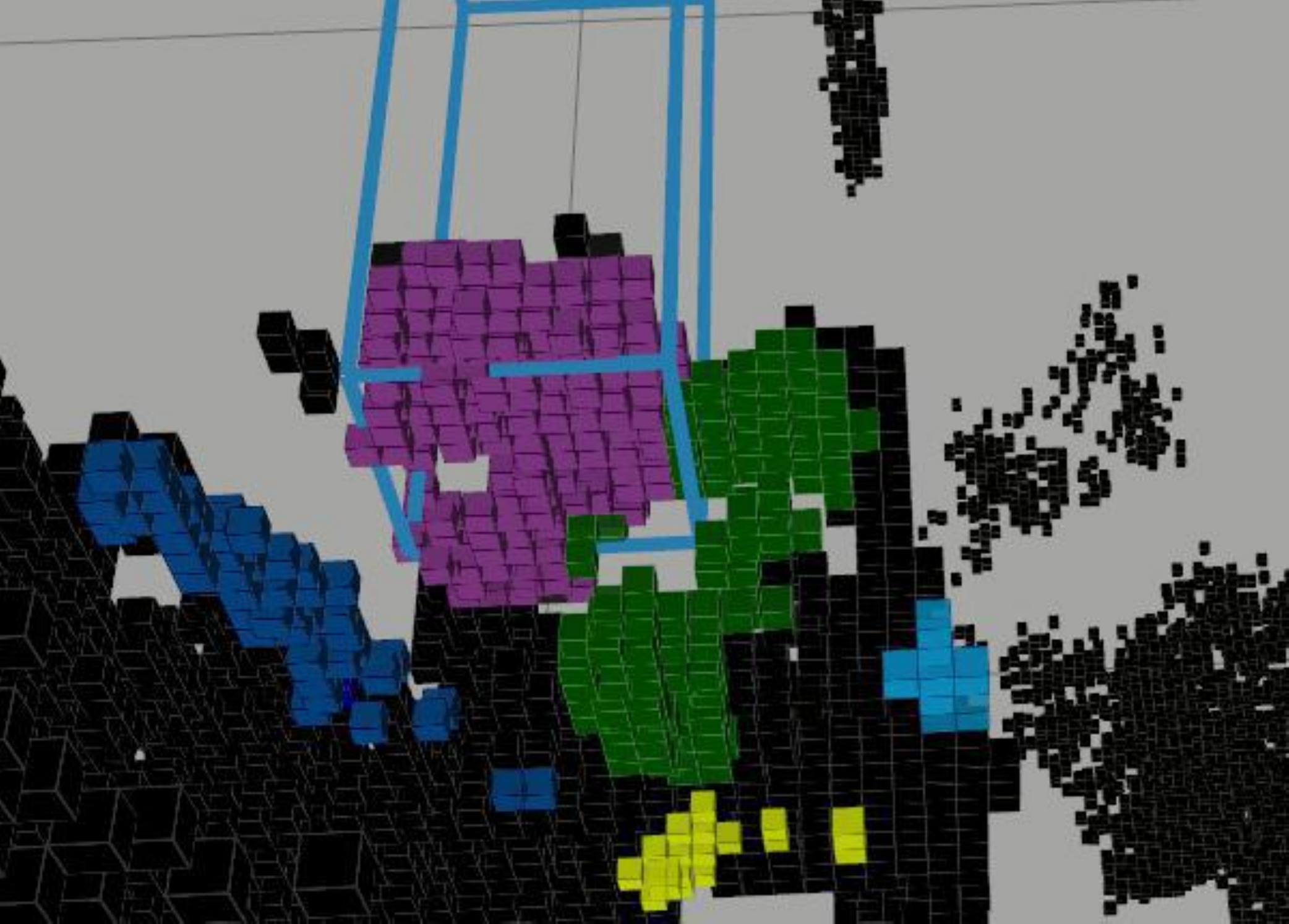} \\
  22          &                                         4.40 &                                         5.25 &                       \textbf{9.28} &  \tabimg{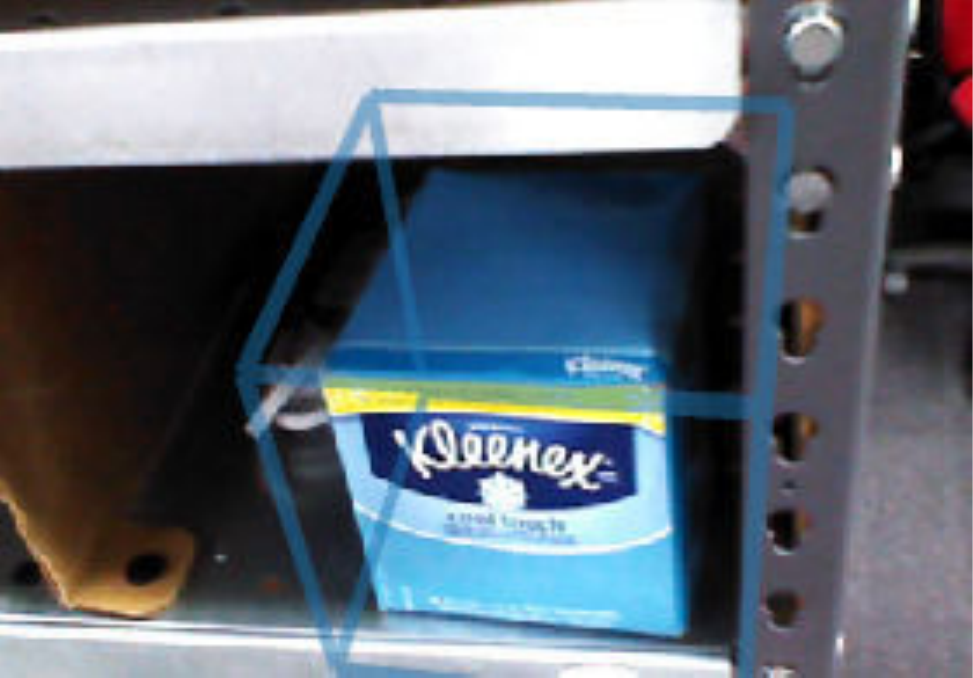} &  \tabimg{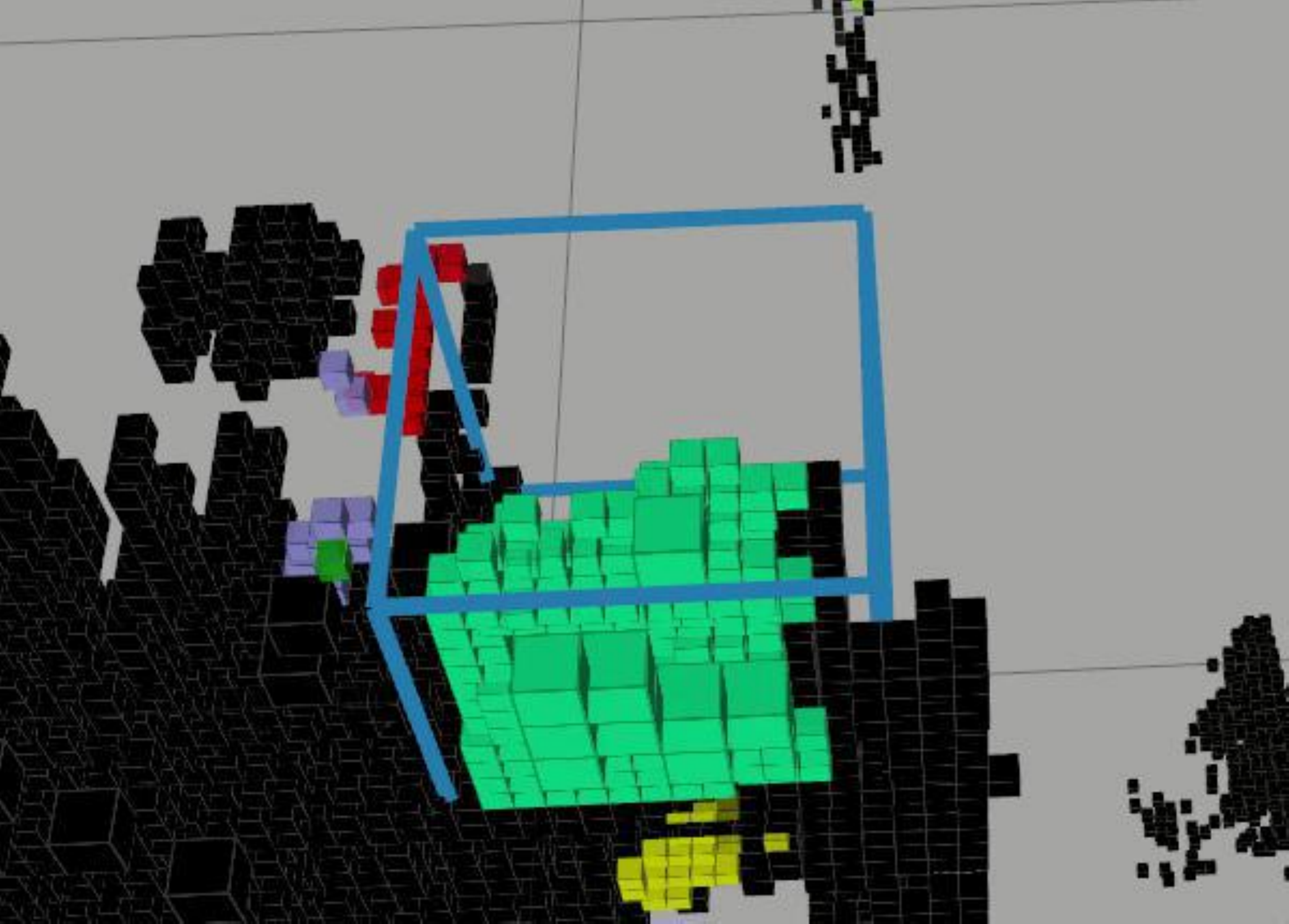} \\
  23          &                                         4.53 &                                         5.28 &                       \textbf{9.95} &  \tabimg{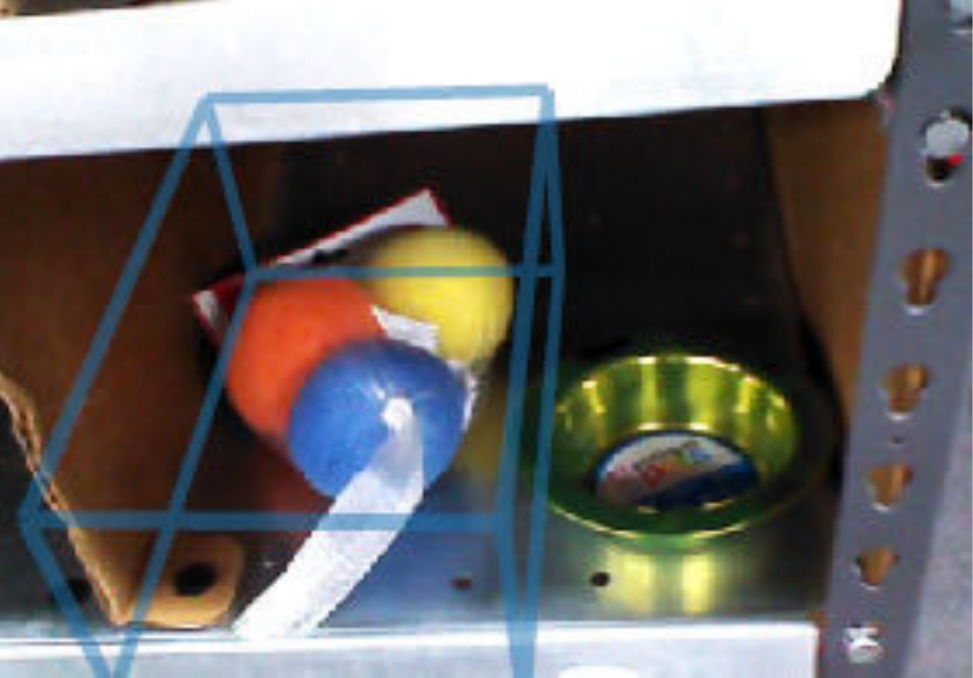} &  \tabimg{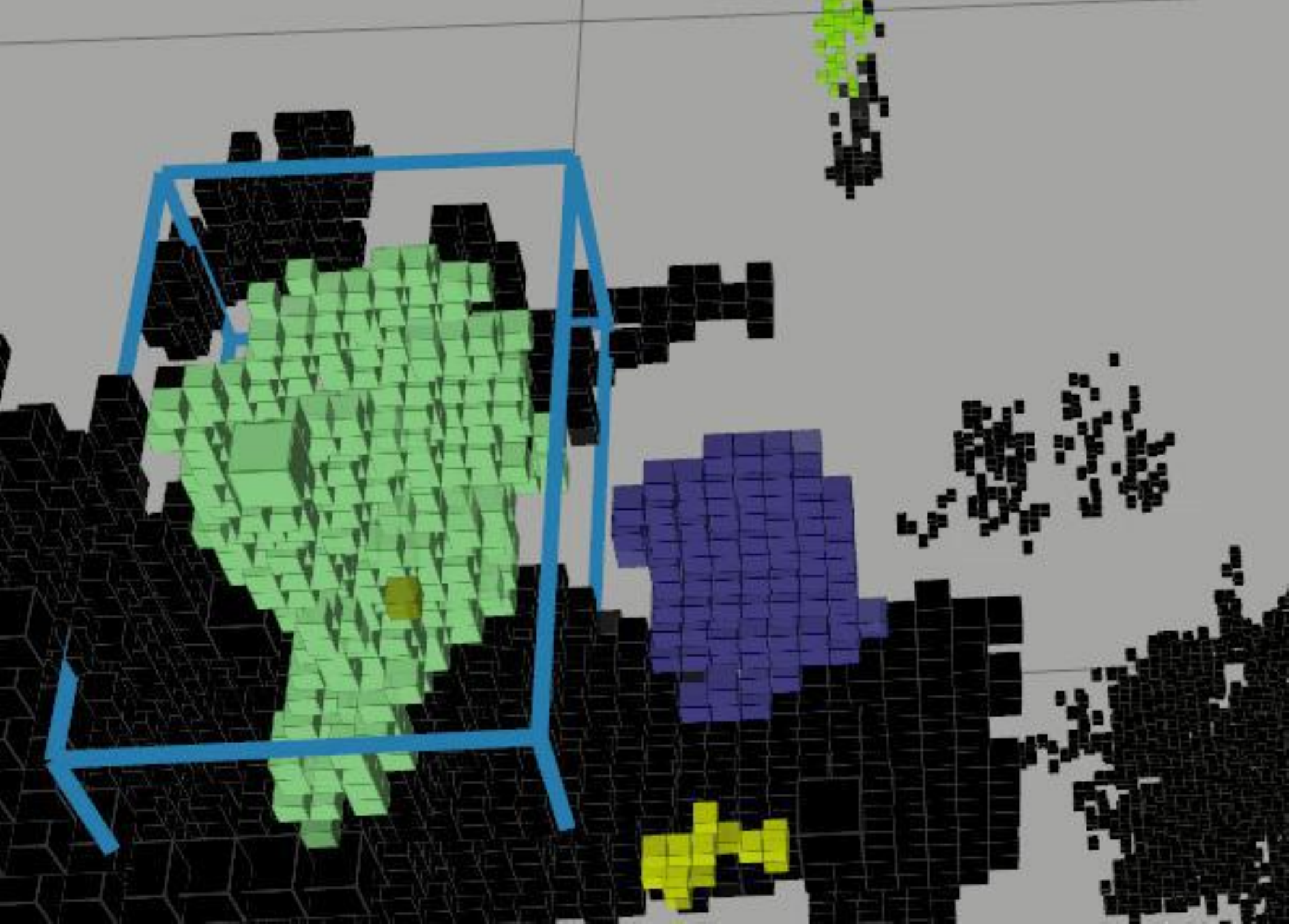} \\
  24          &                                        16.47 &                                        19.92 &                      \textbf{31.43} &  \tabimg{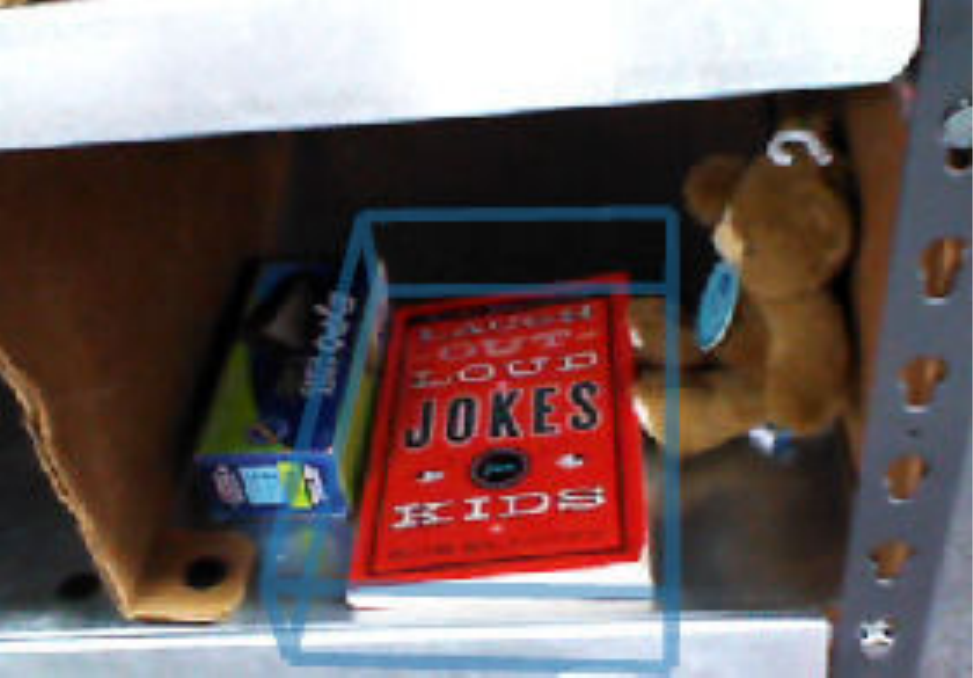} &  \tabimg{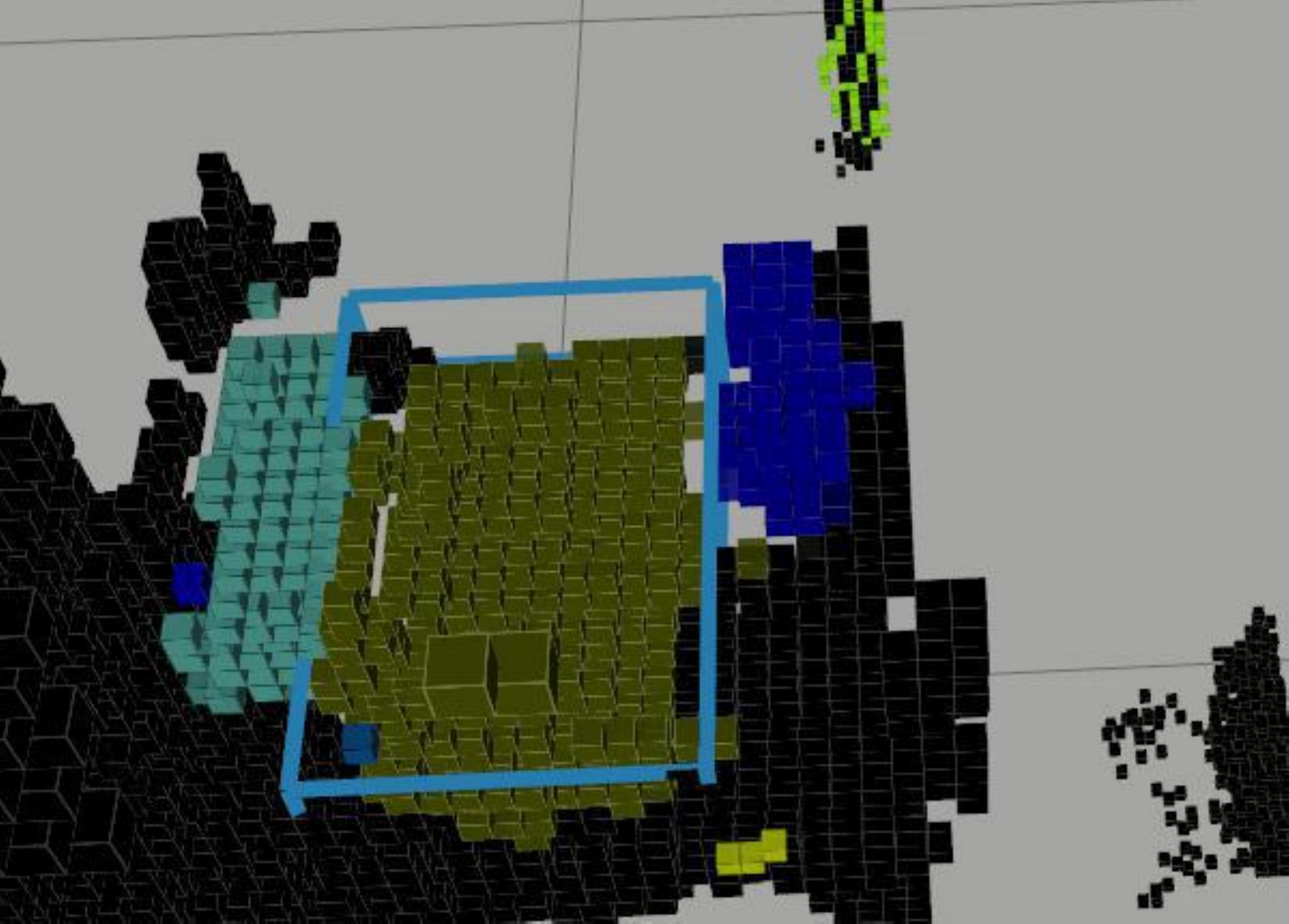} \\
  25          &                                         1.14 &                                \textbf{4.44} &                                0.59 &  \tabimg{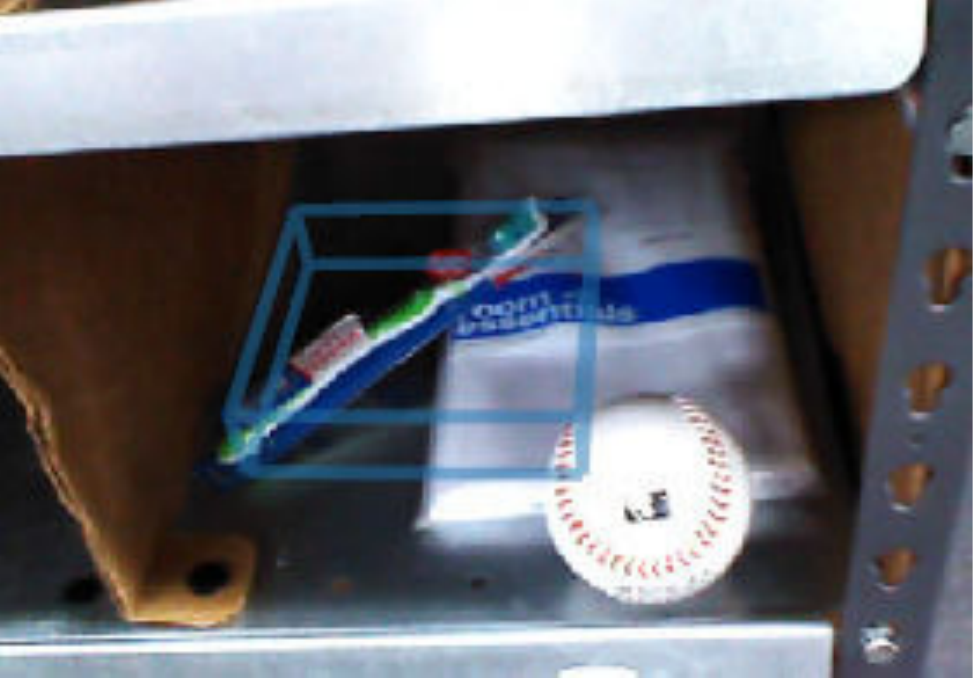} &  \tabimg{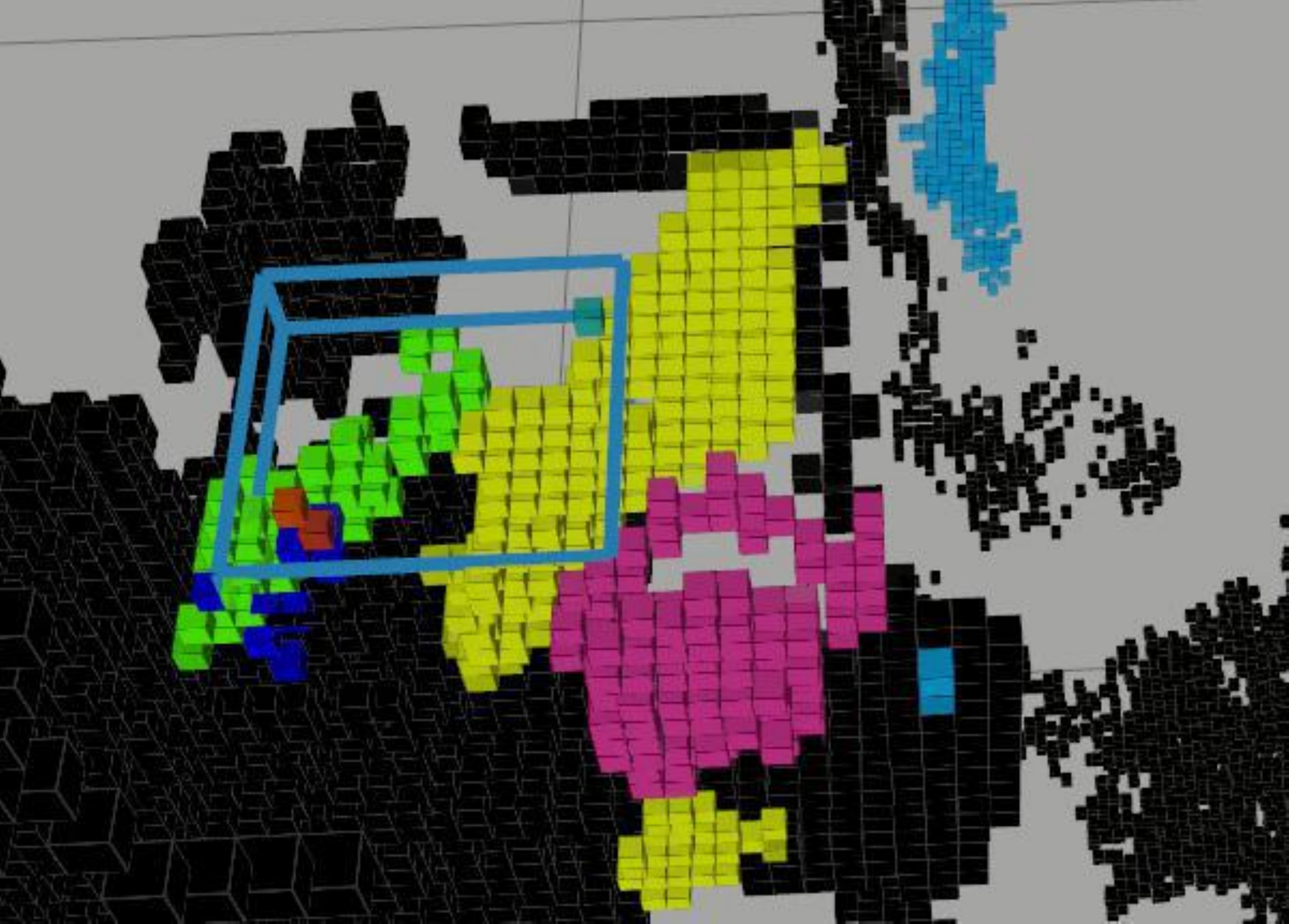} \\
  26          &                                         9.32 &                               \textbf{15.42} &                               11.61 &  \tabimg{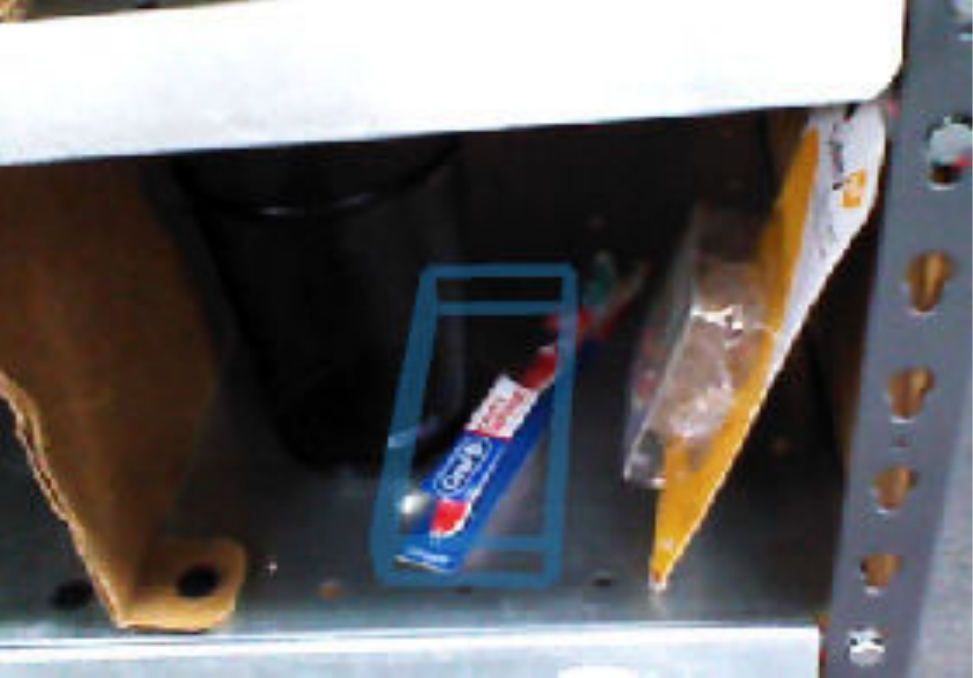} &  \tabimg{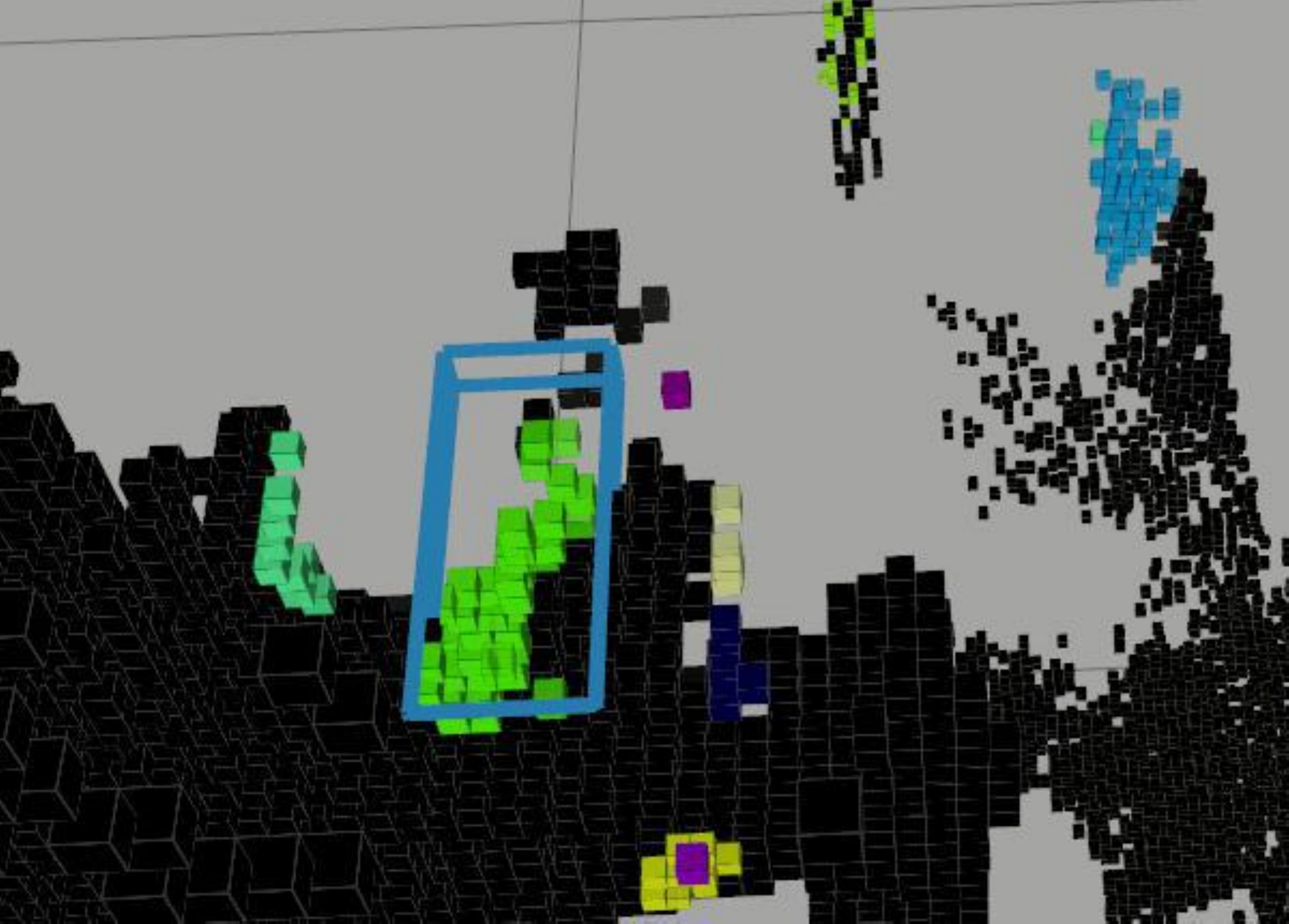} \\
  27          &                                         4.69 &                            \underline{10.86} &                   \underline{10.79} &  \tabimg{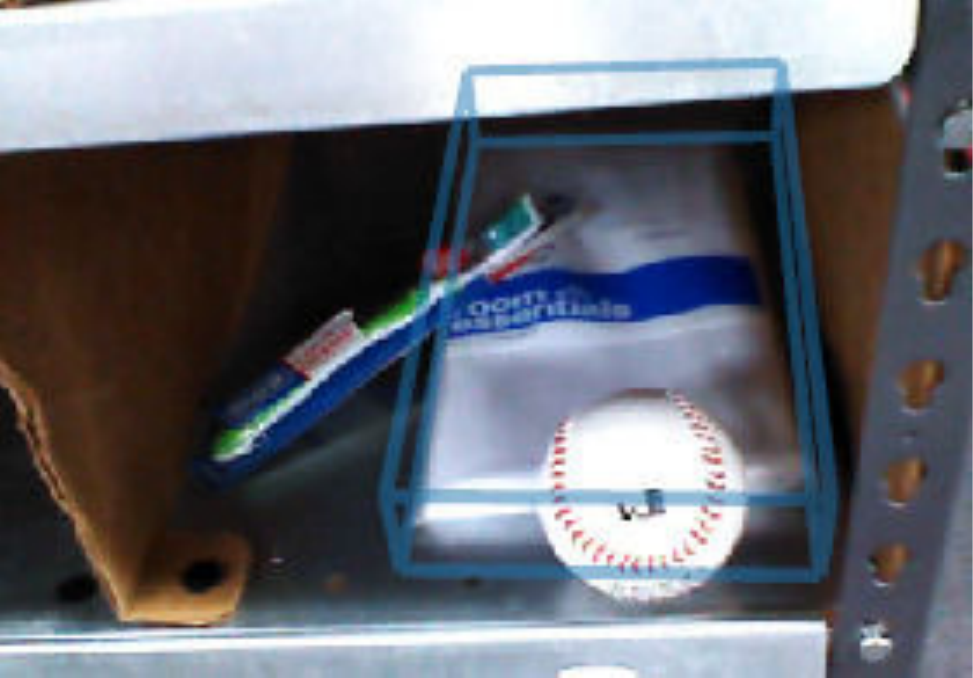} &  \tabimg{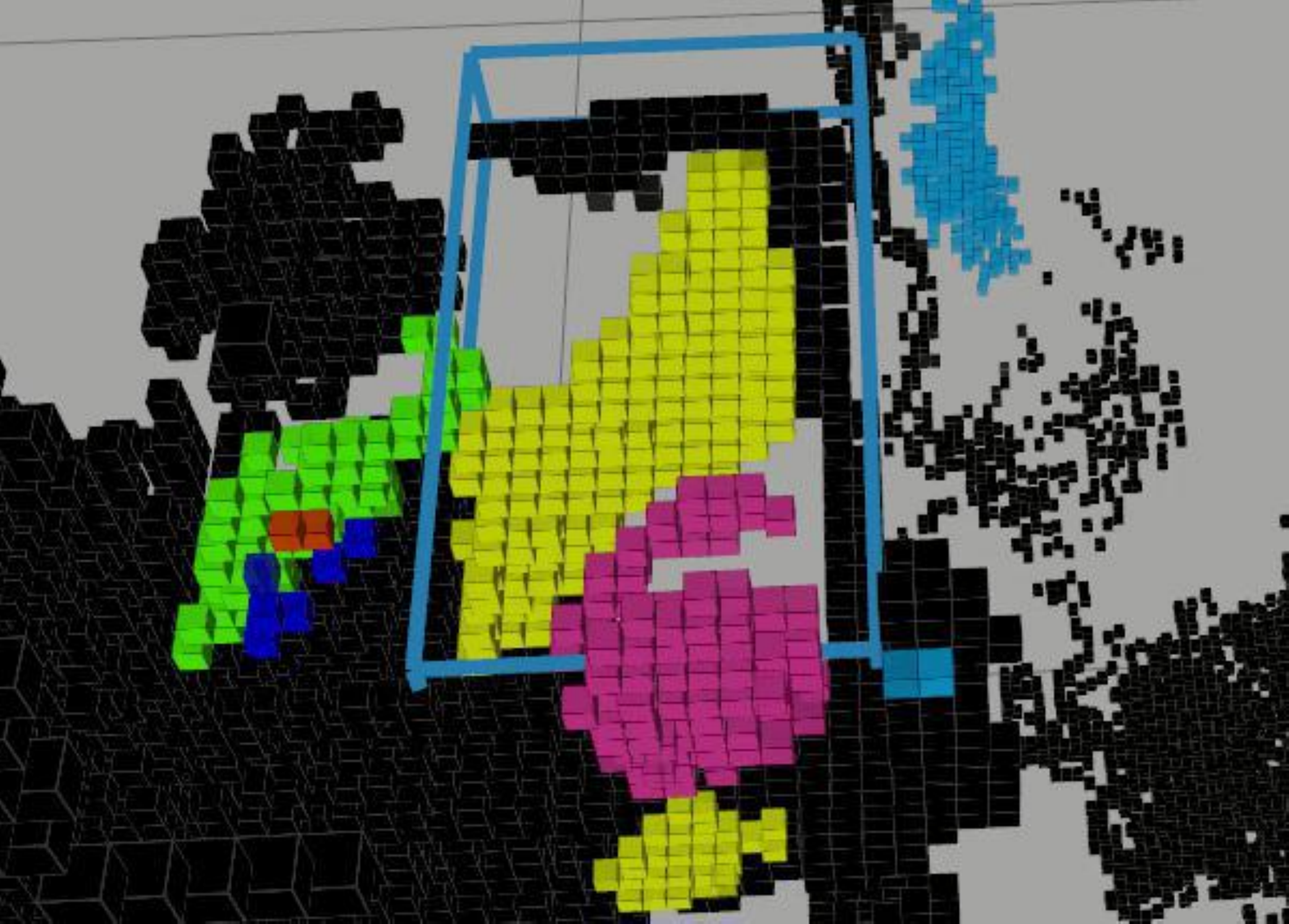} \\
  28          &                                         4.12 &                                         5.33 &                       \textbf{9.16} &  \tabimg{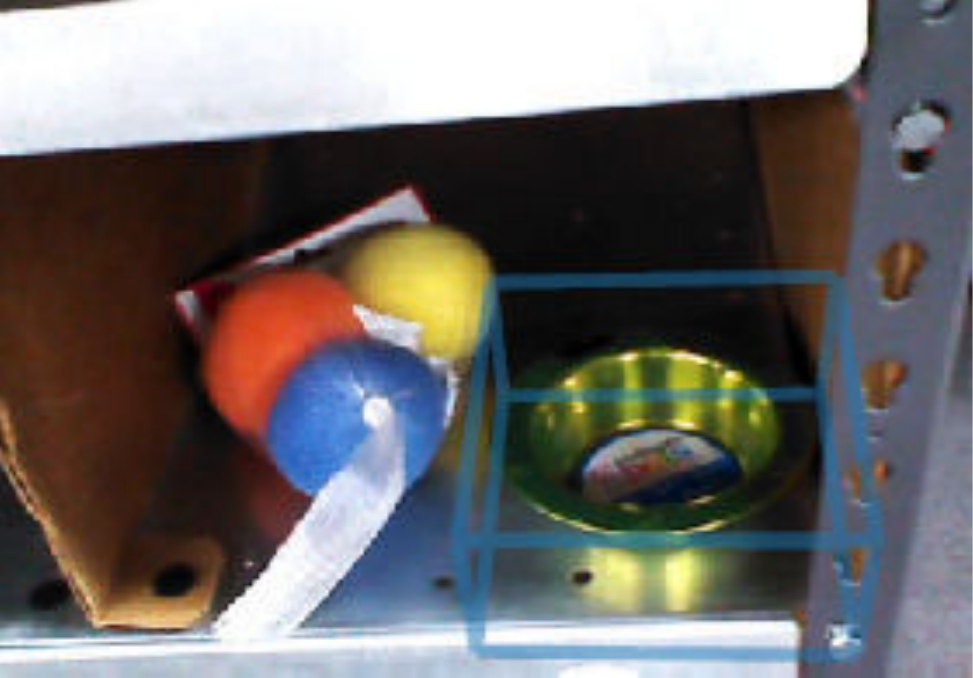} &  \tabimg{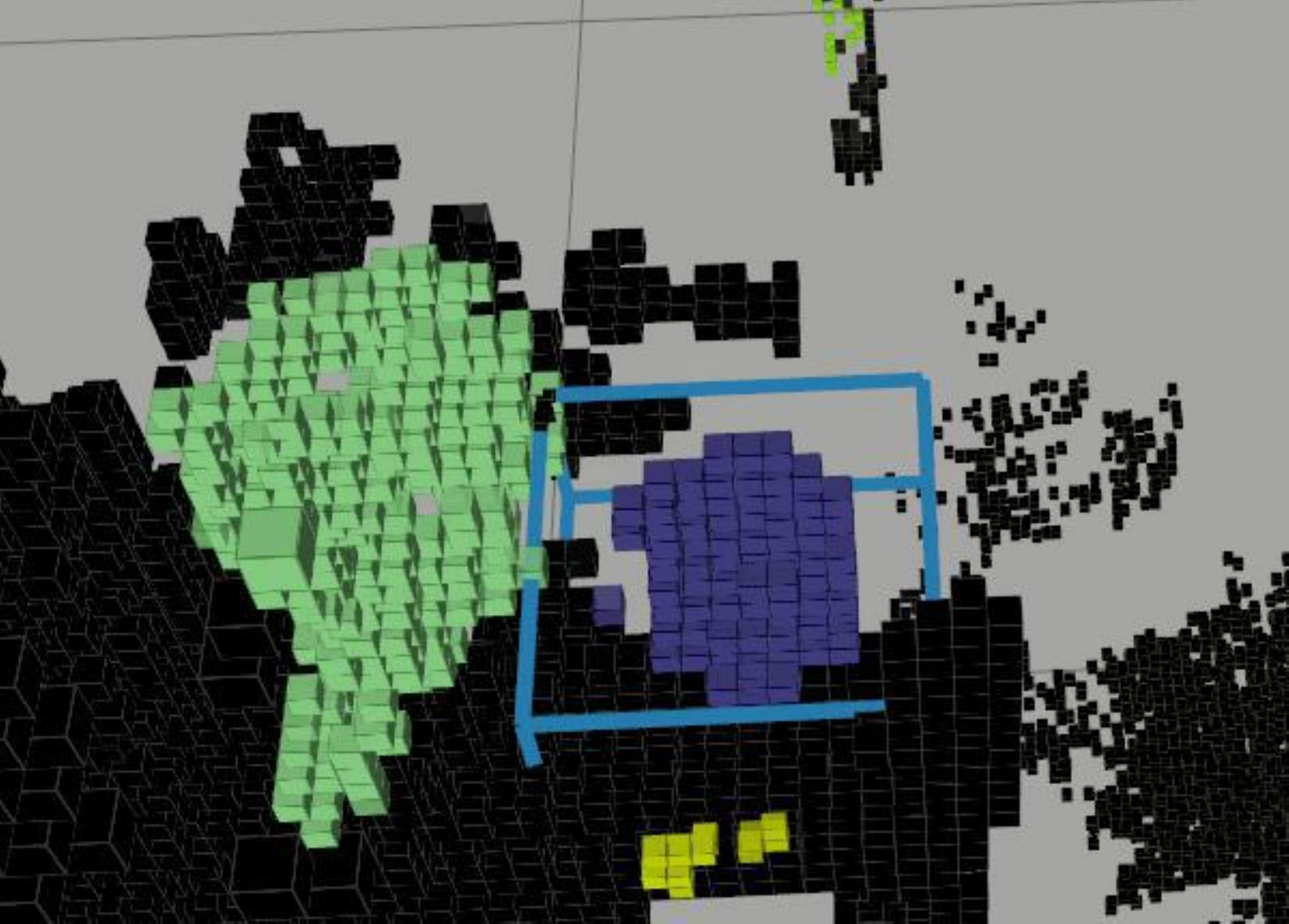} \\
  29          &                                        11.15 &                                        13.32 &                      \textbf{17.06} &  \tabimg{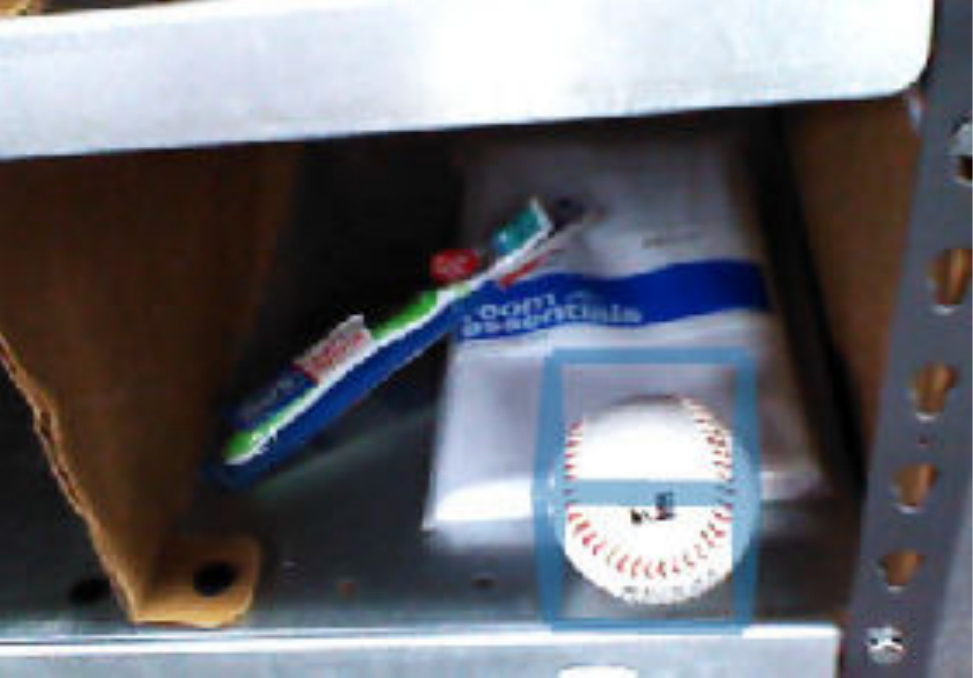} &  \tabimg{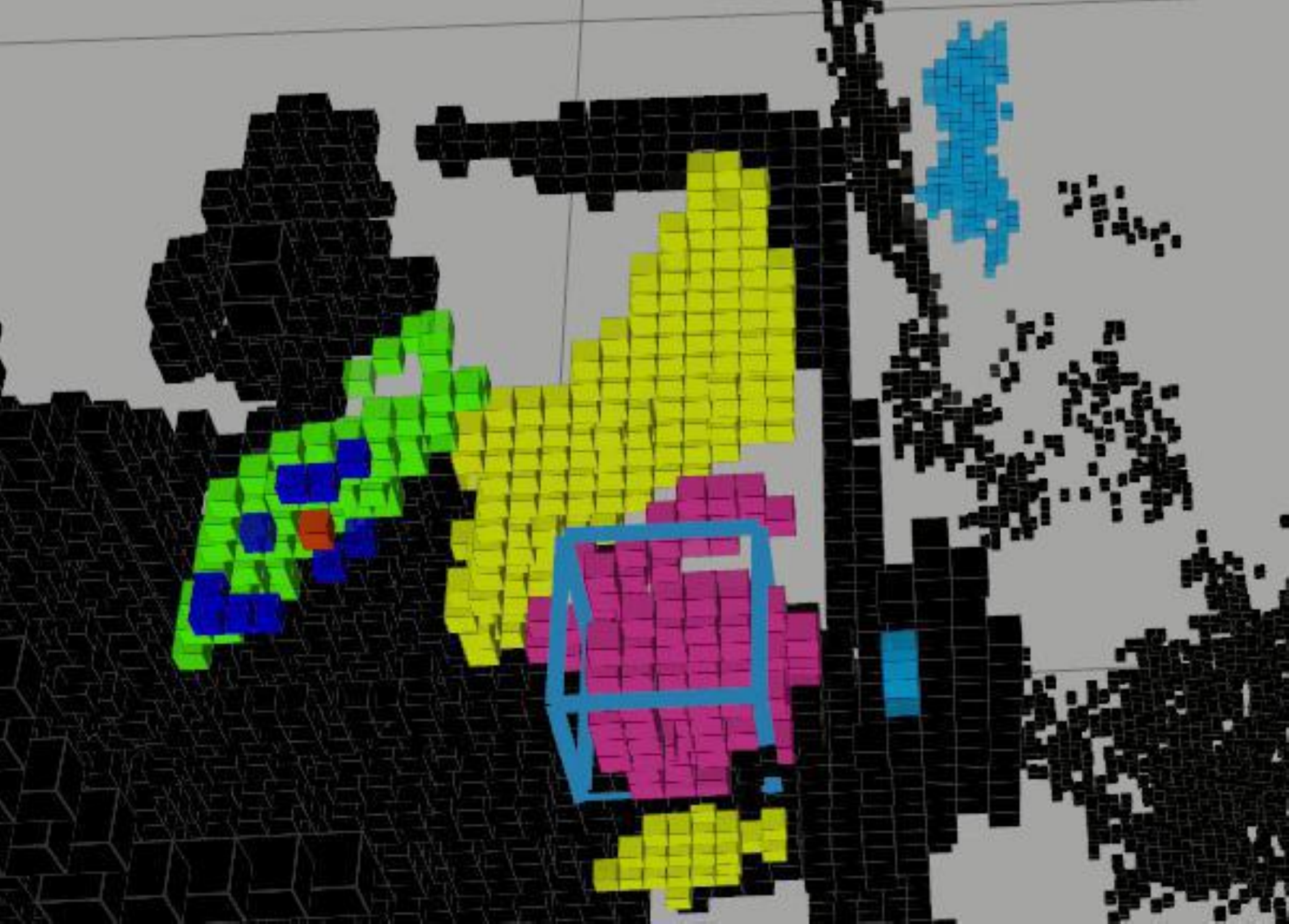} \\
  30          &                                         0.01 &                                         0.10 &                                0.02 &  \tabimg{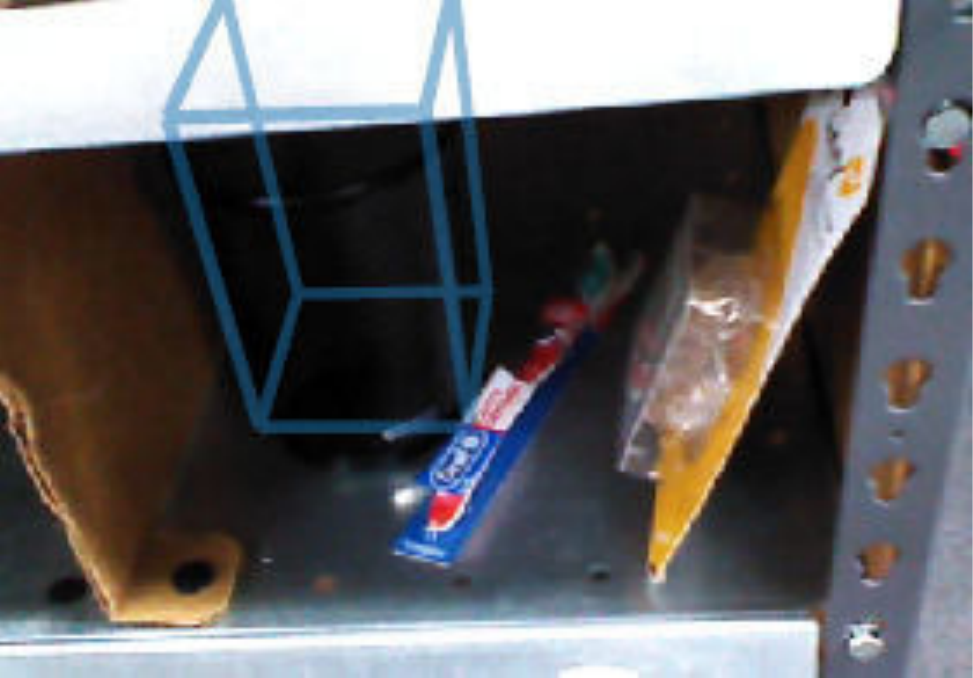} &  \tabimg{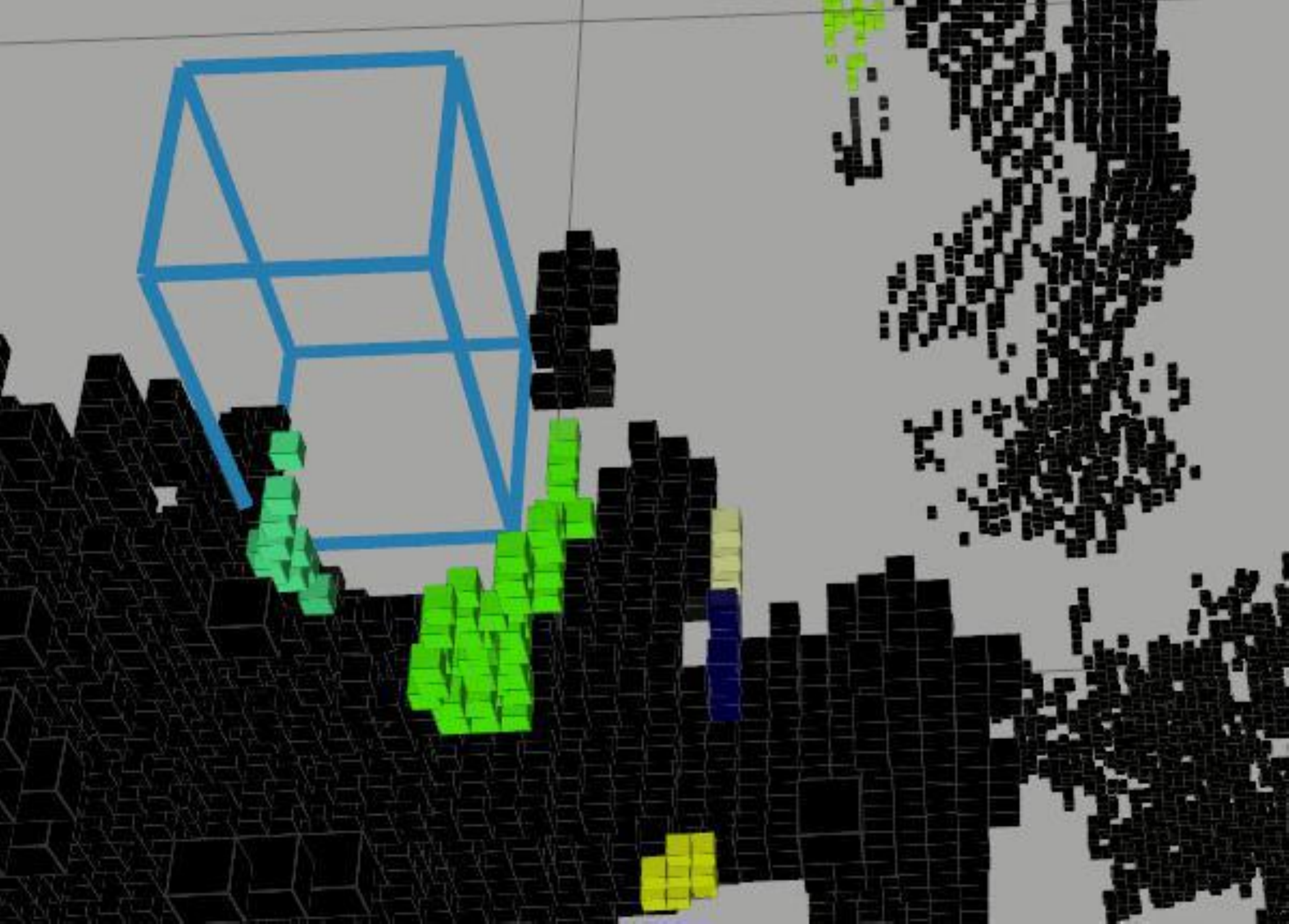} \\
  31          &                                         0.02 &                                         0.19 &                                0.00 &  \tabimg{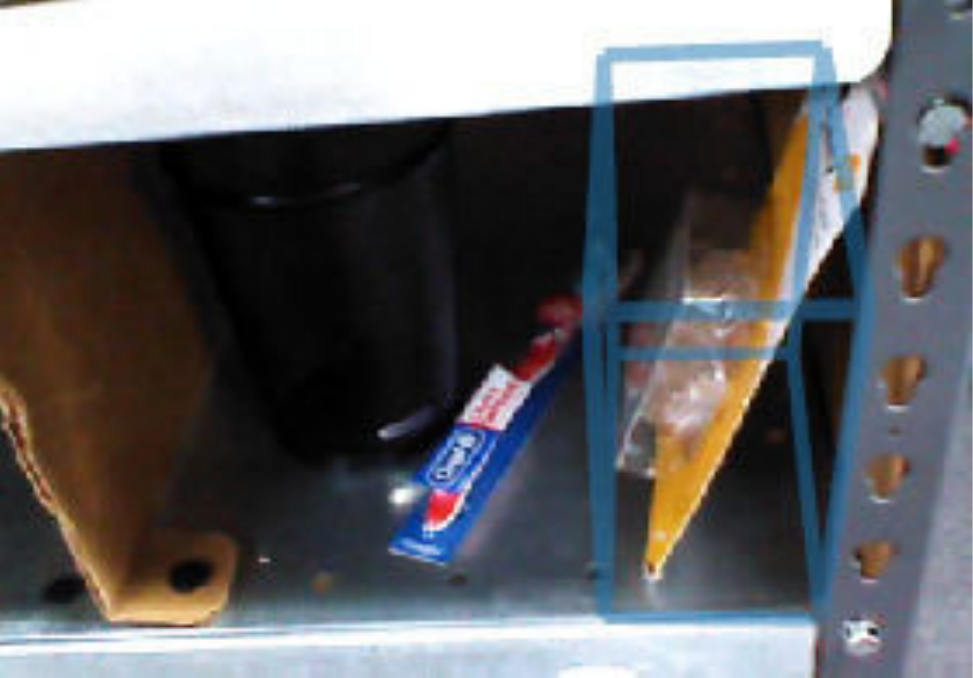} &  \tabimg{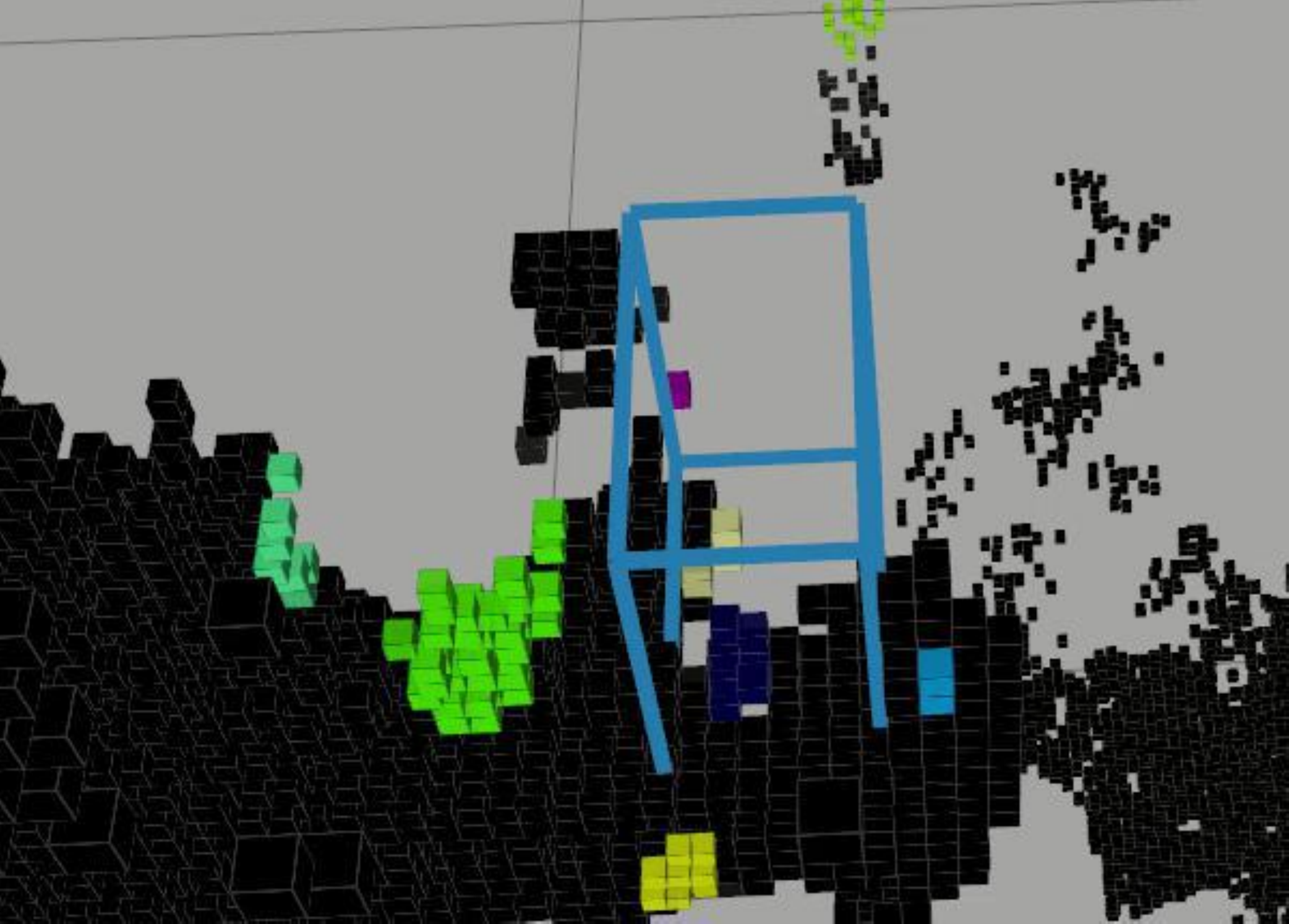} \\
  32          &                                        11.20 &                                        12.53 &                      \textbf{19.25} &  \tabimg{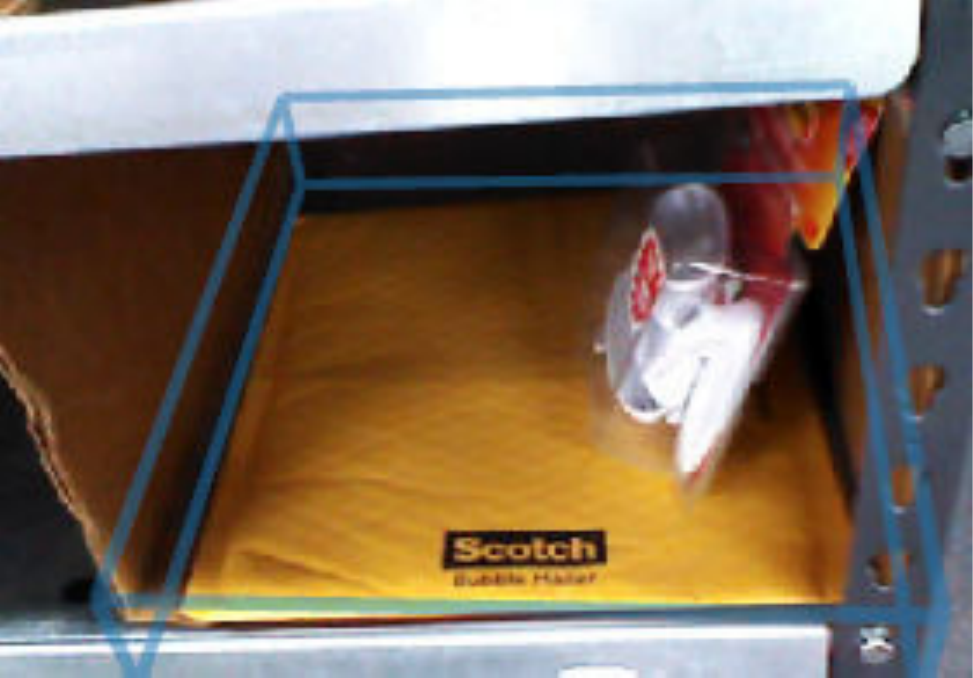} &  \tabimg{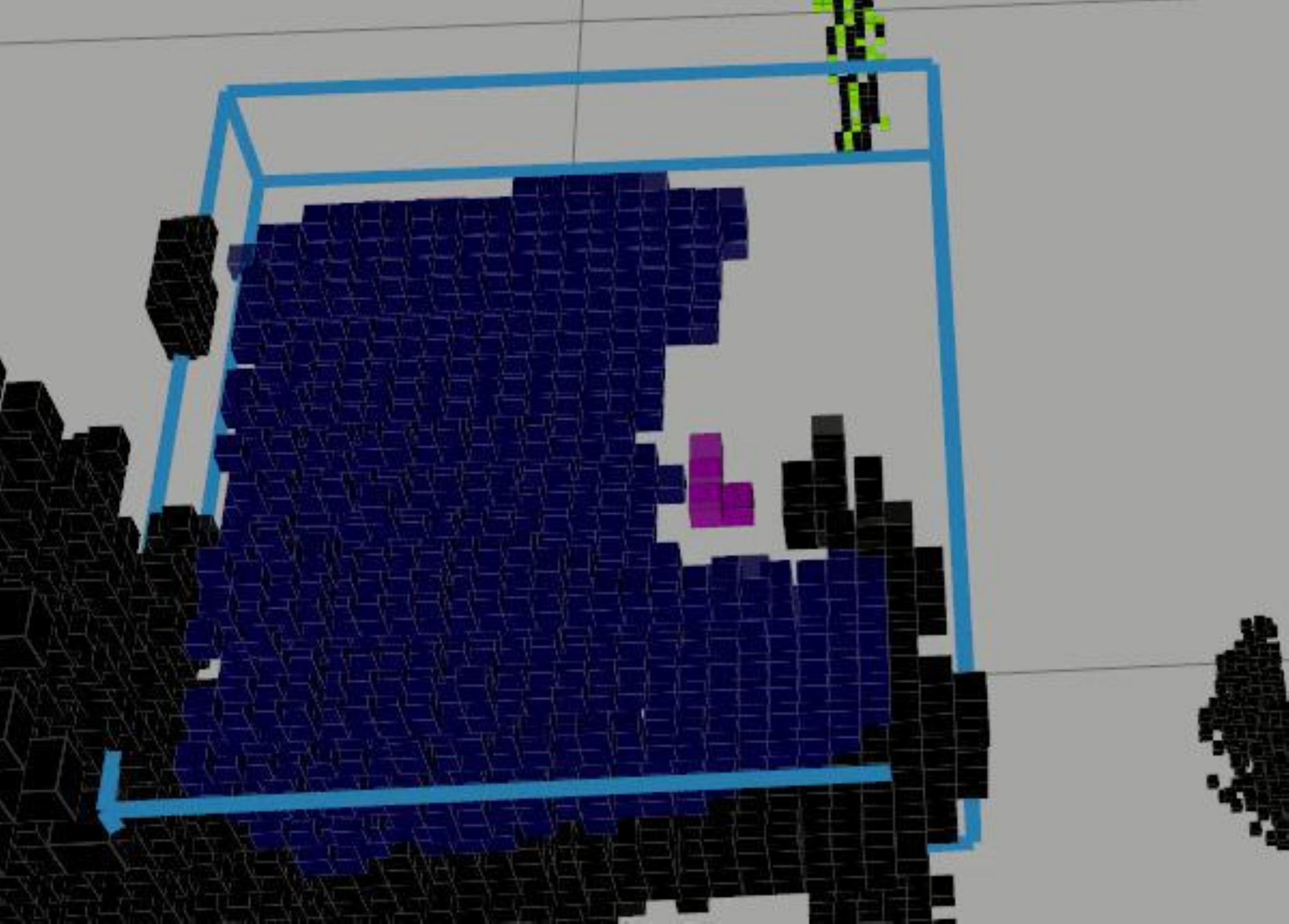} \\
  33          &                                         6.62 &                                        12.54 &                      \textbf{15.74} &  \tabimg{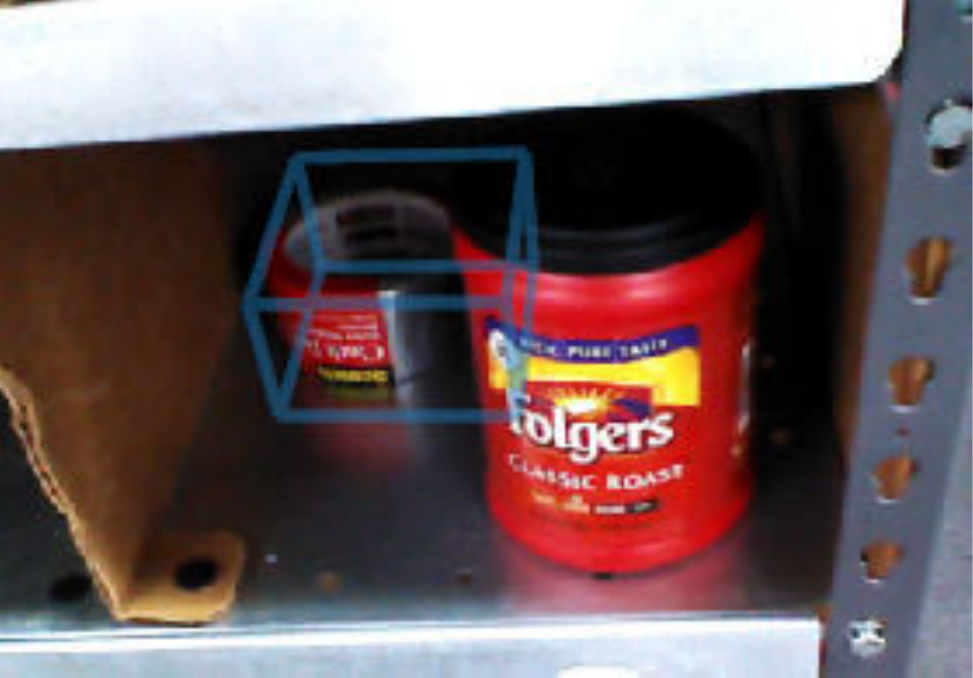} &  \tabimg{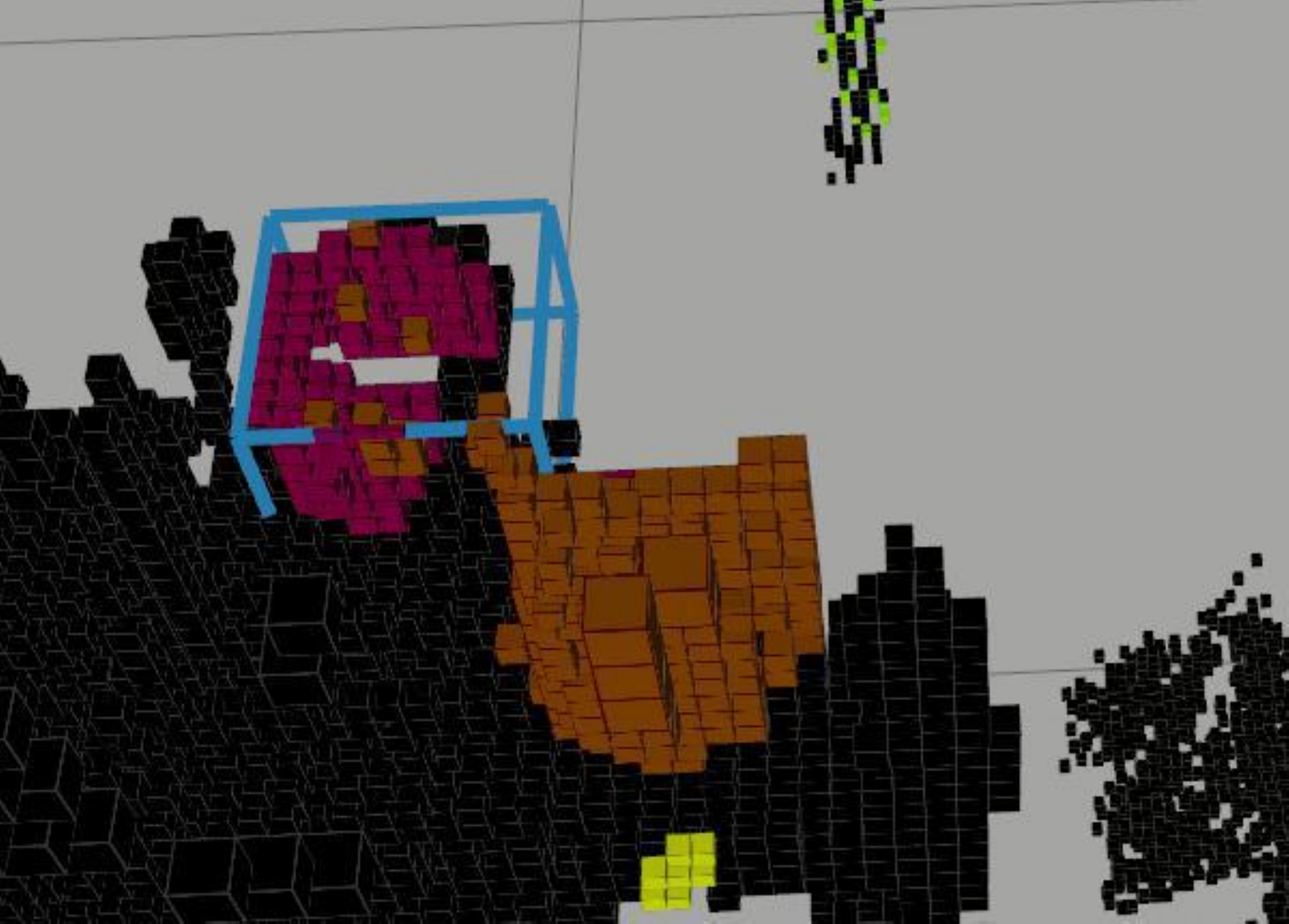} \\
  34          &                                         3.72 &                             \underline{5.62} &                    \underline{5.50} &  \tabimg{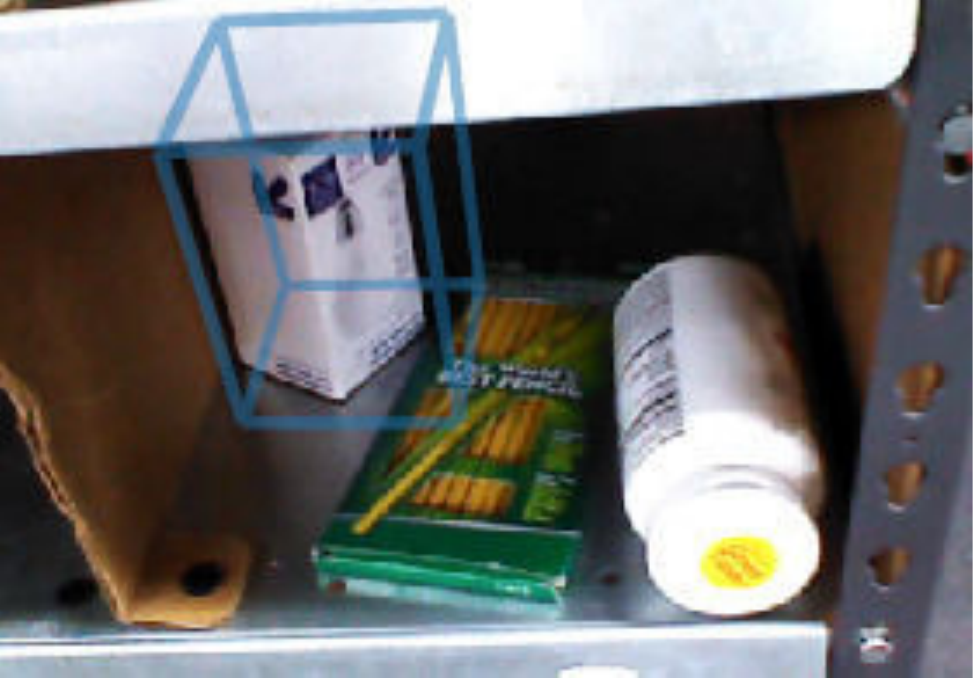} &  \tabimg{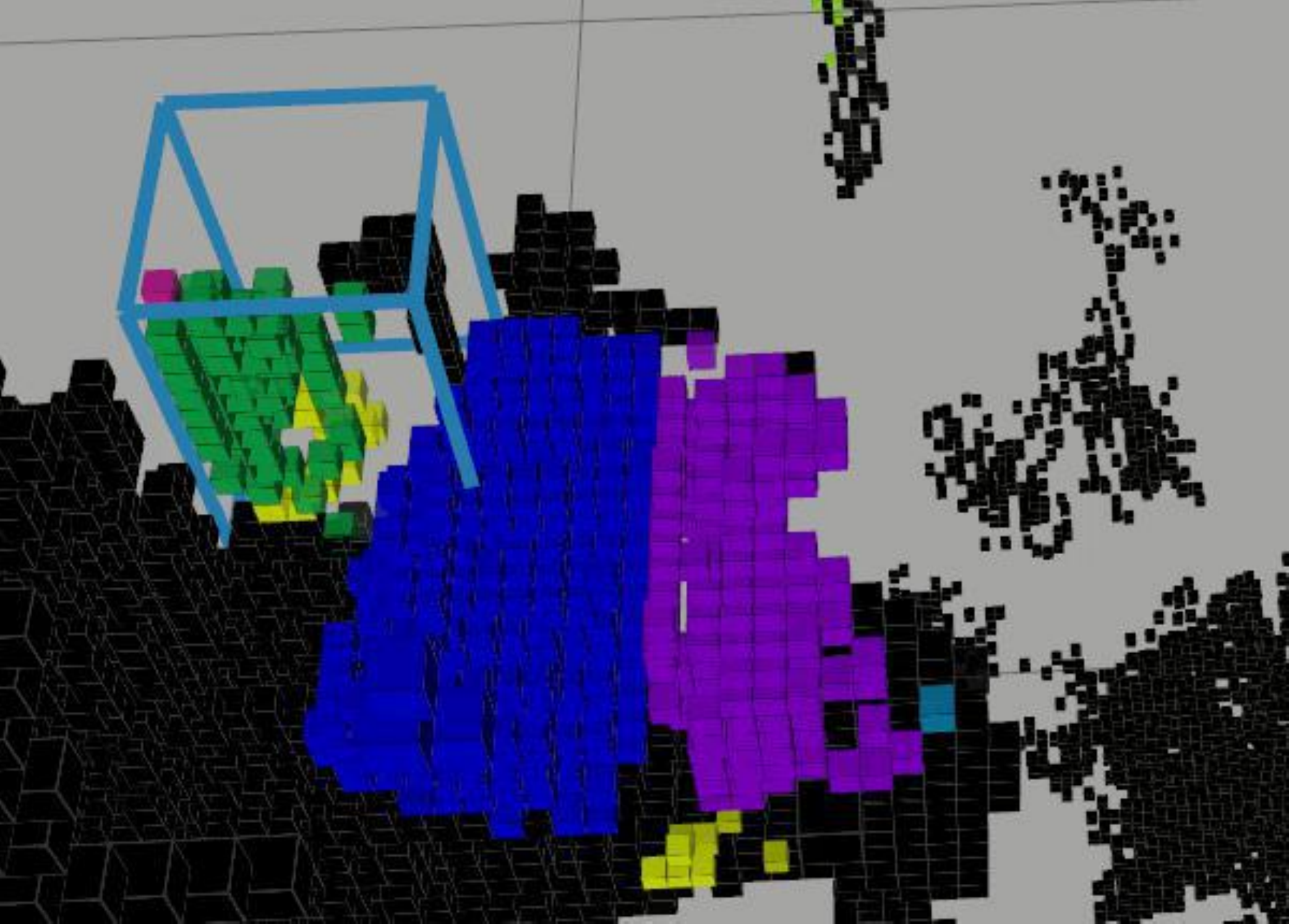} \\
  35          &                                        16.18 &                                        17.90 &                      \textbf{20.07} &  \tabimg{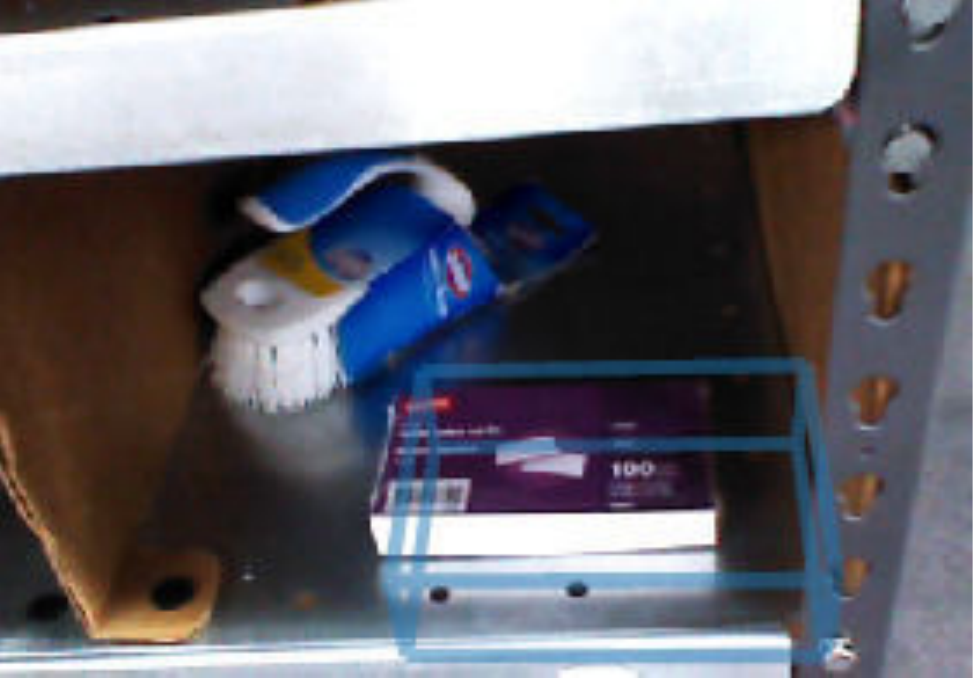} &  \tabimg{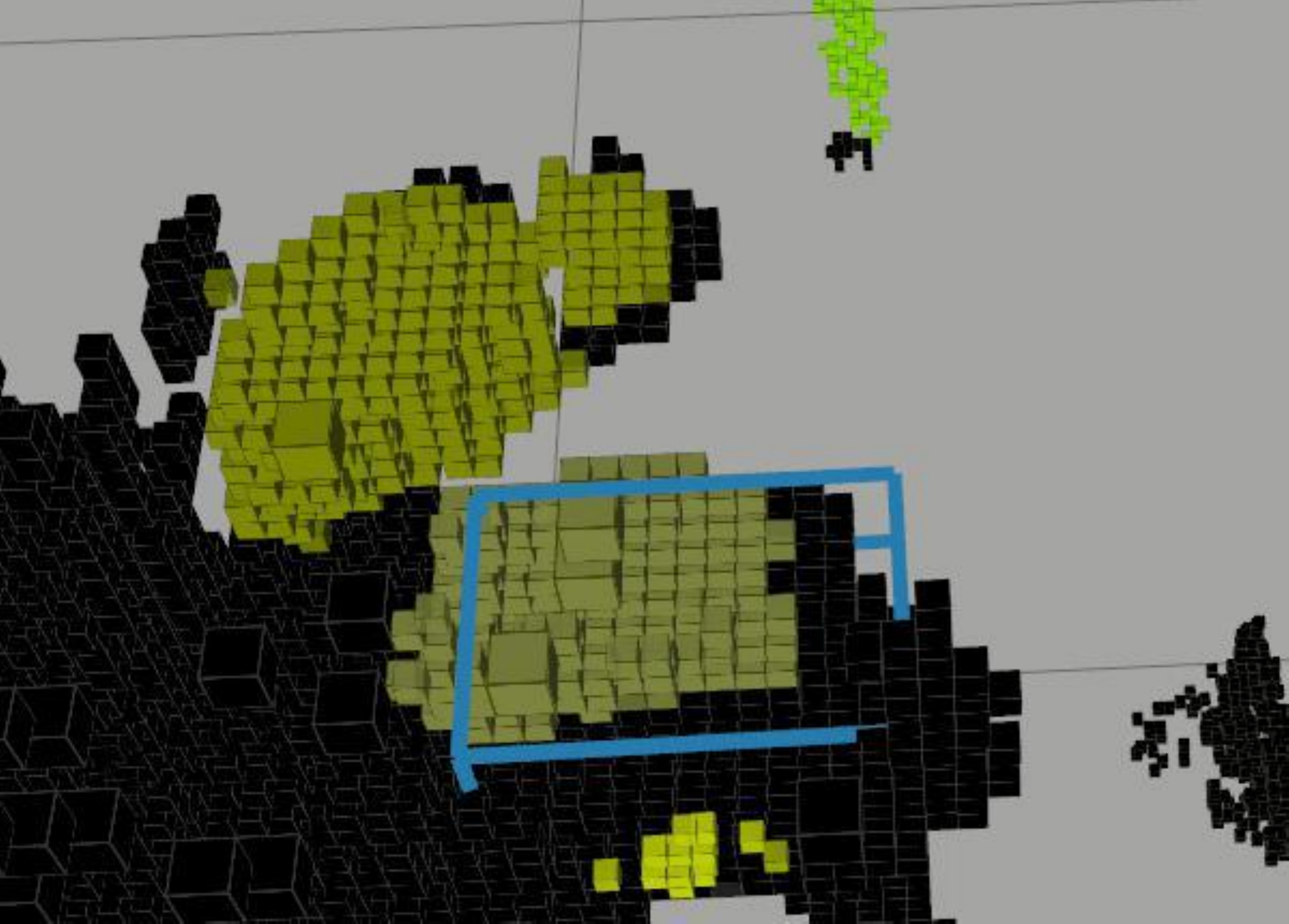} \\
  36          &                                        21.07 &                                        24.00 &                      \textbf{36.99} &  \tabimg{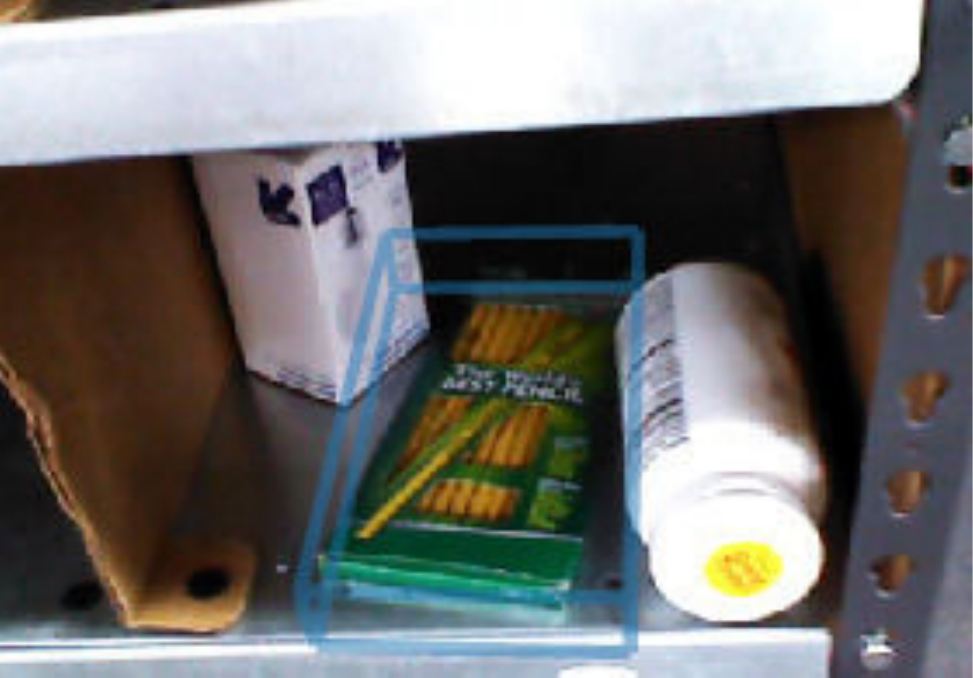} &  \tabimg{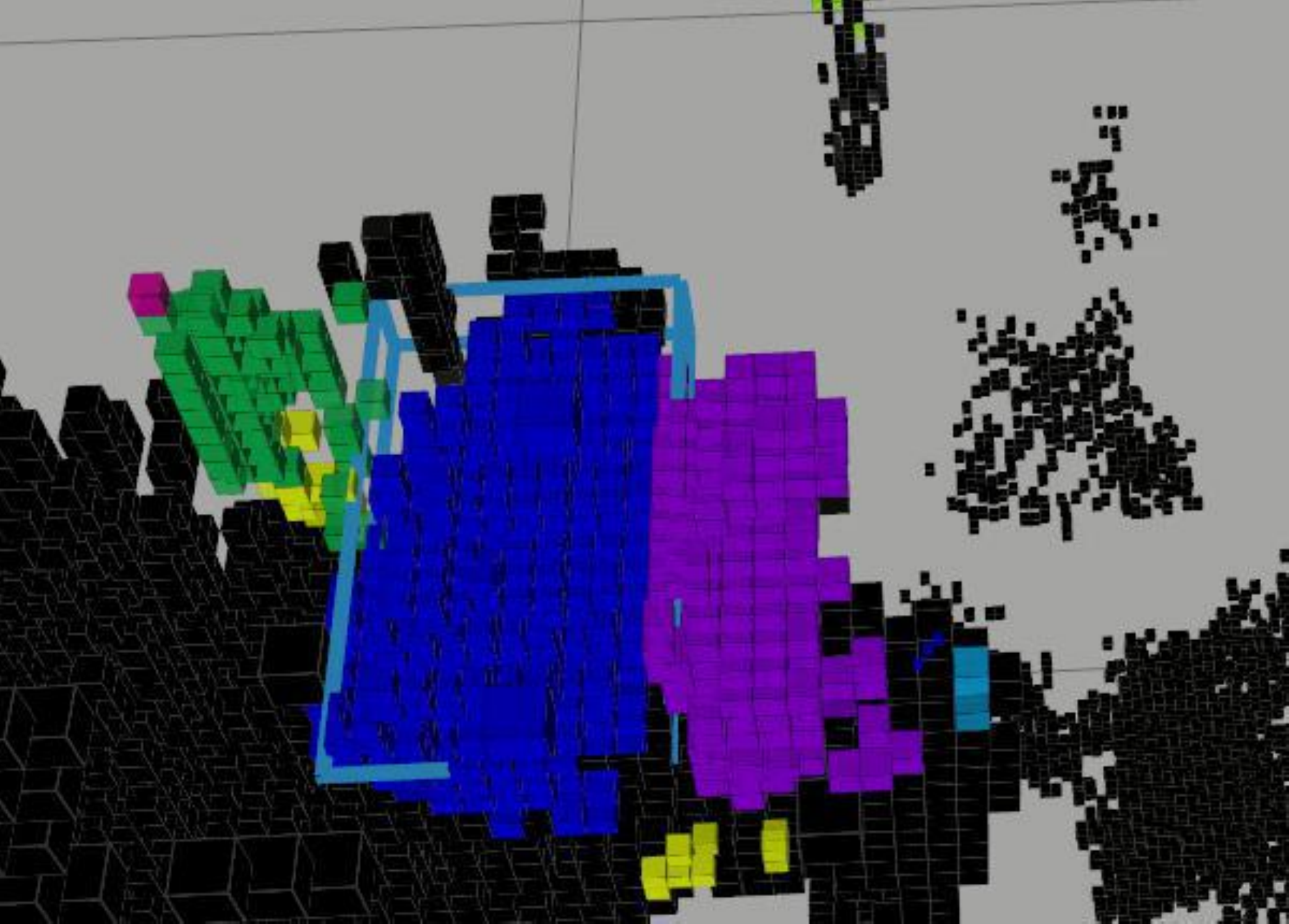} \\
  37          &                                         5.81 &                                         8.15 &                      \textbf{12.19} &  \tabimg{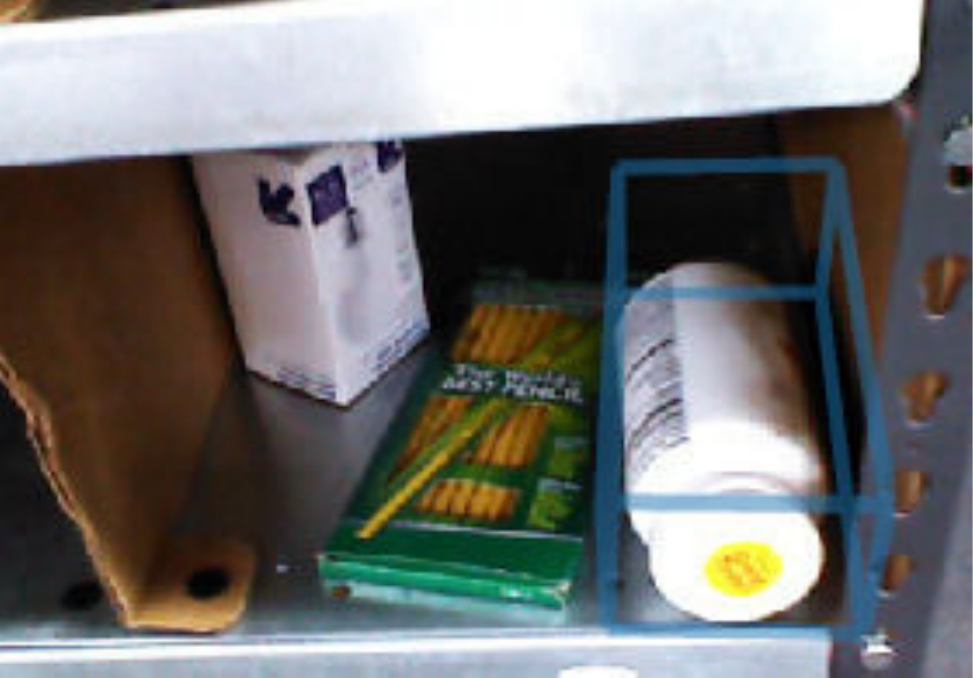} &  \tabimg{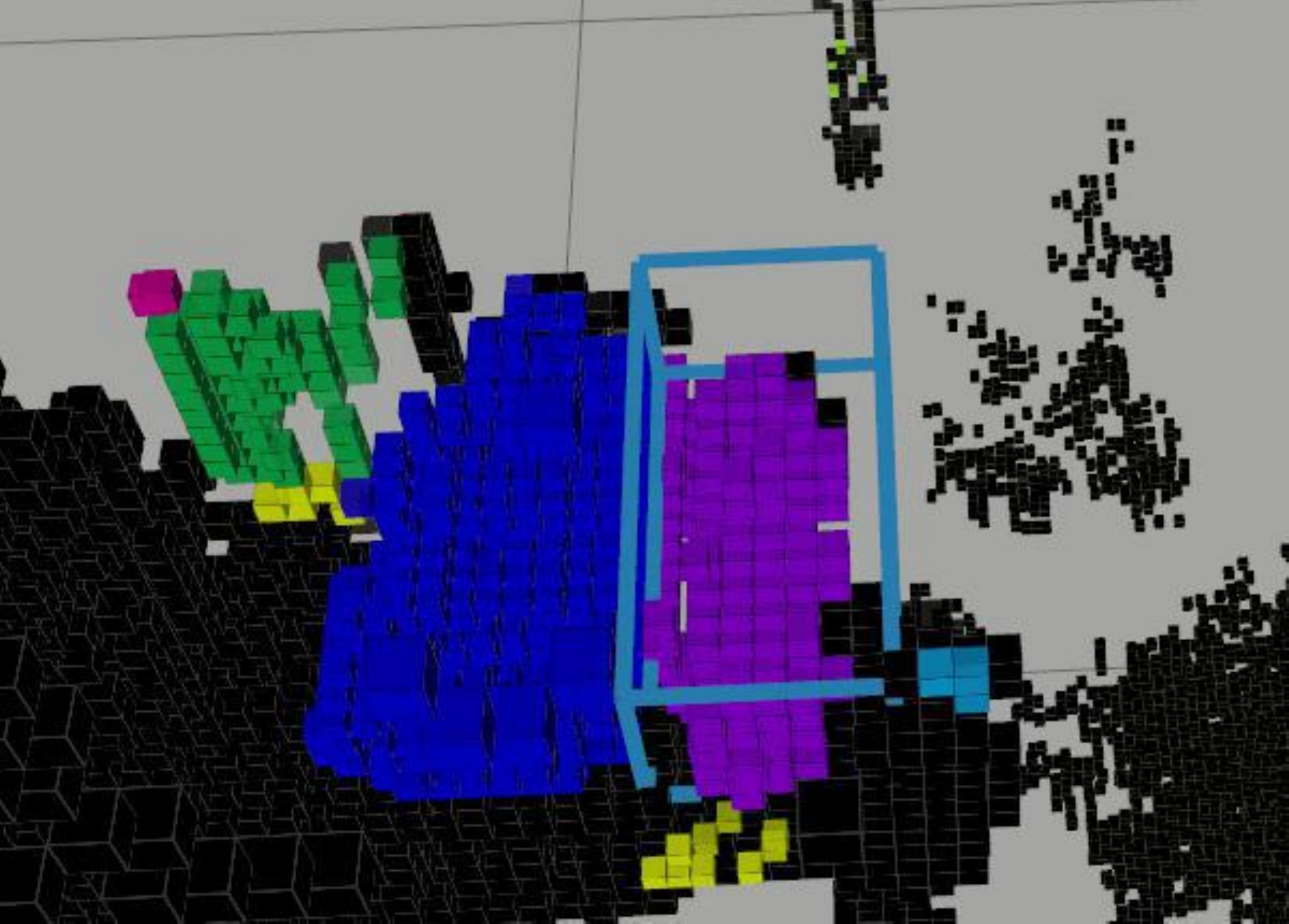} \\
  38          &                                        17.53 &                                        22.70 &                      \textbf{33.32} &  \tabimg{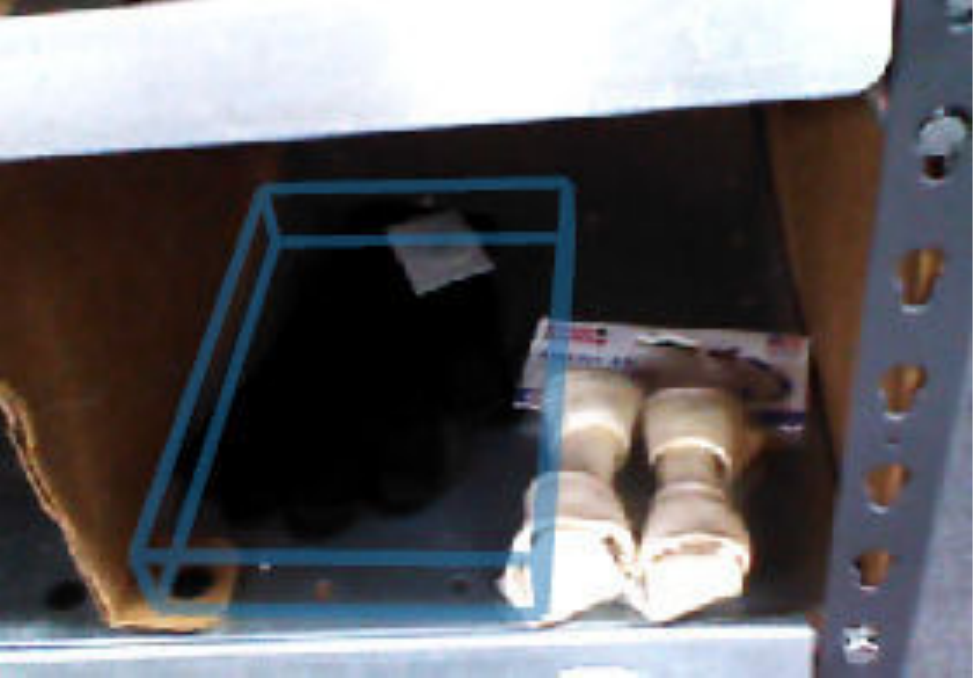} &  \tabimg{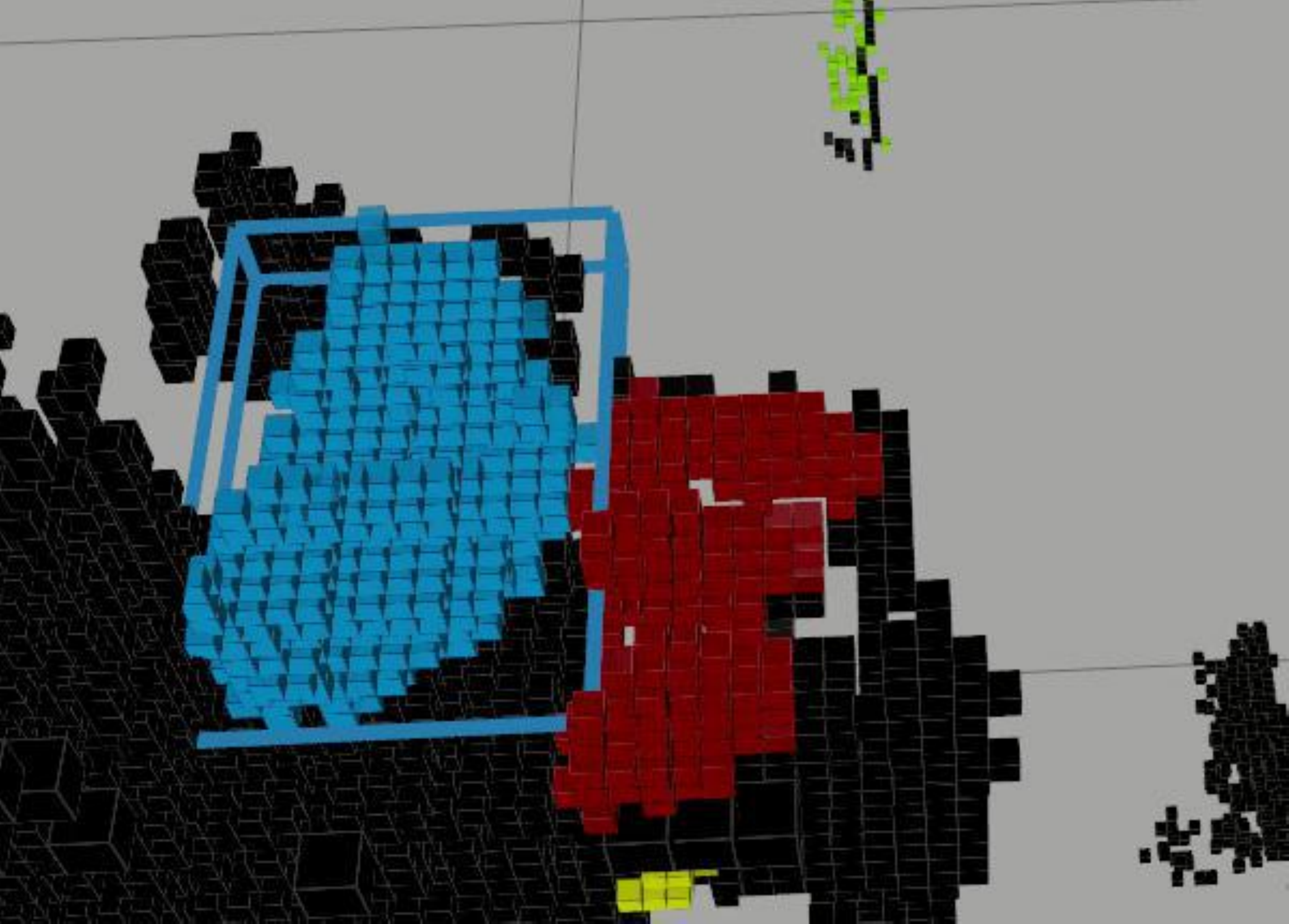} \\
  39          &                                         0.22 &                                         0.95 &                       \textbf{1.81} &  \tabimg{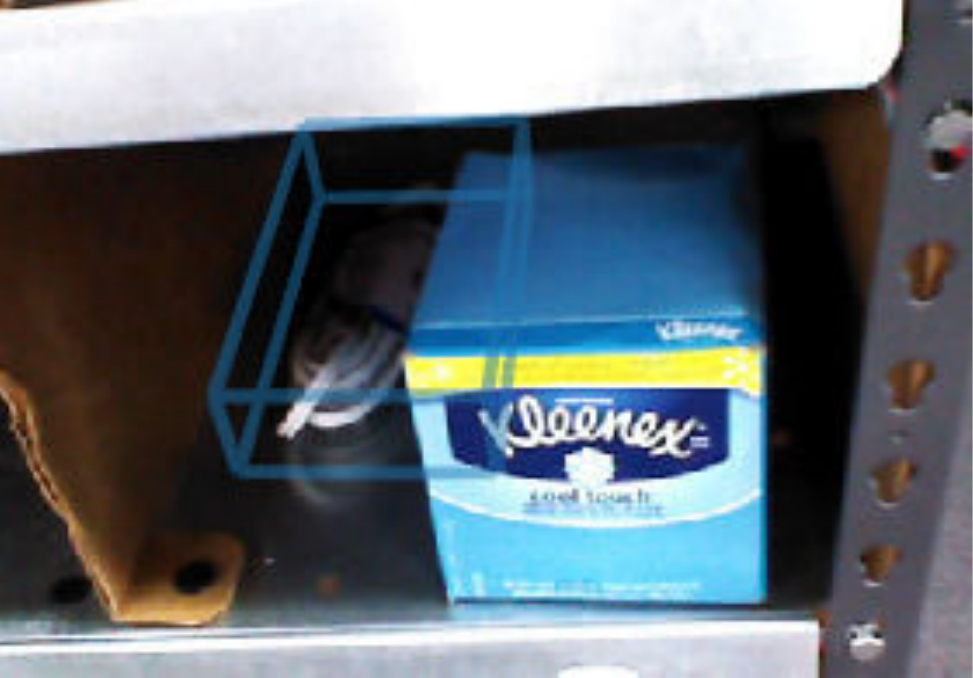} &  \tabimg{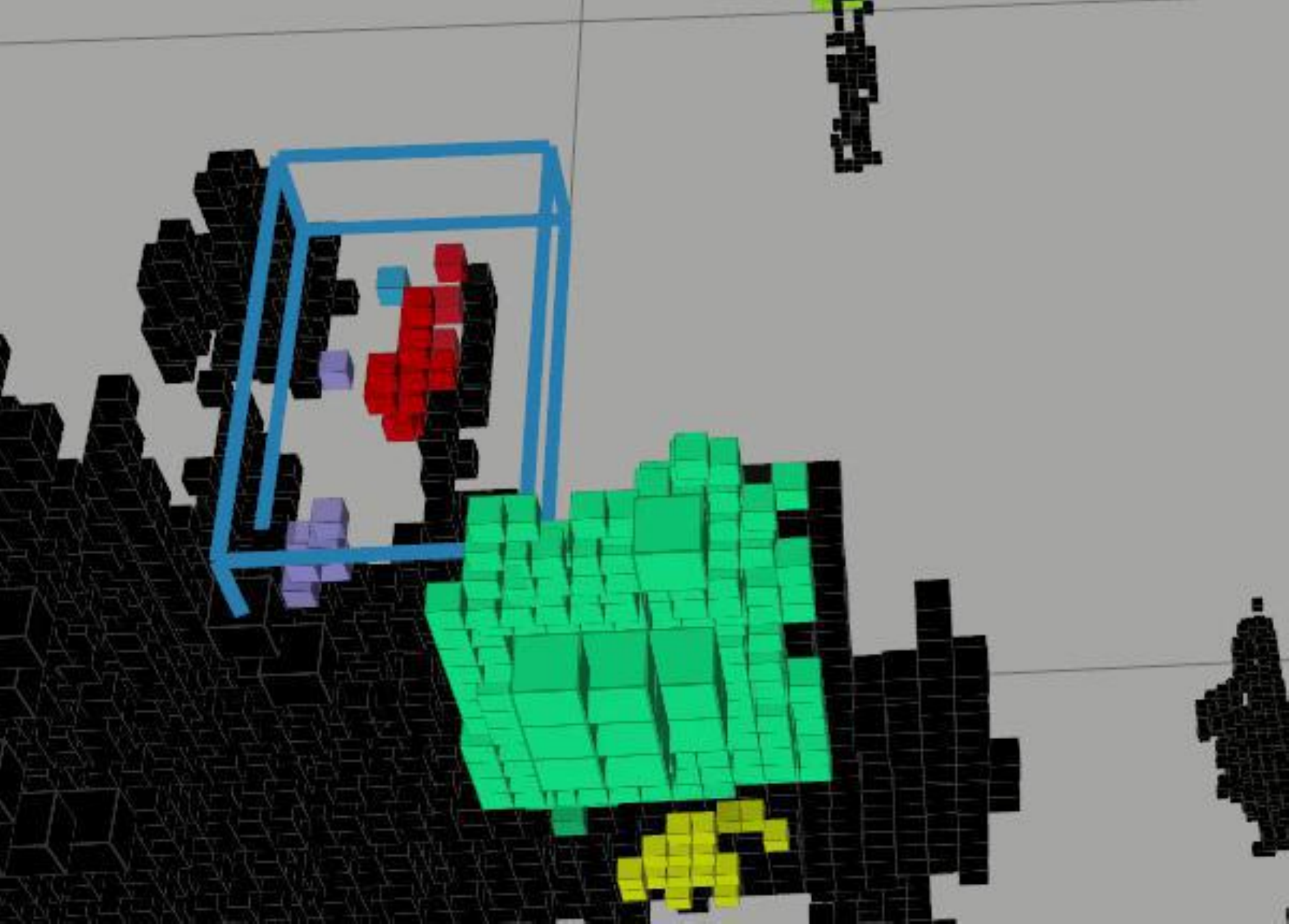} \\
  \vspace{0.7cm}
  \end{tabular}
  \end{minipage}
\end{table*}

% 6. Conclusions
\section{Conclusions}

We presented a method to segment multilabel objects in real world three-dimensionally,
generating object voxels with probabilistic object label representation.
The contributions of this paper are:
1) proposed a 3D segmentation method with mapping multilabel occupancy in real-time
2) evaluated the applicability of the segmentation method in the multi-object manipulation experiment.
By using the proposed method,
% we improved the accuracy of 3D object segmentation in environments with occlusions,
we achieved 3D multi-class segmentation in environments with occlusions,
and picking task which requires multi-object manipulation.
% the candidate objects which needs to be manipulated are segmented,
% and the manipulation task is completed successfully
% with grasps planned from generated voxels.

The future work would be the extension of the proposed method
to object instances in the same object class, because
currently the method is unable to segment multiple instances if they are close to each other.
The optimization of viewpoints and extension to dynamic objects which moves during multi-view
also should be addressed.

% \addtolength{\textheight}{-12cm} % This command serves to balance the column lengths

% \input src/appendix.tex
% \input src/acknowledge.tex

\bibliographystyle{junsrt}
\bibliography{main}

\end{document}